\documentclass{article} % For LaTeX2e
\usepackage{nips12submit_e,times}
\usepackage{graphicx} % more modern
\usepackage{epstopdf}

\usepackage[T1]{fontenc}
\usepackage[utf8]{inputenc}

\usepackage{algorithm}
\usepackage{algorithmic}

\usepackage{subfigure}
\usepackage{caption}
\usepackage{epsf}
\usepackage{amsmath}
\usepackage{amssymb}
\usepackage{amsfonts}
\usepackage{amsthm}

\newlength{\minipagewidth}
\newlength{\minipagewidthx}
\newcommand{\bookbox}[1]{\small
\par\medskip\noindent
\framebox[\columnwidth]{
\begin{minipage}{\minipagewidth} {#1} \end{minipage} } \par\medskip }

\title{Risk--Aversion in Multi--armed Bandits}
\author{Amir Sani $\qquad\qquad$ Alessandro Lazaric $\qquad\qquad$ R\'{e}mi Munos\\
INRIA Lille - Nord Europe, Team SequeL\\
\texttt{\{amir.sani,alessandro.lazaric,remi.munos\}@inria.fr}
}

\newcommand{\A}{\mathcal A}

\newcommand{\E}{\mathbb E}
\newcommand{\calE}{\mathcal E}

\newcommand{\Prob}{\mathbb P}
\newcommand{\I}{\mathbb I}

\newcommand{\bmu}{\boldsymbol \mu}

\newcommand{\bx}{\boldsymbol{x}}
\newcommand{\MV}{\textnormal{MV}}
\newcommand{\hMV}{\widehat{\textnormal{MV}}}

\newcommand{\hmu}{\hat\mu}
\newcommand{\tmu}{\tilde\mu}

\newcommand{\hvar}{\hat\sigma^2}
\newcommand{\tvar}{\tilde\sigma^2}
\newcommand{\R}{\mathcal{R}}

\newcommand{\tR}{\widetilde{\mathcal R}}
\newcommand{\var}{\sigma^2}
\newcommand{\invdelta}{1/\delta}

\newcommand{\eps}{\varepsilon}

\newcommand{\avg}[2]{\frac{1}{#2} \sum_{#1=1}^{#2}}
\newcommand{\hDelta}{\widehat{\Delta}}
\newcommand{\hGamma}{\widehat{\Gamma}}

\newtheorem{lemma}{Lemma}

\newtheorem{definition}{Definition}
\newtheorem{theorem}{Theorem}
\newcommand{\remark}[1]{\textbf{Remark #1.}}

\nipsfinalcopy % Uncomment for camera-ready version

\begin{document}
\maketitle

\begin{abstract}
Stochastic multi--armed bandits solve the Exploration--Exploitation dilemma and ultimately maximize the expected reward. Nonetheless, in many practical problems, maximizing the expected reward is not the most desirable objective. In this paper, we introduce a novel setting based on the principle of risk--aversion where the objective is to compete against the arm with the best risk--return trade--off. This setting proves to be intrinsically more difficult than the standard multi-arm bandit setting due in part to an exploration risk which introduces a regret associated to the variability of an algorithm. Using variance as a measure of risk, we introduce two new algorithms, investigate their theoretical guarantees, and report preliminary empirical results.
\end{abstract}

%%%%%%%%%%%%%%%%%%%%%%%%%%%%%%%%%%%%%%%%%%%%%%%%%%%%%%%%%%%%%%%%%%%%%%%%%%%%%%%
%% Introduction
%%%%%%%%%%%%%%%%%%%%%%%%%%%%%%%%%%%%%%%%%%%%%%%%%%%%%%%%%%%%%%%%%%%%%%%%%%%%%%%

\section{Introduction}\label{s:intro}

The multi--armed bandit~\cite{robbins1952some} elegantly formalizes the problem of on--line learning with partial feedback, which encompasses a large number of real--world applications, such as clinical trials, online advertisements, adaptive routing, and cognitive radio. In the stochastic multi--armed bandit model, a learner chooses among several arms (e.g., different treatments), each characterized by an independent reward distribution (e.g., the treatment effectiveness). At each point in time, the learner selects one arm and receives a noisy reward observation from that arm (e.g., the effect of the treatment on one patient). Given a finite number of $n$ rounds (e.g., patients involved in the clinical trial), the learner faces a dilemma between repeatedly exploring all arms and collecting reward information versus exploiting current reward estimates by selecting the arm with the highest estimated reward. Roughly speaking, the learning objective is to solve this exploration--exploitation dilemma and accumulate as much reward as possible over $n$ rounds. In particular, multi--arm bandit literature typically focuses on the problem of finding a learning algorithm capable of maximizing the expected cumulative reward (i.e., the reward collected over $n$ rounds averaged over all possible observation realizations), thus implying that the best arm returns the highest expected reward. Nonetheless, in many practical problems, maximizing the expected reward is not always the most desirable objective. For instance, in clinical trials, the treatment which works best \textit{on average} might also have considerable \textit{variability}; resulting in adverse side effects for some patients. In this case, a treatment which is less effective on average but consistently effective on different patients may be preferable w.r.t. an effective but risky treatment. More generally, some application objectives require an effective trade--off between risk and reward.

There is no agreed upon definition for risk. A variety of behaviours result in an uncertainty which might be deemed unfavourable for a specific application and referred to as a risk. For example, a solution with guarantees over multiple runs of an algorithm may not satisfy the desire for a solution with low variability over a single implementation of an algorithm. Two foundational risk modeling paradigms are Expected Utility theory \cite{Neumann1947} and the historically popular and accessible Mean-Variance paradigm \cite{markowitz1952portfolio}. A large part of decision--making theory focuses on defining and managing risk (see e.g., \cite{gollier2001the-economics} for an introduction to risk from an expected utility theory perspective).

Risk has mostly been studied in on--line learning within the so--called expert advice setting (i.e., adversarial full--information on--line learning). In particular, \cite{even-dar2006risk-sensitive} showed that in general, although it is possible to achieve a small regret w.r.t. to the expert with the best average performance, it is not possible to compete against the expert which best trades off between average return and risk. On the other hand, it is possible to define no--regret algorithms for simplified measures of risk--return. \cite{warmuth2006online} studied the case of pure risk minimization (notably variance minimization) in an on-line setting where at each step the learner is given a covariance matrix and must choose a weight vector that minimizes the variance. The regret is then computed over horizon $n$ and compared to the fixed weights minimizing the variance in hindsight. In the multi--arm bandit domain, the most interesting results are by \cite{audibert2009exploration-exploitation} and \cite{salomon2011deviations}. \cite{audibert2009exploration-exploitation} introduced an analysis of the expected regret and its distribution, revealing that an anytime version of \textsl{UCB}~\cite{auer2002finite-time} and \textsl{UCB-V} might have large regret with some non-negligible probability.\footnote{Although the analysis is mostly directed to the pseudo--regret, as commented in Remark 2 at page 23 of \cite{audibert2009exploration-exploitation}, it can be extended to the true regret.} This analysis is further extended by~\cite{salomon2011deviations} who derived negative results which show no anytime algorithm can achieve a regret with both a small expected regret and exponential tails. Although these results represent an important step towards the analysis of risk within bandit algorithms, they are limited to the case where an algorithm's cumulative reward is compared to the reward obtained by pulling the arm with the highest expectation.

In this paper, we focus on the problem of competing against the arm with the best risk--return trade--off. In particular, we refer to the first and most popular measure of risk--return, the mean--variance model introduce by~\cite{markowitz1952portfolio}. 
%
%The rest of the paper is organized as follows. 
In Section~\ref{s:multi-arm} we introduce notation and define the mean--variance bandit problem. In Section~\ref{s:mv-lcb} we introduce a confidence--bound algorithm and study its theoretical properties. In Section~\ref{s:simulations} we report a set of numerical simulations aiming at validating the theoretical results. Finally, in Section~\ref{s:conclusions} we conclude with a discussion on possible extensions. The proofs and additional experiments are reported in the appendix.

%%%%%%%%%%%%%%%%%%%%%%%%%%%%%%%%%%%%%%%%%%%%%%%%%%%%%%%%%%%%%%%%%%%%%%%%%%%%%%%
%% Finding the Best Risk-Averse Arm
%%%%%%%%%%%%%%%%%%%%%%%%%%%%%%%%%%%%%%%%%%%%%%%%%%%%%%%%%%%%%%%%%%%%%%%%%%%%%%%

\vspace{-0.1in}
\section{Mean--Variance Multi--arm Bandit}\label{s:multi-arm}
\vspace{-0.1in}

In this section we introduce the main notation used throughout the paper and define the mean--variance multi--arm bandit problem.

%% The setting

%\subsection{The Setting}

We consider the standard multi--arm bandit setting with $K$ arms, each characterized by a distribution $\nu_i$ bounded in the interval $[0,1]$. Each distribution has a mean $\mu_{i}$ and a variance $\var_{i}$. The bandit problem is defined over a finite horizon of $n$ rounds. We denote by $X_{i,s}\sim \nu_i$ the $s$-th random sample drawn from the distribution of arm $i$. All arms and samples are independent. In the multi--arm bandit protocol, at each round $t$, an algorithm selects arm $I_t$ and observes sample $X_{I_t,T_{i,t}}$, where $T_{i,t}$ is the number of samples observed from arm $i$ up to time $t$ (i.e., $T_{i,t} = \sum_{s=1}^t \I\{I_t = i\}$). 

%% The mean-variance bandit problem

%\subsection{The Mean--Variance Bandit Problem}

While in the standard literature on multi--armed bandits the objective is to select the arm leading to the highest reward in \textit{expectation} (the arm with the largest expected value $\mu_{i}$), here we focus on the problem of finding the arm which effectively trades off between its expected reward (i.e., the \textit{return}) and its variability (i.e., the \textit{risk}). Although a large number of models for risk--return trade--off have been proposed, here we focus on the most historically popular and simple model: the mean--variance model proposed by \cite{markowitz1952portfolio},\footnote{We discuss the limitations of this model and possible extensions to other models of risk in Section~\ref{s:conclusions}.} where the return of an arm is measured by the expected reward and its risk by its variance.

\begin{definition}\label{d:mv}
The mean--variance of an arm $i$ with mean $\mu_i$, variance $\var_i$ and coefficient of absolute risk tolerance $\rho$ is defined as\footnote{The coefficient of risk tolerance is the inverse of the more popular coefficient of risk aversion $A=1/\rho$.} $\MV_i =  \var_i - \rho\mu_i$.
%
%\begin{align}\label{eq:mv}
%
%\end{align}
\end{definition}

Thus it easily follows that the best arm minimizes the mean--variance, that is $i^* = \arg\min_{i=1,\ldots,K} \MV_i$.
We notice that we can obtain two extreme settings depending on the value of risk tolerance $\rho$. As $\rho\rightarrow \infty$, the mean--variance of arm $i$ tends to the opposite of its expected value $\mu_i$ and the problem reduces to the standard expected reward maximization traditionally considered in multi--arm bandit problems. With $\rho=0$, the mean--variance reduces to minimizing the variance $\var_i$ and the objective becomes variance minimization.

Given $\{X_{i,s}\}_{s=1}^t$ i.i.d. samples from the distribution $\nu_i$, we define the empirical mean--variance of an arm $i$ with $t$ samples as $\hMV_{i,t} =  \hvar_{i,t} - \rho\hmu_{i,t}$,
%%
%\begin{align}\label{eq:hmv}
%,
%\end{align}
%%
where 
\begin{align}\label{eq:hmu.hvar}
\hmu_{i,t} = \avg{s}{t} X_{i,s}, \quad\quad \hvar_{i,t} = \frac{1}{t}\sum_{s=1}^t \big(X_{i,s} - \hmu_{i,t}\big)^2.
\end{align}
We now consider a learning algorithm $\A$ and its corresponding performance over $n$ rounds. Similar to a single arm $i$ we define its empirical mean--variance as
\begin{align}\label{eq:hmv.alg}
\hMV_{n}(\A) =  \hvar_{n}(\A) - \rho\hmu_{n}(\A),
\end{align}
where
\begin{align}\label{eq:hmu.hvar.alg}
\hmu_{n}(\A) = \avg{t}{n} Z_t, \quad\quad \hvar_n(\A) = \frac{1}{n}\sum_{t=1}^n \big(Z_t - \hmu_n(\A)\big)^2,
\end{align}
with $Z_t = X_{I_t, T_{i,t}}$, that is the reward collected by the algorithm at time $t$. This leads to a natural definition of the (random) regret at each single run of the algorithm as the difference in the mean--variance performance of the algorithm compared to the best arm.

\begin{definition}\label{d:regret}
The regret for a learning algorithm $\A$ over $n$ rounds is defined as
\begin{align}\label{eq:regret}
\R_n(\A) &=  \hMV_n(\A) - \hMV_{i^*,n}.
\end{align}
\end{definition}
Given this definition, the objective is to design an algorithm whose regret decreases as the number of rounds increases (in high probability or in expectation).

We notice that the previous definition actually depends on \textit{unobserved} samples. In fact, $\hMV_{i^*,n}$ is computed on $n$ samples $i^*$ which are not actually observed when running $\A$. This matches the definition of \textit{true} regret in standard bandits (see e.g.,~\cite{audibert2009exploration-exploitation}). Thus, in order to clarify the main components characterizing the regret, we introduce additional notation. Let
\begin{align*}
Y_{i,t} = \begin{cases} 
X_{i^*,t} &\mbox{if } i=i^* \\ 
X_{i^*,t'} \mbox{ with } t'=T_{i^*,n}+\sum\limits_{j < i, j\neq i^*} T_{j,n} + t & \mbox{otherwise} 
\end{cases}
\end{align*}
be a renaming of the samples from the optimal arm, such that while the algorithm was pulling arm $i$ for the $t$-th time, $Y_{i,t}$ is the unobserved sample from $i^*$. Then we define the corresponding mean and variance as
\begin{align}\label{eq:tmu.tvar}
\tmu_{i,T_{i,n}} = \avg{t}{T_{i,n}} Y_{i,t}, \quad\quad \tvar_{i,T_{i,n}} = \frac{1}{T_{i,n}}\sum_{t=1}^{T_{i,n}} \big(Y_{i,t} - \tmu_{i,T_{i,n}}\big)^2.
\end{align}
Given these additional definitions, we can rewrite the regret as (see Appendix~\ref{app:true.regret})
\begin{align}\label{eq:regret2}
\R_n(\A) &= \frac{1}{n}\sum_{i\neq i^*} T_{i,n} \Big[(\hvar_{i,T_{i,n}}-\rho\hmu_{i,T_{i,n}}) - (\tvar_{i,T_{i,n}} - \rho\tmu_{i,T_{i,n}})\Big] \nonumber\\
&\quad+ \frac{1}{n}\sum_{i=1}^K T_{i,n} \big(\hmu_{i,T_{i,n}} - \hmu_n(\A)\big)^2 - \frac{1}{n}\sum_{i=1}^K T_{i,n} \big(\tmu_{i,T_{i,n}} - \hmu_{i^*,n}\big)^2.
\end{align}

Since the last term is always negative and small~\footnote{More precisely, it can be shown that this term decreases with rate $O(K\log(1/\delta)/n)$ with probability $1-\delta$.}, our analysis focuses on the first two terms which reveal two interesting characteristics of $\A$. First, an algorithm $\A$ suffers a regret whenever it chooses a suboptimal arm $i\neq i^*$ and the regret corresponds to the difference in the empirical mean--variance of $i$ w.r.t. the optimal arm $i^*$. Such a definition has a strong similarity to the standard definition of regret, where $i^*$ is the arm with highest expected value and the regret depends on the number of times suboptimal arms are pulled and their respective gaps w.r.t. the optimal arm $i^*$. In contrast to the standard formulation of regret, $\A$ also suffers an additional regret from the variance $\hvar_n(\A)$, which depends on the variability of pulls $T_{i,n}$ over different arms. Recalling the definition of the mean $\hmu_n(\A)$ as the weighted mean of the empirical means $\hmu_{i,T_{i,n}}$ with weights $T_{i,n}/n$ (see eq.~\ref{eq:hmu.hvar.alg}), we notice that this second term is a weighted variance of the means and illustrates the exploration risk of the algorithm. In fact, if an algorithm simply selects and pulls a single arm from the beginning, it would not suffer any exploration risk (secondary regret) since $\hmu_n(\A)$ would coincide with $\hmu_{i,T_{i,n}}$ for the chosen arm and all other components would have zero weight. On the other hand, an algorithm accumulates exploration risk through this second term as the mean $\hmu_n(\A)$ deviates from any specific arm; where the maximum exploration risk peaks at the mean $\hmu_n(\A)$ furthest from all arm means.
%In the next sections we introduce and study two simple algorithm and we study how well they trade off the two components of the regret.

The previous definition of regret can be further elaborated to obtain the upper bound (see App.~\ref{app:true.regret})
\begin{align}\label{eq:regret3}
\R_n(\A) \leq \frac{1}{n}\sum_{i\neq i^*} T_{i,n} \hDelta_i + \frac{1}{n^2}\sum_{i=1}^K\sum_{j\neq i} T_{i,n}T_{j,n} \hGamma_{i,j}^2,
\end{align}
where $\hDelta_i = (\hvar_{i,T_{i,n}}-\tvar_{i,T_{i,n}}) - \rho(\hmu_{i,T_{i,n}} - \tmu_{i,T_{i,n}})$ and $\hGamma_{i,j}^2 = (\hmu_{i,T_{i,n}}-\hmu_{j,T_{j,n}})^2$. Unlike the definition in eq.~\ref{eq:regret2}, this upper bound explicitly illustrates the relationship between the regret and the number of pulls $T_{i,n}$; suggesting that a bound on the pulls is sufficient to bound the regret. 
%This formulation also allows us to have a better understanding of how the regret is composed. Let consider the case of $\rho = 0$ (variance minimization problem). In this case, $\hDelta_i$ represents the different in the empirical variances and $\hGamma_{i,j}$ is the difference in the empirical means. Even in a problem where all the arms have a zero variance (i.e., $\hDelta_i = 0$), an algorithm pulling all the arms uniformly would suffer a constant regret due to the variance introduced by pulling arms with different means. 

Finally, we can also introduce a definition of the pseudo-regret.
\begin{definition}\label{d:pseudo.regret}
The pseudo regret for a learning algorithm $\A$ over $n$ rounds is defined as
\begin{align}\label{eq:pseudo.regret}
\tR_n(\A) = \frac{1}{n}\sum_{i\neq i^*} T_{i,n} \Delta_i + \frac{2}{n^2}\sum_{i=1}^K\sum_{j\neq i} T_{i,n}T_{j,n} \Gamma_{i,j}^2,
\end{align}
where $\Delta_i = \MV_i - \MV_{i^*}$ and $\Gamma_{i,j} = \mu_i - \mu_j$.
\end{definition}
In the following, we denote the two components of the pseudo--regret as
\begin{align}\label{eq:regrets}
\tR^{\Delta}_n(\A) = \frac{1}{n}\sum_{i\neq i^*} T_{i,n} \Delta_i, \quad\text{ and }\quad \tR^{\Gamma}_n(\A) = \frac{2}{n^2}\sum_{i=1}^K\sum_{j\neq i} T_{i,n}T_{j,n} \Gamma_{i,j}^2.
\end{align}

Where $\tR^{\Delta}_n(\A)$ constitutes the standard regret derived from the traditional formulation of the multi-arm bandit problem and $\tR^{\Gamma}_n(\A)$ denotes the exploration risk. This regret can be shown to be close to the true regret up to small terms with high probability.
\begin{lemma}\label{l:psuedo.regret}
Given definitions~\ref{d:regret} and \ref{d:pseudo.regret},
\begin{align*}
\R_n(\A) \leq \tR_n(\A) + (5+\rho)\sqrt{\frac{2K\log (6nK/\delta)}{n}} + 4\sqrt{2}\frac{K\log (6nK/\delta)}{n},
\end{align*}
with probability at least $1-\delta$.
\end{lemma}

The previous lemma shows that any (high--probability) bound on the pseudo--regret immediately translates into a bound on the true regret. Thus, we report most of the theoretical analysis according to $\tR_n(\A)$. Nonetheless, it is interesting to notice the major difference between the true and pseudo--regret when compared to the standard bandit problem. In fact, it is possible to show in the risk--averse case that the pseudo--regret is not an unbiased estimator of the true regret, i.e., $\E[\R_n] \neq \E[\tR_n]$. Thus, in order to bound the expectation of $\R_n$ we build on the high--probability result from Lemma~\ref{l:psuedo.regret}.

%%%%%%%%%%%%%%%%%%%%%%%%%%%%%%%%%%%%%%%%%%%%%%%%%%%%%%%%%%%%%%%%%%%%%%%%%%%%%%%
%% The Mean--Variance Lower Confidence Bound Algorithm
%%%%%%%%%%%%%%%%%%%%%%%%%%%%%%%%%%%%%%%%%%%%%%%%%%%%%%%%%%%%%%%%%%%%%%%%%%%%%%%

\section{The Mean--Variance Lower Confidence Bound Algorithm}\label{s:mv-lcb}

In this section we introduce a novel risk--averse bandit algorithm whose objective is to identify the arm which best trades off risk and return. The algorithm is a natural extension of \textsl{UCB1}~\cite{auer2002finite-time} and we report a theoretical performance analysis on how well it balances the exploration needed to identify the best arm versus the risk of pulling arms with different means.

%% The algorithm

\subsection{The Algorithm}

\begin{figure}[t]
\bookbox{
\begin{algorithmic}
\STATE \textbf{Input:} Confidence $\delta$
\FOR{$t = 1,\ldots,n$}
\FOR{$i = 1,\ldots,K$}
\STATE Compute $B_{i,T_{i,t-1}} = \hMV_{i,T_{i,t-1}} - (5+\rho) \sqrt{\frac{\log\invdelta}{2T_{i,t-1}}}$ 
\ENDFOR
\STATE Return $I_t = \arg\min_{i=1,\ldots,K} B_{i,T_{i,t-1}}$
\STATE Update $T_{i,t} = T_{i,t-1}+1$
\STATE Observe $X_{I_t, T_{i,t}} \sim \nu_{I_t}$
\STATE Update $\hMV_{i,T_{i,t}}$
\ENDFOR
\end{algorithmic}}
\vspace{-0.1in}
\caption{Pseudo-code of the \textsl{MV-LCB} algorithm.}\label{f:lcb}
\vspace{-0.15in}
\end{figure}

We propose an index--based bandit algorithm which estimates the mean--variance of each arm and selects the optimal arm according to the optimistic confidence--bounds on the current estimates. A sketch of the algorithm is reported in Figure~\ref{f:lcb}.
For each arm, the algorithm keeps track of the empirical mean--variance $\hMV_{i,s}$ computed according to $s$ samples. We can build high--probability confidence bounds on empirical mean--variance through an application of the Chernoff--Hoeffding inequality (see e.g.,~\cite{antos2010active} for the bound on the variance) on terms $\hmu$ and $\hvar$.

\begin{lemma}\label{l:ch.mean-var}
Let $\{X_{i,s}\}$ be i.i.d.~random variables bounded in $[0,1]$ from the distribution $\nu_i$ with mean $\mu_i$ and variance $\var_i$, and the empirical mean $\hmu_{i,s}$ and variance $\hvar_{i,s}$ computed as in Equation~\ref{eq:hmu.hvar}, then
\begin{equation*}\label{eq:ch.mean}
\Prob\Bigg[\exists i=1,\ldots,K, s=1,\ldots,n, |\hMV_{i,s}-\MV_i| \geq (5+\rho)\sqrt{\frac{\log\invdelta}{2s}}\Bigg] \leq 6nK\delta,
\end{equation*}
\end{lemma}

The algorithm in Figure~\ref{f:lcb} implements the principle of optimism in the face of uncertainty used in many multi--arm bandit algorithms. On the basis of the previous confidence bounds, we define a lower--confidence bound on the mean--variance of arm $i$ when it has been pulled $s$ times as
\begin{align}\label{eq:index}
B_{i, s} = \hMV_{i,s} - (5+\rho) \sqrt{\frac{\log\invdelta}{2s}},
\end{align}
where $\delta$ is an input parameter of the algorithm. Given the index of each arm at each round $t$, the algorithm simply selects the arm with the smallest mean--variance index, i.e., $I_t = \arg\min_{i} B_{i,T_{i,t-1}}$. We refer to this algorithm as the mean--variance lower--confidence bound (\textsl{MV-LCB}) algorithm.

\remark{1} We notice that the algorithm reduces to \textsl{UCB1} whenever $\rho\rightarrow\infty$. This is coherent with the fact that for $\rho\rightarrow\infty$ the mean--variance problem reduces to the maximization of the cumulative reward, for which \textsl{UCB1} is already known to be nearly-optimal. On the other hand, for $\rho=0$, which leads to the problem of cumulative reward variance minimization, the algorithm plays according to a lower--confidence--bound on the variances.

%\remark{2} The algorithm can be easily improved by using tighter bounds on the mean and variance estimates. In particular, we can use Bernstein's inequality on the mean (see e.g.,~\cite{audibert2009exploration-exploitation}) and a tighter deviation on the variance~\cite{maurer2009empirical}, obtaining the index\footnote{We notice that in this case the estimated variance is computed as $\hvar_{i,s} = \frac{1}{s-1} \sum_{s'=1}^{s}X_{i,s'}^2-\hmu_{i,s}^2$}
%%
%\begin{align}
%B^V_{i, s, t} &= \Big( \hat\sigma_{i,s} + \sqrt{\frac{\log 1/\delta}{2s}} \Big)^2  \\
%&\quad-\rho \Big(\hmu_{i,s} + \hat\sigma_{i,s} \sqrt{\frac{\log 1/\delta}{s}} + \frac{\log 1/\delta}{s}\Big) \nonumber.
%\end{align}
%%
%While this version of \textsl{MV-LCB} should work better whenever the variance of the arms is small, its theoretical properties would not differ much w.r.t. \textsl{MV-LCB} (see \cite{audibert2009exploration-exploitation} for a comparison between \textsl{UCB-V} and \textsl{UCB}).

\remark{2} The \textsl{MV-LCB} algorithm is parameterized by a parameter $\delta$ which defines the confidence level of the bounds employed in the definition of the index (\ref{eq:index}). In Theorem~\ref{thm:mvlcb.regret} we show how to optimize the parameter when the horizon $n$ is known in advance. On the other hand, if $n$ is not known, it is possible to design an anytime version of \textsl{MV-LCB} by defining a non-decreasing exploration sequence $(\eps_t)_t$ instead of the term $\log 1/\delta$.

%% Theoretical Analysis

\subsection{Theoretical Analysis}\label{ss:theory}

In this section we report the analysis of the regret $\R_n(\A)$ of \textsl{MV-LCB} (Fig.~\ref{f:lcb}). As highlighted in eq.~\ref{eq:regret3}, it is enough to analyze the number of pulls for each of the arms to recover a bound on the regret. The proofs (reported in the appendix) are mostly based on similar arguments to the proof of \textsl{UCB}.

%We first report a high--probability bound on the number of pulls. 
% and that the high--probability event over which the statement holds coincides with the event form which Lemma~\ref{l:psuedo.regret} which thus allows us to combine the two result to obtain a high--probability bound for the true regret $\R_n(\A)$.

%\begin{lemma}\label{l:expected}
%Let $b = 2(5+\rho)$, for any $\delta \in (0,1)$, the number of times each suboptimal arm $i\neq i^*$ is pulled by \textsl{MV-LCB} is
%%
%\begin{align}\label{eq:mvlcb.pulls.hp}
%T_{i,n} \leq \frac{b^2}{\Delta_i^2} \log \frac{1}{\delta} + 1,
%\end{align}
%%
%with probability at least $1-6nK\delta$.
%\end{lemma}

%From the previous result, 

We derive the following regret bound in high probability and expectation.

%\TODO{Report the corresponding high-prob and expectation bounds for the true regret in the supplementary material}

\begin{theorem}\label{thm:mvlcb.regret}
Let the optimal arm $i^*$ be unique and $b = 2(5+\rho)$, the \textsl{MV-LCB} algorithm achieves a pseudo--regret bounded as
\begin{align*}
\tR_n(\A) &\leq \frac{b^2 \log 1/\delta}{n}\bigg(\sum_{i\neq i^*}\frac{1}{\Delta_i} + 4\sum_{i\neq i^*}\frac{\Gamma_{i^*,i}^2}{\Delta_i^2} + \frac{2b^2 \log 1/\delta}{n}\sum_{i\neq i^*}\mathop{\sum_{j\neq i}}_{j\neq i^*}\frac{\Gamma_{i,j}^2}{\Delta_i^2 \Delta_j^2}\bigg) + \frac{5K}{n},
\end{align*}
with probability at least $1-6nK\delta$. Similarly, if \textsl{MV-LCB} is run with $\delta = 1/n^2$ then
\begin{align*}
\E[\tR_n(\A)] &\leq \frac{2b^2 \log n}{n}\bigg(\sum_{i\neq i^*}\frac{1}{\Delta_i} + 4\sum_{i\neq i^*}\frac{\Gamma_{i^*,i}^2}{\Delta_i^2} + \frac{4b^2 \log n}{n}\sum_{i\neq i^*}\mathop{\sum_{j\neq i}}_{j\neq i^*}\frac{\Gamma_{i,j}^2}{\Delta_i^2 \Delta_j^2}\bigg) + (17 + 6\rho)\frac{K}{n}.
\end{align*}
\end{theorem}

\remark{1 (the bound)} Let $\Delta_{\min} = \min_{i\neq i^*} \Delta_i$ and $\Gamma_{\max} = \max_{i} |\Gamma_i|$, then a rough simplification of the previous bound leads to
\begin{align*}
\E[\tR_n(\A)]\leq O\Big(\frac{K}{\Delta_{\min}} \frac{\log n}{n} + K^2 \frac{\Gamma_{\max}^2}{\Delta_{\min}^4} \frac{\log^2 n}{n}\Big).
\end{align*}
First we notice that the regret decreases as $O(\log^2 n/n)$, implying that \textsl{MV-LCB} is a consistent algorithm. As already highlighted in Definition~\ref{d:regret}, the regret is mainly composed by two terms. The first term is due to the difference in the mean--variance of the best arm and the arms pulled by the algorithm, while the second term denotes the additional variance introduced by the exploration risk of pulling arms with different means. In particular, it is interesting to note that this additional term depends on the squared difference in the means of the arms $\Gamma_{i,j}^2$. Thus, if all the arms have the same mean, this term would be zero.

%\remark{xxx (comparison with \textsl{UCB})} Although targeting different problems, it is interesting to compare \textsl{MV-LCB} to \textsl{UCB} and their regret. We already noticed that as $\tau$ tends to infinity \textsl{MV-LCB} reduces to \textsl{UCB}. Similarly, in that case the regret bound indeed coincides with the average cumulative regret bound derived for \textsl{UCB}.

\remark{2 (worst--case analysis)} We can further study the result of Theorem~\ref{thm:mvlcb.regret} by considering the worst--case performance of \textsl{MV-LCB}, that is the performance when the distributions of the arms are chosen so as to maximize the regret. In order to illustrate our argument we consider the simple case of $K=2$ arms, $\rho=0$ (variance minimization), $\mu_1 \neq \mu_2$, and $\var_1 = \var_2 = 0$ (deterministic arms).~\footnote{Note that in this case (i.e., $\Delta=0$), Theorem~\ref{thm:mvlcb.regret} does not hold, since the optimal arm is not unique.} In this case we have a variance gap $\Delta = 0$ and $\Gamma^2 > 0$. According to the definition of \textsl{MV-LCB}, the index $B_{i,s}$ would simply reduce to $B_{i,s} = \sqrt{\log(1/\delta)/s}$,
%%
%\begin{align*}
%B_{i,s} = \sqrt{\frac{\log 1/\delta}{s}},
%\end{align*}
%%
thus forcing the algorithm to pull both arms uniformly (i.e., $T_{1,n}=T_{2,n} = n/2$ up to rounding effects). Since the arms have the same variance, there is no direct regret in pulling either one or the other. Nonetheless, the algorithm has an additional variance due to the difference in the samples drawn from distributions with different means. In this case, the algorithm suffers a constant (true) regret
\begin{align*}
\R_n(\textsl{MV-LCB}) = 0 + \frac{T_{1,n}T_{2,n}}{n^2} \Gamma^2 = \frac{1}{4} \Gamma^2,
\end{align*}
independent from the number of rounds $n$. This argument can be generalized to multiple arms and $\rho\neq 0$, since it is always possible to design an environment (i.e., a set of distributions) such that $\Delta_{\min}=0$ and $\Gamma_{\max}\neq 0$.~\footnote{Notice that this is always possible for a large majority of distributions for which the mean and variance are independent or mildly correlated.} This result is not surprising. In fact, two arms with the same mean--variance are likely to produce similar observations, thus leading \textsl{MV-LCB} to pull the two arms repeatedly over time, since the algorithm is designed to try to discriminate between similar arms. Although this behavior does not suffer from any regret in pulling the ``suboptimal'' arm (the two arms are equivalent), it does introduce an additional variance, due to the difference in the means of the arms ($\Gamma\neq 0$), which finally leads to a regret the algorithm is not ``aware'' of. This argument suggests that, for any $n$, it is always possible to design an environment for which \textsl{MV-LCB} has a constant regret. This is particularly interesting since it reveals a huge gap between the mean--variance problem and the standard expected regret minimization problem and will be further investigated in the numerical simulations presented in Section~\ref{s:simulations}. In fact, in the latter case, \textsl{UCB} is known to have a worst--case regret per round of $\Omega(1/\sqrt{n})$~\cite{audibert2010regret}, while in the worst case, \textsl{MV-LCB} suffers a constant regret. In the next section we introduce a simple algorithm able to deal with this problem and achieve a vanishing worst--case regret.

%%%%%%%%%%%%%%%%%%%%%%%%%%%%%%%%%%%%%%%%%%%%%%%%%%%%%%%%%%%%%%%%%%%%%%%%%%%%%%%
%% The Mean--Variance Lower Confidence Bound Algorithm
%%%%%%%%%%%%%%%%%%%%%%%%%%%%%%%%%%%%%%%%%%%%%%%%%%%%%%%%%%%%%%%%%%%%%%%%%%%%%%%

%\vspace{-0.1in}
\section{The Exploration--Exploitation Algorithm}\label{s:ee}
%\vspace{-0.1in}

%Although for any fixed problem (with $\Delta_{\min}> 0$) the \textsl{MV-LCB} algorithm has a vanishing regret, for any value of $n$, it is always possible to find an environment for which its regret is constant. In this section, we analyze a simple algorithm where exploration and exploitation are two distinct phases.
%
%% The algorithm
%
%\subsection{The Algorithm}
%
%\begin{figure}[t]
%\bookbox{
%\begin{algorithmic}
%\STATE \textbf{Input:} length of the exploration phase $\tau$
%\STATE \textit{Exploration phase}
%\FOR{$t = 1,\ldots,\tau/K$}
%\FOR{$i = 1,\ldots,K$}
%\STATE Observe $X_{i, t} \sim \nu_{i}$
%\ENDFOR
%\ENDFOR
%\STATE \textit{Exploitation phase}
%\STATE Compute the estimates $\hMV_{i,\tau/K}$
%\STATE Compute $\hat i^* = \arg\min_i \hMV_{i,\tau/K}$
%\FOR{$t = \tau+1,\ldots,n$}
%\STATE Select $\hat i^*$
%\ENDFOR
%\end{algorithmic}}
%\vspace{-0.1in}
%\caption{Pseudo-code of the \textsl{ExpExp} algorithm.}\label{f:ee}
%\vspace{-0.15in}
%\end{figure}
%
%As shown in Figure~\ref{f:ee}, 
The \textsl{ExpExp} algorithm divides the time horizon $n$ into two distinct phases of length $\tau$ and $n-\tau$ respectively. During the first phase all the arms are explored uniformly, thus collecting $\tau/K$ samples each~\footnote{In the definition and in the following analysis we ignore rounding effects.}. Once the exploration phase is over, the mean--variance of each arm is computed and the arm with the smallest estimated mean--variance $\MV_{i,\tau/K}$ is repeatedly pulled until the end.

%% The Theoretical Analysis

%\subsection{Theoretical Analysis}

The \textsl{MV-LCB} is specifically designed to minimize the probability of pulling the wrong arms, so whenever there are two equivalent arms (i.e., arms with the same mean--variance), the algorithm tends to pull them the same number of times, at the cost of potentially introducing an additional variance which might result in a constant regret. On the other hand, \textsl{ExpExp} stops exploring the arms after $\tau$ rounds and then elicits one arm as the best and keeps pulling it for the remaining $n-\tau$ rounds. Intuitively, the parameter $\tau$ should be tuned so as to meet different requirements. The first part of the regret (i.e., the regret coming from pulling the suboptimal arms) suggests that the exploration phase $\tau$ should be long enough for the algorithm to select the empirically best arm $\hat i^*$ at $\tau$ equivalent to the actual optimal arm $i^*$ with high probability; and at the same time, as short as possible to reduce the number of times the suboptimal arms are explored. On the other hand, the second part of the regret (i.e., the variance of pulling arms with different means) is minimized by taking $\tau$ as small as possible (e.g., $\tau=0$ would guarantee a zero regret). The following theorem illustrates the optimal trade-off between these contrasting needs.

\begin{theorem}\label{thm:ee.regret}
Let \textsl{ExpExp} be run with $\tau = K(n/14)^{2/3}$, then for any choice of distributions $\{\nu_i\}$ the expected regret is $\E[\tR_n(\A)] \leq 2\frac{K}{n^{1/3}}.$
%
%\begin{align}
%\E[\tR_n(\A)] \leq 2\frac{K}{n^{1/3}}.
%\end{align}
%%
\end{theorem}

\remark{1 (the bound)} We first notice that this bound suggests that \textsl{ExpExp} performs worse than \textsl{MV-LCB} on easy problems. In fact, Theorem~\ref{thm:mvlcb.regret} demonstrates that \textsl{MV-LCB} has a regret decreasing as $O(K\log(n)/n)$ whenever the gaps $\Delta$ are not small compared to $n$, while in the remarks of Theorem~\ref{thm:mvlcb.regret} we highlighted the fact that for any value of $n$, it is always possible to design an environment which leads \textsl{MV-LCB} to suffer a constant regret. On the other hand, the previous bound for \textsl{ExpExp} is distribution independent and indicates the regret is still a decreasing function of $n$ even in the worst case. This opens the question whether it is possible to design an algorithm which works as well as \textsl{MV-LCB} on easy problems and as robustly as \textsl{ExpExp} on difficult problems.

\remark{2 (exploration phase)} The previous result can be improved by changing the exploration strategy used in the first $\tau$ rounds. Instead of a pure uniform exploration of all the arms, we could adopt a best--arm identification algorithms such as \textsl{Successive Reject} or \textsl{UCB-E}, which maximize the probability of returning the best arm given a fixed budget of rounds $\tau$ (see e.g.,~\cite{audibert2010best}).

%%%%%%%%%%%%%%%%%%%%%%%%%%%%%%%%%%%%%%%%%%%%%%%%%%%%%%%%%%%%%%%%%%%%%%%%%%%%%%%
%% Numerical Simulations
%%%%%%%%%%%%%%%%%%%%%%%%%%%%%%%%%%%%%%%%%%%%%%%%%%%%%%%%%%%%%%%%%%%%%%%%%%%%%%%

\vspace{-0.1in}
\section{Numerical Simulations}\label{s:simulations}
\vspace{-0.1in}

\begin{figure*}[t]
\begin{center}
\includegraphics[width=0.38\textwidth]{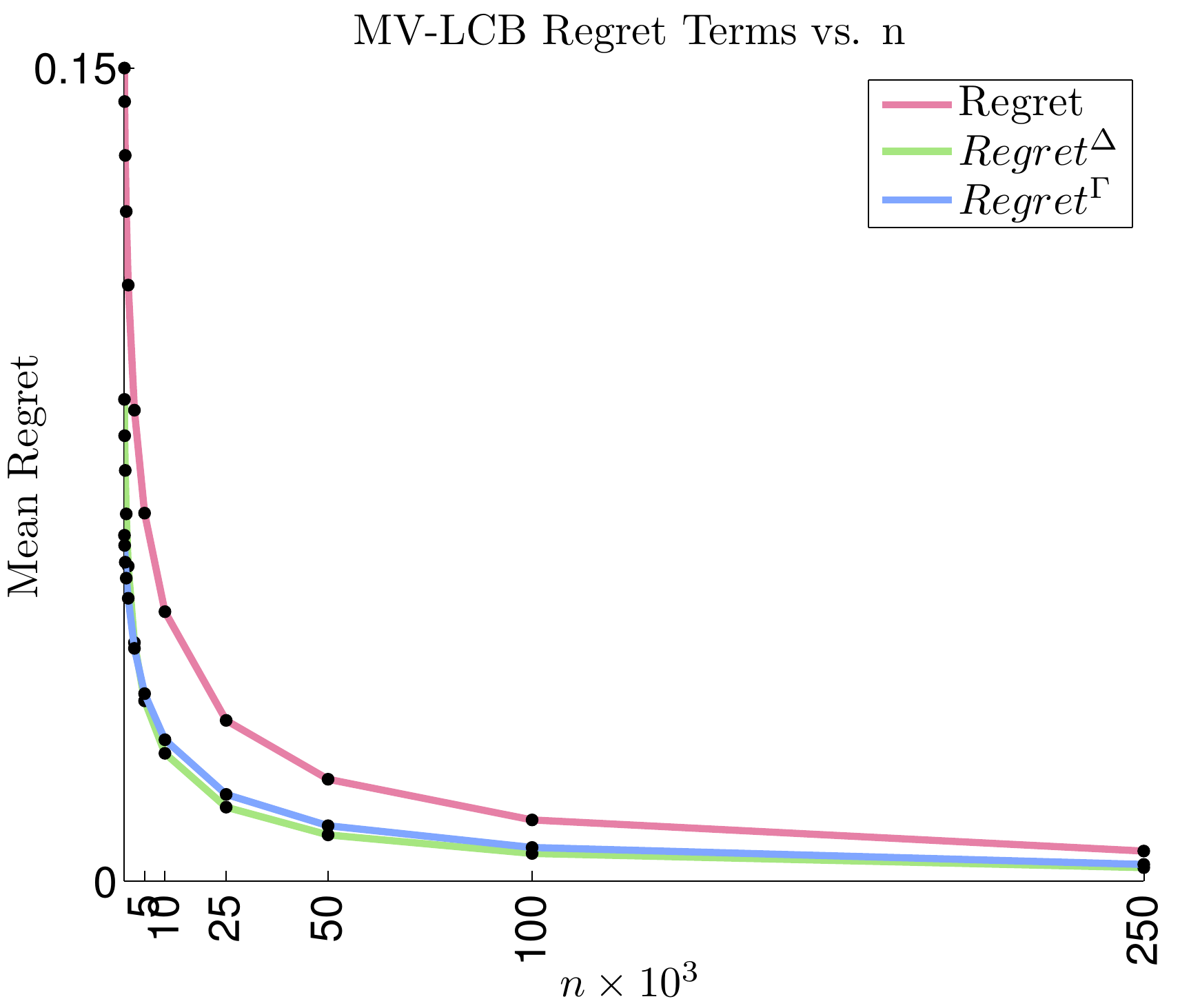}
\hspace{0.5in}
\includegraphics[width=0.38\textwidth]{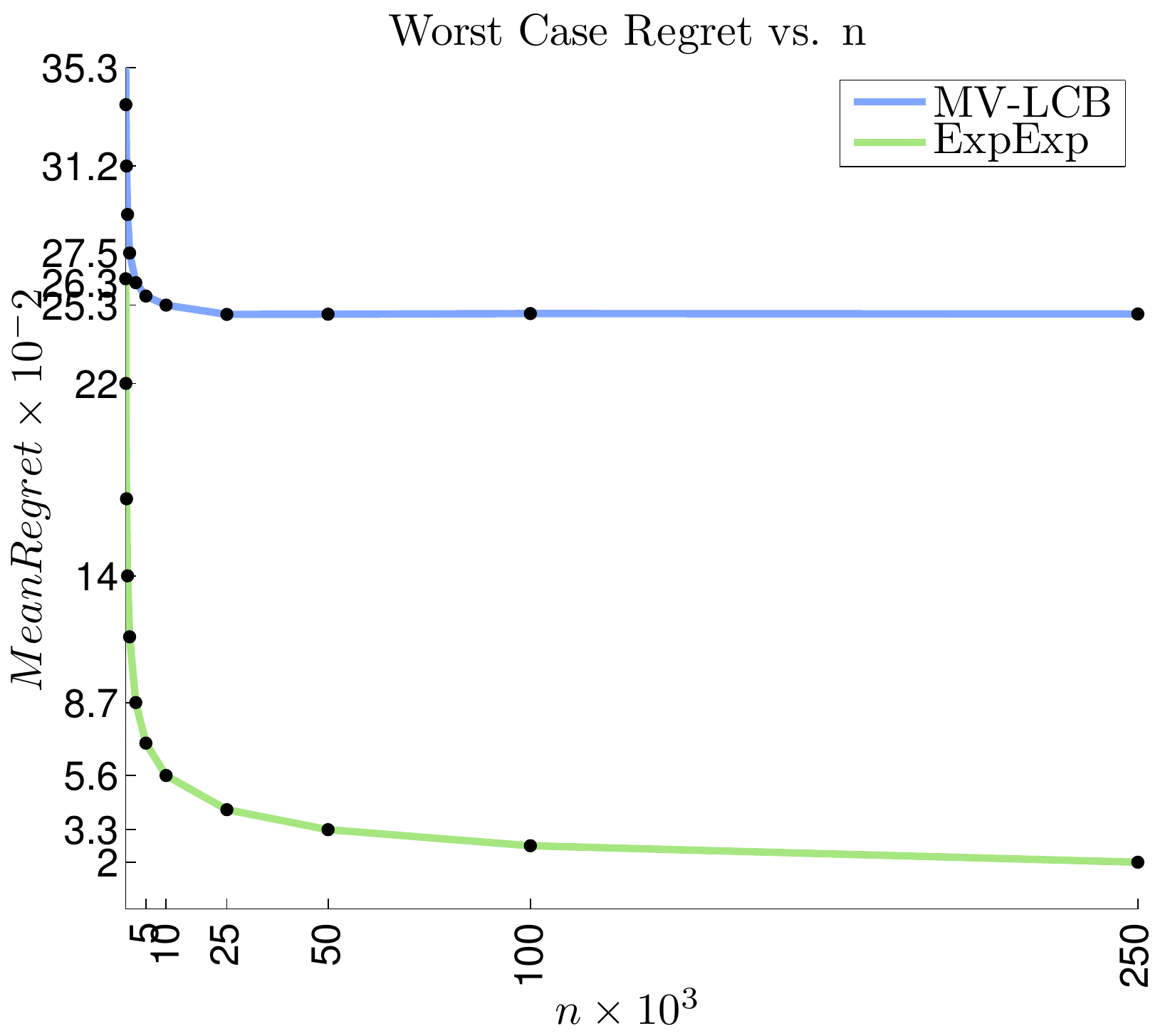}
\end{center}
\vspace{-0.4cm}
\caption{Regret of \textsl{MV-LCB} and \textsl{ExpExp} in different scenarios.}\label{f:mvlcb}
\vspace{-0.4cm}
\end{figure*}

In this section we report numerical simulations aimed at validating the main theoretical findings reported in the previous sections. In the following graphs we study the true regret $\R_n(\A)$ averaged over 500 runs.
We first consider the variance minimization problem ($\rho=0$) with $K=2$ Gaussian arms set to $\mu_1 = 1.0$, $\mu_2=0.5$, $\var_1=0.05$, and $\var_2=0.25$ and run \textsl{MV-LCB}~\footnote{Notice that although in the paper we assumed the distributions to be bounded in $[0,1]$ all the results can be extended to sub-Gaussian distributions.}. In Figure~\ref{f:mvlcb} we report the true regret $\R_n$ (as in the original definition in eq.~\ref{eq:regret}) and its two components $\R_n^{\hDelta}$ and $\R_n^{\hGamma}$ (these two values are defined as in eq.~\ref{eq:regrets} with $\hDelta$ and $\hGamma$ replacing $\Delta$ and $\Gamma$). As expected (see e.g., Theorem~\ref{thm:mvlcb.regret}), the regret is characterized by the regret realized from pulling suboptimal arms and arms with different means (Exploration Risk) and tends to zero as $n$ increases. Indeed, if we considered two distributions with equal means ($\mu_1=\mu_2$), the average regret coincides with $\R_n^{\hDelta}$. Furthermore, as shown in Theorem~\ref{thm:mvlcb.regret} the two regret terms decrease with the same rate $O(\log n/n)$.

A detailed analysis of the impact of $\Delta$ and $\Gamma$ on the performance of \textsl{MV-LCB} is reported in Appendix~\ref{app:simulations}. Here we only compare the worst--case performance of \textsl{MV-LCB} to \textsl{ExpExp} (see Figure~\ref{f:mvlcb}). In order to have a fair comparison, for any value of $n$ and for each of the two algorithms, we select the pair $\Delta_{w}, \Gamma_{w}$ which corresponds to the largest regret (we search in a grid of values with $\mu_1=1.5$, $\mu_2\in [0.4; 1.5]$, $\var_1\in [0.0; 0.25]$, and $\var_2 = 0.25$, so that $\Delta\in [0.0;0.25]$ and $\Gamma\in [0.0; 1.1]$). As discussed in Section~\ref{s:ee}, while the worst--case regret of \textsl{ExpExp} keeps decreasing over $n$, it is always possible to find a problem for which regret of \textsl{MV-LCB} stabilizes to a constant. For numerical results with multiple values of $\rho$ and 15 arms, please see Appendix~\ref{app:simulations}.

%%%%%%%%%%%%%%%%%%%%%%%%%%%%%%%%%%%%%%%%%%%%%%%%%%%%%%%%%%%%%%%%%%%%%%%%%%%%%%%
%% Discussion
%%%%%%%%%%%%%%%%%%%%%%%%%%%%%%%%%%%%%%%%%%%%%%%%%%%%%%%%%%%%%%%%%%%%%%%%%%%%%%%

\vspace{-0.1in}
\section{Discussion}\label{s:discussion}
\vspace{-0.1in}

In this paper we evaluate the \textit{risk} of an algorithm in terms of the variability of the sequences of samples that it actually generates. Although this notion might resemble other analyses of UCB-based algorithms (see e.g., the high-probability analysis in~\cite{audibert2009exploration-exploitation}), it captures different features of the learning algorithm. Whenever a bandit algorithm is run over $n$ rounds, its behavior, combined with the arms' distributions, generates a probability distribution over sequences of $n$ rewards. While the \textit{quality} of this sequence is usually defined by its cumulative sum (or average), here we say that a sequence of rewards is \textit{good} if it displays a good trade-off between its (empirical) mean and variance. It is important to notice that this notion of risk-return tradeoff does not coincide with the variance of the algorithm over multiple runs.

Let us consider a simple case with two arms that deterministically generate 0s and 1s respectively, and two different algorithms. Algorithm $\A_1$ pulls the arms in a fixed sequence at each run (e.g., arm 1, arm 2, arm 1, arm 2, and so on), so that each arm is always pulled $n/2$ times. Algorithm $\A_2$ chooses one arm uniformly at random at the beginning of the run and repeatedly pulls this arm for $n$ rounds. Algorithm $\A_1$ generates sequences such as $010101...$ which have high variability within each run, incurs a high regret (e.g., if $\rho=0$), but has no variance over multiple runs because it always generates the same sequence. On the other hand, $\A_2$ has no variability in each run, since it generates sequences with only 0s or only 1s, suffers no regret in the case of variance minimization, but has high variance over multiple runs since the two completely different sequences are generated with equal probability. This simple example demonstrates that an algorithm with a very small standard regret w.r.t. the cumulative reward (e.g., $\A_1$), might result in a very high variability in a single run of the algorithm, while an algorithm with small mean-variance regret (e.g., $\A_2$) could have a high variance over multiple runs.

%%%%%%%%%%%%%%%%%%%%%%%%%%%%%%%%%%%%%%%%%%%%%%%%%%%%%%%%%%%%%%%%%%%%%%%%%%%%%%%
%% Conclusions
%%%%%%%%%%%%%%%%%%%%%%%%%%%%%%%%%%%%%%%%%%%%%%%%%%%%%%%%%%%%%%%%%%%%%%%%%%%%%%%

\vspace{-0.1in}
\section{Conclusions}\label{s:conclusions}
\vspace{-0.1in}

The majority of multi--armed bandit literature focuses on the problem of minimizing the regret w.r.t. the arm with the highest return in expectation. We study the notion of risk associated to the variance over multiple runs and risk of variability associated to a single run of an algorithm. The later case highlights an interesting effect on the regret due to the need to estimate variability within a single sequence of finite random samples before making a risk-averse decision. Further, controling the variance risk over multiple runs does not necessarily control the risk of variability over a single run. In this paper, we introduced a novel multi--armed bandit setting where the objective is to perform as well as the arm with the best risk--return trade--off. In particular, we relied on the mean--variance model introduced in~\cite{markowitz1952portfolio} to measure the performance of the arms and define the regret of a learning algorithm. We proposed two novel algorithms to solve the mean--variance bandit problem and we reported their corresponding theoretical analysis. While \textsl{MV-LCB} shows a small regret of order $O(\log n/n)$ on ``easy'' problems (i.e., where the mean--variance gaps $\Delta$ are big w.r.t. $n$), we showed that it has a constant worst--case regret. On the other hand, we proved that \textsl{ExpExp} has a vanishing worst--case regret at the cost of worse performance on ``easy'' problems. To the best of our knowledge this is the first work introducing risk--aversion in the multi--armed bandit setting and it opens a series of interesting questions.

%% Extensions

%\subsection{Extensions}

\textbf{Lower bound.} In this paper we introduced two algorithms, \textsl{MV-LCB} and \textsl{ExpExp}. As discussed in the remarks of Theorem~\ref{thm:mvlcb.regret} and Theorem~\ref{thm:ee.regret}, \textsl{MV-LCB} has a regret of order $O(\sqrt{K/n})$ on easy problems and $O(1)$ on difficult problems, while \textsl{ExpExp} achieves the same regret $O(K/n^{1/3})$ over all problems. The primary open question is whether $O(K/n^{1/3})$ is actually the best possible achievable rate (in the worst--case) for this problem or a better rate is possible. This question is of particular interest since the standard reward expectation maximization problem has a known lower--bound of $\Omega(\sqrt{1/n})$, and a minimax rate of $\Omega(1/n^{1/3})$ for the mean--variance problem would imply that the risk--averse bandit problem is intrinsically more difficult than standard bandit problems.

\textbf{Different measures of return--risk.} Considering alternative notions of risk is a straightforward extension to the previous setting. In fact, over the years the mean--variance model has often been criticized. From a point of view of the expected utility theory, the mean--variance model is only justified under a Gaussianity assumption on the arm distributions. It also violates the monotonocity condition due to the different orders of the mean and variance and is not a coherent measure of risk \cite{Artzner1999}. 
Furthermore, the variance is a symmetric measure of risk, while it is often the case that only one--sided deviations from the mean are undesirable (e.g., in finance only losses w.r.t. to the expected return are considered as a risk, while any positive deviation is not considered as a real risk). A popular replacement for the mean--variance is to use the $\alpha$ value--at--risk (i.e., the quantile) to measure the risk of a random variable. The main challenge in this case is the estimation of the value--at--risk for each arm. In fact, while the cumulative distribution of a random variable can be reliably estimated (see e.g.,~\cite{massart1990the-tight}), estimating the quantile might be more difficult.

In \cite{Artzner1999} axiomatic rules are listed to define coherent measures of risk. Though $\alpha$ value--at--risk violates these rules, Conditional Value at Risk (otherwise known as average value at risk, tail value at risk, expected shortfall and lower tail risk) passes these rules as a coherent measure of risk. One can easily imagine a lower confidence bound algorithm based on \cite{Brown2007} in the same composition as \textsl{MV-LCB} which replaces the variance by the conditional value at risk.

The notion of optimality in the risk sensitive setting also depends on the selection of a single-period or multi-period risk evaluation. While the single-period risk of an arm is simply the risk of its distribution, in a multi-period evaluation we consider the risk of the sum of rewards obtained by repeatedly pulling the same arm over $n$ rounds. Unlike the variance, for which the variance of a sum of $n$ independent realizations of the same random variable is simply $n$ times its variance, for other measures of risk (e.g., $\alpha$ value--at--risk) this is not necessarily the case. As a result, an arm with the smallest single-period risk might not be the optimal choice over an horizon of $n$ rounds. Therefore, the performance of a learning algorithm should be compared to the smallest risk that can be achieved by any sequence of arms over $n$ rounds, thus requiring a new definition of regret.

%A single period evaluation corresponds to is similar to choosing the best arm in hindsight, while in the multi-period case optimality does not necessarily coincide with the best single arm in hindsight. It may correspond to the best sequence of arms. In the mean variance setting, the best sequence of arms corresponds to the best single arm, so the single and multi-period cases coincide. This is not necessarily the case for other notions of risk, such as the conditional value at risk or value at risk, where the optimal single period risk may coincide with a single arm, but the multi-period risk is defined by the minimum risk over the best sequence of arms.

\textbf{Linear bandits.} In linear bandits, each arm is characterized by a marginal distribution with expected value $\mu_i$ and a covariance matrix $C$. At each step the learner chooses a combination of arms and observes the corresponding combined reward. In this case, the best combination is obtained by solving the mean--variance quadratic program $\min_{\bx} (\bx^\top C \bx - \rho\bx^\top \bmu)$ where $\bx$ is usually a point in the $K$-dimensional simplex (e.g., in finance $\bx$ is in the simplex when no short--selling is allowed). Similar to the multi--arm case, the objective is to define an algorithm able to achieve a mean--variance as small as the best point in the simplex over $n$ rounds.

\textbf{Simple regret.} Finally, an interesting related problem is the simple regret setting where the learner is allowed to explore over $n$ rounds and it only suffers a regret defined on the solution returned at the end. It is known that it is possible to design algorithm able to effectively estimate the mean of the arms and finally return the best arm with high probability. In the risk-return setting, the objective would be to return the arm with the best risk-return tradeoff.

{\bf Acknowledgments} This work was supported by Ministry of Higher Education and Research, Nord-Pas de Calais Regional Council and FEDER through the ``contrat de projets {\'e}tat region 2007--2013", French National Research Agency (ANR) under project LAMPADA $n^\circ$ ANR-09-EMER-007, European Community's Seventh Framework Programme (FP7/2007-2013) under grant agreement $n^\circ$ 270327, and PASCAL2 European Network of Excellence.

\bibliographystyle{plain}
\bibliography{risk_bandit}	

\begin{thebibliography}{10}

\bibitem{antos2010active}
Andr\'{a}s Antos, Varun Grover, and Csaba Szepesv\'{a}ri.
\newblock Active learning in heteroscedastic noise.
\newblock {\em Theoretical Computer Science}, 411:2712--2728, June 2010.

\bibitem{Artzner1999}
P~Artzner, F~Delbaen, JM~Eber, and D~Heath.
\newblock {Coherent measures of risk}.
\newblock {\em Mathematical finance}, (June 1996):1--24, 1999.

\bibitem{audibert2010regret}
Jean-Yves Audibert and S{\'e}bastien Bubeck.
\newblock Regret bounds and minimax policies under partial monitoring.
\newblock {\em Journal of Machine Learning Research}, 11:2785--2836, 2010.

\bibitem{audibert2010best}
Jean-Yves Audibert, S{\'e}bastien Bubeck, and R{\'e}mi Munos.
\newblock Best arm identification in multi-armed bandits.
\newblock In {\em Proceedings of the Twenty-third Conference on Learning Theory
  (COLT'10)}, 2010.

\bibitem{audibert2009exploration-exploitation}
Jean-Yves Audibert, R{\'e}mi Munos, and Csaba Szepesv\'{a}ri.
\newblock Exploration-exploitation trade-off using variance estimates in
  multi-armed bandits.
\newblock {\em Theoretical Computer Science}, 410:1876--1902, 2009.

\bibitem{auer2002finite-time}
Peter Auer, Nicol\`{o} Cesa-Bianchi, and Paul Fischer.
\newblock Finite-time analysis of the multi-armed bandit problem.
\newblock {\em Machine Learning}, 47:235--256, 2002.

\bibitem{Brown2007}
David~B. Brown.
\newblock Large deviations bounds for estimating conditional value-at-risk.
\newblock {\em Operations Research Letters}, 35:722--730, 2007.

\bibitem{even-dar2006risk-sensitive}
Eyal Even-Dar, Michael Kearns, and Jennifer Wortman.
\newblock Risk-sensitive online learning.
\newblock In {\em Proceedings of the 17th international conference on
  Algorithmic Learning Theory (ALT'06)}, pages 199--213, 2006.

\bibitem{gollier2001the-economics}
Christian Gollier.
\newblock {\em The Economics of Risk and Time}.
\newblock {The MIT Press}, 2001.

\bibitem{markowitz1952portfolio}
Harry Markowitz.
\newblock Portfolio selection.
\newblock {\em The Journal of Finance}, 7(1):77--91, 1952.

\bibitem{massart1990the-tight}
Pascal Massart.
\newblock The tight constant in the dvoretzky-kiefer-wolfowitz inequality.
\newblock {\em The Annals of Probability}, 18(3):pp. 1269--1283, 1990.

\bibitem{Neumann1947}
J~Neumann and O~Morgenstern.
\newblock {Theory of games and economic behavior}.
\newblock {\em Princeton University, Princeton}, 1947.

\bibitem{robbins1952some}
Herbert Robbins.
\newblock Some aspects of the sequential design of experiments.
\newblock {\em Bulletin of the AMS}, 58:527--535, 1952.

\bibitem{salomon2011deviations}
Antoine Salomon and Jean-Yves Audibert.
\newblock Deviations of stochastic bandit regret.
\newblock In {\em Proceedings of the 22nd international conference on
  Algorithmic learning theory (ALT'11)}, pages 159--173, 2011.

\bibitem{warmuth2006online}
Manfred~K. Warmuth and Dima Kuzmin.
\newblock Online variance minimization.
\newblock In {\em Proceedings of the 19th Annual Conference on Learning Theory
  (COLT'06)}, pages 514--528, 2006.

\end{thebibliography}

\newpage
\appendix

%%%%%%%%%%%%%%%%%%%%%%%%%%%%%%%%%%%%%%%%%%%%%%%%%%%%%%%%%%%%%%%%%%%%%%%
%REGRET
%%%%%%%%%%%%%%%%%%%%%%%%%%%%%%%%%%%%%%%%%%%%%%%%%%%%%%%%%%%%%%%%%%%%%%%

\section{The Regret}\label{app:regret}

%% Definition of the regret

\subsection{The True Regret}\label{app:true.regret}

We recall the definition of the (empirical) regret as
\begin{align*}
\R_n(\A) &=  \hMV_n(\A) - \hMV_{i^*,n}.
\end{align*}

Given the definitions reported in the main paper, we first elaborate on the two mean terms in the regret as
\begin{align*}
\hmu_{i^*,n} = \frac{1}{n}\sum_{i=1}^K \sum_{t=1}^{T_{i,n}} Y_{i,t} = \frac{1}{n}\sum_{i=1}^K T_{i,n} \tmu_{i,T_{i,n}},
\end{align*}
and
\begin{align*}
\hmu_{n}(\A) = \frac{1}{n}\sum_{i=1}^K \sum_{t=1}^{T_{i,n}} X_{i,t} = \frac{1}{n}\sum_{i=1}^K T_{i,n} \hmu_{i,T_{i,n}}.
\end{align*}

Similarly, the two variance terms can be written as
\begin{align*}
\hvar_n(\A) &= \frac{1}{n} \sum_{i=1}^K \sum_{t=1}^{T_{i,n}} \big(X_{i,t} - \hmu_n(\A)\big)^2 \\
&= \frac{1}{n}\sum_{i=1}^K\sum_{t=1}^{T_{i,n}} \big(X_{i,t} - \hmu_{i,T_{i,n}}\big)^2 + \frac{1}{n}\sum_{i=1}^K\sum_{t=1}^{T_{i,n}} \big(\hmu_{i,T_{i,n}} - \hmu_n(\A)\big)^2 + \frac{2}{n}\sum_{i=1}^K \sum_{t=1}^{T_{i,n}} \big(X_{i,t} - \hmu_{i,T_{i,n}}\big)\big(\hmu_{i,T_{i,n}} - \hmu_n(\A)\big) \\
&= \frac{1}{n}\sum_{i=1}^K T_{i,n} \hvar_{i,T_{i,n}} + \frac{1}{n}\sum_{i=1}^K T_{i,n} \big(\hmu_{i,T_{i,n}} - \hmu_n(\A)\big)^2 + 0,
\end{align*}
and
\begin{align*}
\var_{i^*,n} &= \frac{1}{n} \sum_{i=1}^K \sum_{t=1}^{T_{i,n}} \big(Y_{i,t} - \hmu_{i^*,n}\big)^2 \\
&= \frac{1}{n}\sum_{i=1}^K \sum_{t=1}^{T_{i,n}} \big(Y_{i,t} - \tmu_{i,T_{i,n}}\big)^2 + \frac{1}{n}\sum_{i=1}^K \sum_{t=1}^{T_{i,n}} \big(\tmu_{i,T_{i,n}} - \hmu_{i^*,n}\big)^2 + \frac{2}{n}\sum_{i=1}^K \sum_{t=1}^{T_{i,n}} \big(Y_{i,t} - \tmu_{i,T_{i,n}}\big)\big(\tmu_{i,T_{i,n}} - \hmu_{i^*,n}\big)\\
&= \frac{1}{n}\sum_{i=1}^K T_{i,t} \tvar_{i,T_{i,n}} + \frac{1}{n}\sum_{i=1}^K T_{i,n} \big(\tmu_{i,T_{i,n}} - \hmu_{i^*,n}\big)^2 + 0.
\end{align*}

Putting together these terms, we obtain the regret (see eq.~\ref{eq:regret})
\begin{align*}
\R_n(\A) &= \frac{1}{n}\sum_{i\neq i^*} T_{i,n} \Big[(\hvar_{i,T_{i,n}}-\tvar_{i,T_{i,n}}) - \rho(\hmu_{i,T_{i,n}} - \tmu_{i,T_{i,n}})\Big] \\
&\quad+ \frac{1}{n}\sum_{i=1}^K T_{i,n} \big(\hmu_{i,T_{i,n}} - \hmu_n(\A)\big)^2 - \frac{1}{n}\sum_{i=1}^K T_{i,n} \big(\tmu_{i,T_{i,n}} - \hmu_{i^*,n}\big)^2
\end{align*}

If we further elaborate the second term, we obtain
\begin{align*}
\frac{1}{n}\sum_{i=1}^K T_{i,n} \big(\hmu_{i,T_{i,n}} - \hmu_n(\A)\big)^2 &= \frac{1}{n} \sum_{i=1}^K T_{i,n} \Big(\hmu_{i,T_{i,n}} - \frac{1}{n} \sum_{j=1}^K T_{j,n} \hmu_{j,T_{j,n}} \Big)^2 \\
&= \frac{1}{n} \sum_{i=1}^K T_{i,n} \Big(\sum_{j=1}^K \frac{T_{j,n}}{n} (\hmu_{i,T_{i,n}} -  \hmu_{j,T_{j,n}}) \Big)^2 \\
&\leq \frac{1}{n} \sum_{i=1}^K T_{i,n} \sum_{j=1}^K \frac{T_{j,n}}{n} (\hmu_{i,T_{i,n}} -  \hmu_{j,T_{j,n}})^2 \\
&= \frac{1}{n^2} \sum_{i=1}^K \sum_{j\neq i} T_{i,n}  T_{j,n} (\hmu_{i,T_{i,n}} -  \hmu_{j,T_{j,n}})^2.
\end{align*}
Using the definitions $\hDelta_i = (\hvar_{i,T_{i,n}}-\tvar_{i,T_{i,n}}) - \rho(\hmu_{i,T_{i,n}} - \tmu_{i,T_{i,n}})$ and $\hGamma_{i,j}^2 = (\hmu_{i,T_{i,n}}-\hmu_{j,T_{j,n}})^2$ we finally obtain an upper--bound on the regret of the form

\begin{align*}
\R_n(\A) \leq \frac{1}{n}\sum_{i\neq i^*} T_{i,n} \hDelta_i + \frac{1}{n^2}\sum_{i=1}^K\sum_{j\neq i} T_{i,n}T_{j,n} \hGamma_{i,j}^2.
\end{align*}

In the following we refer to the two terms as $\R_{n}^{\hDelta}$ and $\R_n^{\hGamma}$.

%% Psuedo-regret

\subsection{The Pseudo--Regret}\label{app:psuedo.regret}

Similar to what is done in the standard bandit problem, we can introduce a different notion of regret. Starting from the last equation in the previous section, we define the pseudo--regret
\begin{align*}
\tR_n(\A) = \frac{1}{n}\sum_{i\neq i^*} T_{i,n} \Delta_i + \frac{2}{n^2}\sum_{i=1}^K\sum_{j\neq i} T_{i,n}T_{j,n} \Gamma_{i,j}^2,
\end{align*}
where the empirical values $\hDelta_i$ and $\hGamma_{i,j}$ are substituted by their corresponding exact values~\footnote{Notice that the factor 2 in front of the second term is due to a rough upper bounding used in the proof of Lemma~\ref{l:psuedo.regret}.}. In the following we show that the true and pseudo regrets different for values that tend to zero with high probability. 

\begin{proof}(Lemma~\ref{l:psuedo.regret})

We define a high--probability event in which the empirical values and the true values only differ for small quantities
\begin{align*}
\calE = \bigg\lbrace& \forall i=1,\ldots, K, \;\forall s=1,\ldots,n, \;\;\big| \hmu_{i,s} - \mu_i \big| \leq \sqrt{\frac{\log 1/\delta}{2s}} \;\text{ and }\; \big| \hvar_{i,s} - \var_i \big| \leq 5 \sqrt{\frac{\log 1/\delta}{2s}} \bigg\rbrace.
\end{align*}
Using Chernoff--Hoeffding inequality and a union bound over arms and rounds, we have that $\Prob[\calE^C] \leq 6nK\delta$. Under this event we rewrite the empirical $\hDelta_i$ as
\begin{align*}
\hDelta_i &= \Delta_i - (\var_{i}-\var_{i^*}) + \rho(\mu_{i} - \mu_{i^*}) + (\hvar_{i,T_{i,n}}-\tvar_{i,T_{i,n}}) - \rho(\hmu_{i,T_{i,n}} - \tmu_{i,T_{i,n}}) \\
&\leq \Delta_i + 2(5+\rho) \sqrt{\frac{\log 1/\delta}{2T_{i,n}}}.
\end{align*}
Similarly, $\hGamma_{i,j}$ is upper--bounded as
\begin{align*}
|\hGamma_{i,j}| &= |\Gamma_{i,j} -\mu_i + \mu_{j} + \hmu_{i,T_{i,n}}-\hmu_{j,T_{j,n}}| \\
&\leq |\Gamma_{i,j}| + \sqrt{\frac{\log 1/\delta}{2T_{i,n}}} + \sqrt{\frac{\log 1/\delta}{2T_{j,n}}}.
\end{align*}
Thus the regret can be written as

\begin{small}
\begin{align*}
\R_n(\A) &\leq \frac{1}{n}\sum_{i\neq i^*} T_{i,n} \Big(\Delta_i + 2(5+\rho) \sqrt{\frac{\log 1/\delta}{2T_{i,n}}}\Big) +  \frac{1}{n^2}\sum_{i=1}^K\sum_{j\neq i} T_{i,n}T_{j,n} \Big(|\Gamma_{i,j}| + \sqrt{\frac{\log 1/\delta}{2T_{i,n}}} + \sqrt{\frac{\log 1/\delta}{2T_{j,n}}}\Big)^2 \\
&\leq \frac{1}{n}\sum_{i\neq i^*} T_{i,n} \Delta_i + \frac{5+\rho}{n}\sum_{i\neq i^*}\sqrt{2T_{i,n}\log 1/\delta} + \frac{2}{n^2}\sum_{i=1}^K\sum_{j\neq i} T_{i,n}T_{j,n} \Gamma_{i,j}^2 \\
&\quad+ \frac{2\sqrt{2}}{n^2}\sum_{i=1}^K\sum_{j\neq i} T_{j,n}\log 1/\delta + \frac{2\sqrt{2}}{n^2}\sum_{i=1}^K\sum_{j\neq i} T_{i,n} \log 1/\delta\\
&\leq \frac{1}{n}\sum_{i\neq i^*} T_{i,n} \Delta_i + \frac{2}{n^2}\sum_{i=1}^K\sum_{j\neq i} T_{i,n}T_{j,n} \Gamma_{i,j}^2+ (5+\rho)\sqrt{\frac{2K\log 1/\delta}{n}} + 4\sqrt{2}\frac{K\log 1/\delta}{n}.
\end{align*}
\end{small}

where in the next to last passage we used Jensen's inequality for concave functions and rough upper bounds on other terms ($K-1< K$, $\sum_{i\neq i^*} T_{i,n}\leq n$). By recalling the definition of $\tR_n(\A)$ we finally obtain
\begin{align*}
\R_n(\A) \leq \tR_n(\A) + (5+\rho)\sqrt{\frac{2K\log 1/\delta}{n}} + 4\sqrt{2}\frac{K\log 1/\delta}{n},
\end{align*}
with probability $1-6nK\delta$. Thus we can conclude that any upper bound on the pseudo--regret $\tR_n(\A)$ is a valid upper bound for the true regret $\R_n(\A)$, up to a decreasing term of order $O(\sqrt{K/n})$.

\end{proof}

%%%%%%%%%%%%%%%%%%%%%%%%%%%%%%%%%%%%%%%%%%%%%%%%%%%%%%%%%%%%%%%%%%%%%%%
% MVLCB THEORETICAL ANALYSIS
%%%%%%%%%%%%%%%%%%%%%%%%%%%%%%%%%%%%%%%%%%%%%%%%%%%%%%%%%%%%%%%%%%%%%%%

\section{\textsl{MV-LCB} Theoretical Analysis}\label{app:mvlcb}

In order to simplify the notation in the following we use $b = 2(5+\rho)$.

\begin{proof}(Theorem~\ref{thm:mvlcb.regret})

We begin by defining a high--probability event $\calE$ as
\begin{align*}
\calE = \bigg\lbrace& \forall i=1,\ldots, K, \;\forall s=1,\ldots,n, \;\;\big| \hmu_{i,s} - \mu_i \big| \leq \sqrt{\frac{\log 1/\delta}{2s}} \;\text{ and }\; \big| \hvar_{i,s} - \var_i \big| \leq 5 \sqrt{\frac{\log 1/\delta}{2s}} \bigg\rbrace.
\end{align*}
Using Chernoff--Hoeffding inequality and a union bound over arms and rounds, we have that $\Prob[\calE^C] \leq 6nK\delta$.

We now introduce the definition of the algorithm. Consider any time $t$ when arm $i\neq i^*$ is pulled (i.e., $I_t = i$). By definition of the algorithm in Figure~\ref{f:lcb}, $i$ is selected if its corresponding index $B_{i, T_{i,t-1}}$ is bigger than for any other arm, notably the best arm $i^*$. By recalling the definition of the index and the empirical mean--variance at time $t$, we have
\begin{align*}
\hvar_{i, T_{i,t-1}} - \rho \hmu_{i, T_{i,t-1}} &-(5+\rho) \sqrt{\frac{\log 1/\delta}{2T_{i,t-1}}} = B_{i, T_{i,t-1}} \leq\\
& \leq B_{i^*, T_{i*, t-1}} = \hvar_{i^*, T_{i^*,t-1}} - \rho \hmu_{i^*, T_{i^*,t-1}} -(5+\rho) \sqrt{\frac{\log 1/\delta}{2T_{i^*,t-1}}}.
\end{align*}
Over all the possible realizations, we now focus on the realizations in $\calE$. In this case, we can rewrite the previous condition as
\begin{align*}
\var_{i} - \rho \mu_{i} &-2(5+\rho) \sqrt{\frac{\log 1/\delta}{2T_{i,t-1}}} \leq B_{i, T_{i,t-1}} \leq B_{i^*, T_{i*, t-1}} \leq \var_{i^*}- \rho \mu_{i^*}.
\end{align*}
Let time $t$ be the last time when arm $i$ is pulled until the final round $n$, then $T_{i,t-1} = T_{i,n} - 1$ and
\begin{align*}
T_{i,n} \leq \frac{2 (5+\rho)^2}{\Delta_i^2} \log \frac{1}{\delta} + 1,
\end{align*}
which suggests that the suboptimal arms are pulled only few times with high probability. Plugging the bound in the regret in eq.~\ref{eq:pseudo.regret} leads to the final statement
\begin{align*}
\tR_n(\A) &\leq \frac{1}{n}\sum_{i\neq i^*}\frac{b^2 \log 1/\delta}{\Delta_i} + \frac{1}{n}\sum_{i\neq i^*}\frac{4b^2 \log 1/\delta}{\Delta_i^2}\Gamma_{i^*,i}^2 + \frac{1}{n^2}\sum_{i\neq i^*}\mathop{\sum_{j\neq i}}_{j\neq i^*}\frac{2b^4 (\log 1/\delta)^2}{\Delta_i^2 \Delta_j^2}\Gamma_{i,j}^2 + \frac{5K}{n},
\end{align*}
with probability $1-6nK\delta$.

We now move from the previous high--probability bound to a bound in expectation. The pseudo--regret is (roughly) bounded as $\tR_n(\A) \leq 2+\rho$ (by bounding $\Delta_i\leq 1+\rho$ and $\Gamma\leq 1$), thus
\begin{align*}
\E[\tR_{n}(\A)] = \E\big[\tR_n(\A) \I\{\calE\}\big] + \E\big[\tR_n(\A) \I\{\calE^C\}\big] \leq \E\big[\tR_n(\A) \I\{\calE\}\big] + (2+\rho) \Prob\big[\calE^C\big].
\end{align*}
%\begin{align*}
%\E[\tR_{n}(\A)] = \int_{-\infty}^u t f_t(t) dt + \int_{u}^{2+\rho} t f_t(t) dt \leq u \Prob[\R_{n}(\A) \leq u] + \Big(2+\rho\Big)\Prob[\R_{n}(\A) > u].
%\end{align*}
%
By using the previous high--probability bound and recalling that $\Prob[\calE^C] \leq 6nK\delta$, we have
\begin{align*}
\E[\tR_n(\A)] &\leq \frac{1}{n}\sum_{i\neq i^*}\frac{b^2 \log 1/\delta}{\Delta_i} + \frac{1}{n}\sum_{i\neq i^*}\frac{4b^2 \log 1/\delta}{\Delta_i^2}\Gamma_{i^*,i}^2 + \frac{1}{n^2}\sum_{i\neq i^*}\mathop{\sum_{j\neq i}}_{j\neq i^*}\frac{2b^4 (\log 1/\delta)^2}{\Delta_i^2 \Delta_j^2}\Gamma_{i,j}^2 \\
&+ \frac{5K}{n} + (2+\rho) 6nK\delta.
\end{align*}
The final statement of the lemma follows by tuning the parameter $\delta=1/n^2$ so as to have a regret bound decreasing with $n$.
\end{proof}

While a high--probability bound for $\R_n$ can be immediately obtained from Lemma~\ref{l:psuedo.regret}, the expectation of $\R_n$ is reported in the next corollary.

\begin{proof}
Since the mean--variance $-\rho \leq \hMV \leq 1/4$, the regret is bounded by $-1/4 - \rho \leq \R_n(\A) \leq 1/4 + \rho$. Thus we have
\begin{align*}
\E[\R_{n}(\A)] = \leq u \Prob[\R_{n}(\A) \leq u] + \Big(\frac{1}{4}+\rho\Big)\Prob[\R_{n}(\A) > u].
\end{align*}
By taking $u$ equal to the previous high--probability bound and recalling that $\Prob[\calE^C] \leq 6nK\delta$, we have
\begin{align*}
\E[\R_n(\A)] &\leq \frac{1}{n}\sum_{i\neq i^*}\frac{b^2 \log 1/\delta}{\Delta_i} + \frac{1}{n}\sum_{i\neq i^*}\frac{4b^2 \log 1/\delta}{\Delta_i^2}\Gamma_{i^*,i}^2 + \frac{1}{n^2}\sum_{i\neq i^*}\mathop{\sum_{j\neq i}}_{j\neq i^*}\frac{2b^4 (\log 1/\delta)^2}{\Delta_i^2 \Delta_j^2}\Gamma_{i,j}^2 \nonumber\\
&\quad+ \frac{5K}{n} + b\sqrt{\frac{K\log 1/\delta}{2n}} + 4\sqrt{2}\frac{K\log 1/\delta}{n} + \Big(\frac{1}{4}+\rho\Big)6nK\delta.
\end{align*}
The final statement of the lemma follows by tuning the parameter $\delta=1/n^2$ so as to have a regret bound decreasing with $n$.
\end{proof}

%%%%%%%%%%%%%%%%%%%%%%%%%%%%%%%%%%%%%%%%%%%%%%%%%%%%%%%%%%%%%%%%%%%%%%%
% EXP-EXP THEORETICAL ANALYSIS
%%%%%%%%%%%%%%%%%%%%%%%%%%%%%%%%%%%%%%%%%%%%%%%%%%%%%%%%%%%%%%%%%%%%%%%

\section{Exp--Exp Theoretical Analysis}\label{app:expexp}

During the exploitation phase the algorithm pulls arm $\hat i^*$ with the smallest empirical variance estimated during the exploration phase of length $\tau$. As a result, the number of pulls of each arm is

\begin{align}
T_{i,n} = \frac{\tau}{K} + (n-\tau)\I\{i=\hat i^*\}
\end{align}

We analyze the two terms of the regret separately.

\begin{align*}
\tR_n^{\Delta} = \frac{1}{n} \sum_{i\neq i^*} \Big(\frac{\tau}{K} + (n-\tau)\I\{i=\hat i^*\}\Big) \Delta_i = \frac{\tau}{nK} \sum_{i\neq i^*} \Delta_i + \frac{n-\tau}{n} \sum_{i\neq i^*} \underbrace{\Delta_i \I\{i=\hat i^*\}}_{(a)}.
\end{align*}

We notice that the only random variable in this formulation is the best arm $\hat i^*$ at the end of the exploration phase. We thus compute the expected value of $\tR_n^\Delta$.

\begin{align*}
\E[(a)] &= \Prob[i=\hat i^*] \Delta_i = \Prob[\forall j\neq i, \; \hvar_{i,\tau/K} \leq \hvar_{j,\tau/K}]\Delta_i \\
&\leq \Prob[\hvar_{i,\tau/K} \leq \hvar_{i^*,\tau/K}]\Delta_i = \Prob\big[(\hvar_{i,\tau/K} - \var_i) + (\var_{i^*} - \hvar_{i^*, \tau/K}) \leq \Delta_i\big]\Delta_i \\
&\leq 2\Delta_i\exp\Big(-\frac{\tau}{K}\Delta_i^2\Big)
\end{align*}

The second term in the regret can be bounded as follows.

\begin{align*}
\tR_n^{\Gamma} &= \frac{1}{n^2} \sum_{i=1}^K \sum_{j\neq i} \Big(\frac{\tau}{K} + (n-\tau)\I\{i=\hat i^*\} \Big)\Big(\frac{\tau}{K} + (n-\tau)\I\{j=\hat i^*\} \Big) \Gamma^2_{i,j}\\
&= \frac{1}{n^2} \sum_{i=1}^K \sum_{j\neq i} \Big(\frac{\tau^2}{K^2} + (n-\tau)^2\I\{i=\hat i^*\}\I\{j=\hat i^*\} + \frac{\tau}{K}(n-\tau)\I\{j=\hat i^*\} + \frac{\tau}{K}(n-\tau)\I\{i=\hat i^*\} \Big) \Gamma^2_{i,j}\\
&= \frac{\tau^2}{n^2K^2} \sum_{i=1}^K \sum_{j\neq i} \Gamma^2_{i,j} + 2\frac{(n-\tau)\tau}{K n^2} \sum_{i=1}^K \sum_{j\neq i} \Gamma^2_{i,j} \I\{i=\hat i^*\} \\
&\leq \frac{\tau}{n^2} + 2\frac{(n-\tau)\tau}{n^2}\leq 2\frac{\tau}{n}
\end{align*}

Grouping all the terms, \textsl{ExpExp} has an expected regret bounded as

\begin{align*}
\E[\tR_n(\A)] \leq 2\frac{\tau}{n} + 2\sum_{i\neq i^*} \Delta_i\exp\Big(-\frac{\tau}{K}\Delta_i^2\Big)
\end{align*}

We can now move to the worst--case analysis of the regret. Let $f(\Delta_i) = \Delta_i\exp\Big(-\frac{\tau}{K}\Delta_i^2\Big)$, the ``adversarial'' choice of the gap is determined by maximizing the regret, which corresponds to

\begin{align*}
f'(\Delta_i) &= \exp\Big(-\frac{\tau}{K}\Delta_i^2\Big) + \Delta_i\Bigg(-2 \frac{\tau}{K}\Delta_i\exp\Big(-\frac{\tau}{K}\Delta_i^2\Big)\Bigg)\\
&= \Big(1-2\frac{\tau}{K}\Delta_i^2\Big)\exp\Big(-\frac{\tau}{K}\Delta_i^2\Big) = 0,
\end{align*}

and leads to a worst--case choice for the gap of 

\begin{align*}
\Delta_i = \sqrt{\frac{K}{2\tau}}.
\end{align*}

The worst--case regret is then

\begin{align*}
\E[\tR_n(\A)] \leq 2\frac{\tau}{n} + (K-1) \sqrt{2K} \frac{1}{\sqrt{\tau}} \exp(-0.5) \leq2\frac{\tau}{n} + K^{3/2} \frac{1}{\sqrt{\tau}} 
\end{align*}

We can now choose the parameter $\tau$ minimizing the worst--case regret. Taking the derivative of the regret w.r.t. $\tau$ we obtain

\begin{align*}
\frac{d\E[\tR_n(\A)]}{d\tau} = \frac{2}{n} -\frac{1}{2}\Big(\frac{K}{\tau}\Big)^{3/2} = 0,
\end{align*}

thus leading to the optimal parameter $\tau = (n/ 4)^{2/3} K$. The final regret is thus bounded as

\begin{align*}
\E[\tR_n(\A)] \leq 3\frac{K}{n^{1/3}}.
\end{align*}

\newpage
\section{Additional Simulations}\label{app:simulations}

%%% Grid comparison
\subsection{Comparison between \textsl{MV-LCB} and \textsl{ExpExp} with $K=2$}

\begin{figure*}[th]
\begin{center}
\includegraphics[width=0.3\textwidth]{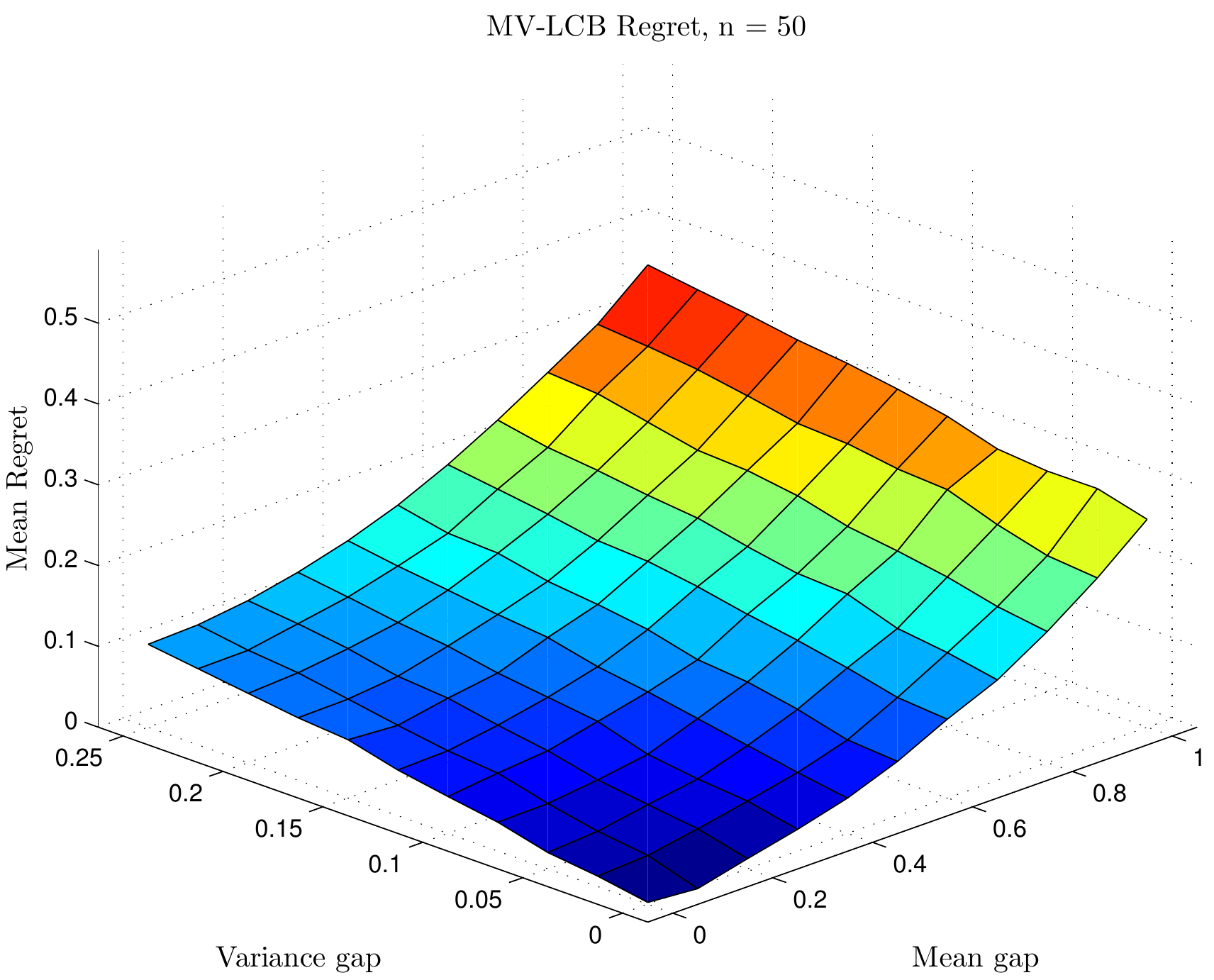}
\includegraphics[width=0.3\textwidth]{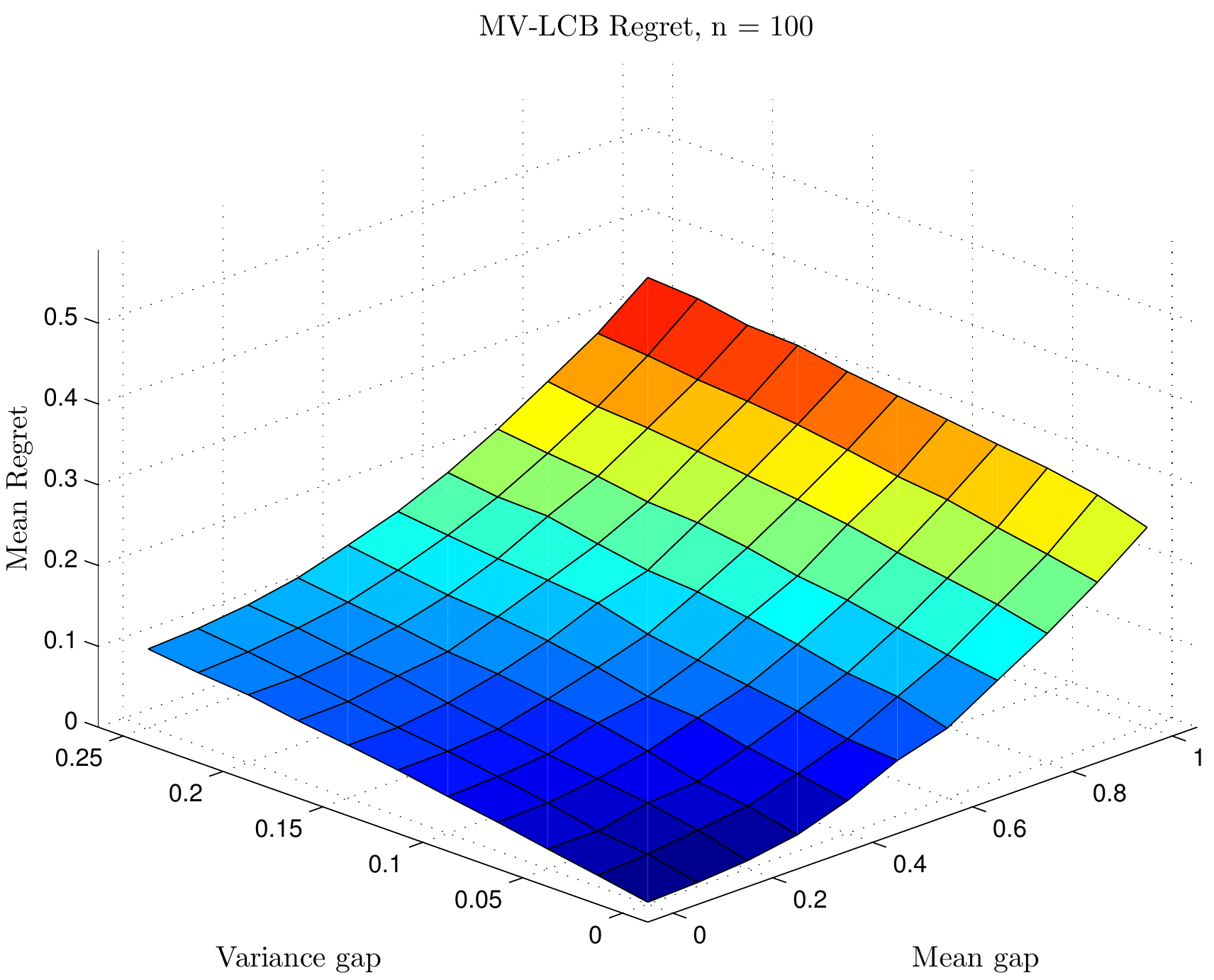}
\includegraphics[width=0.3\textwidth]{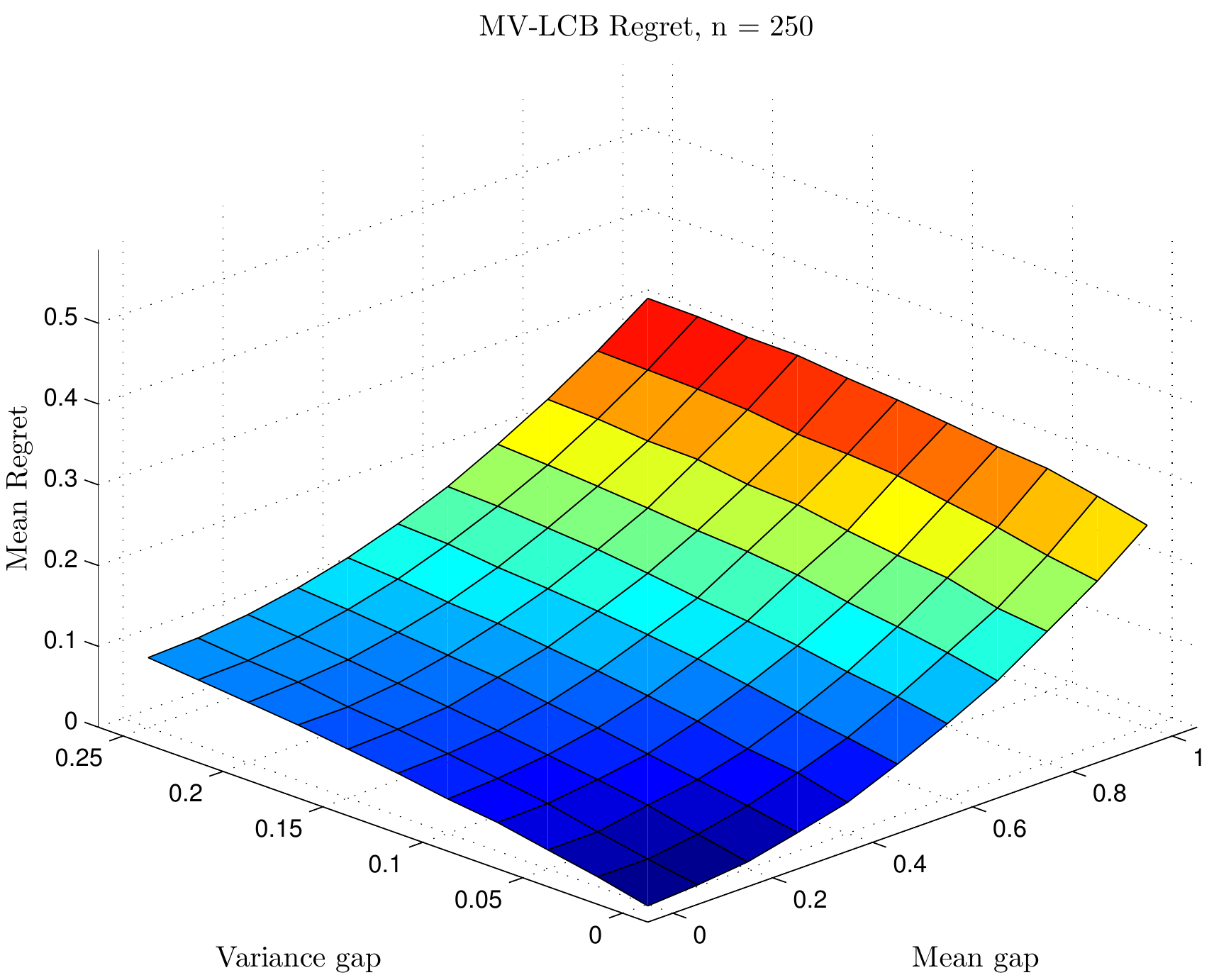}
\includegraphics[width=0.3\textwidth]{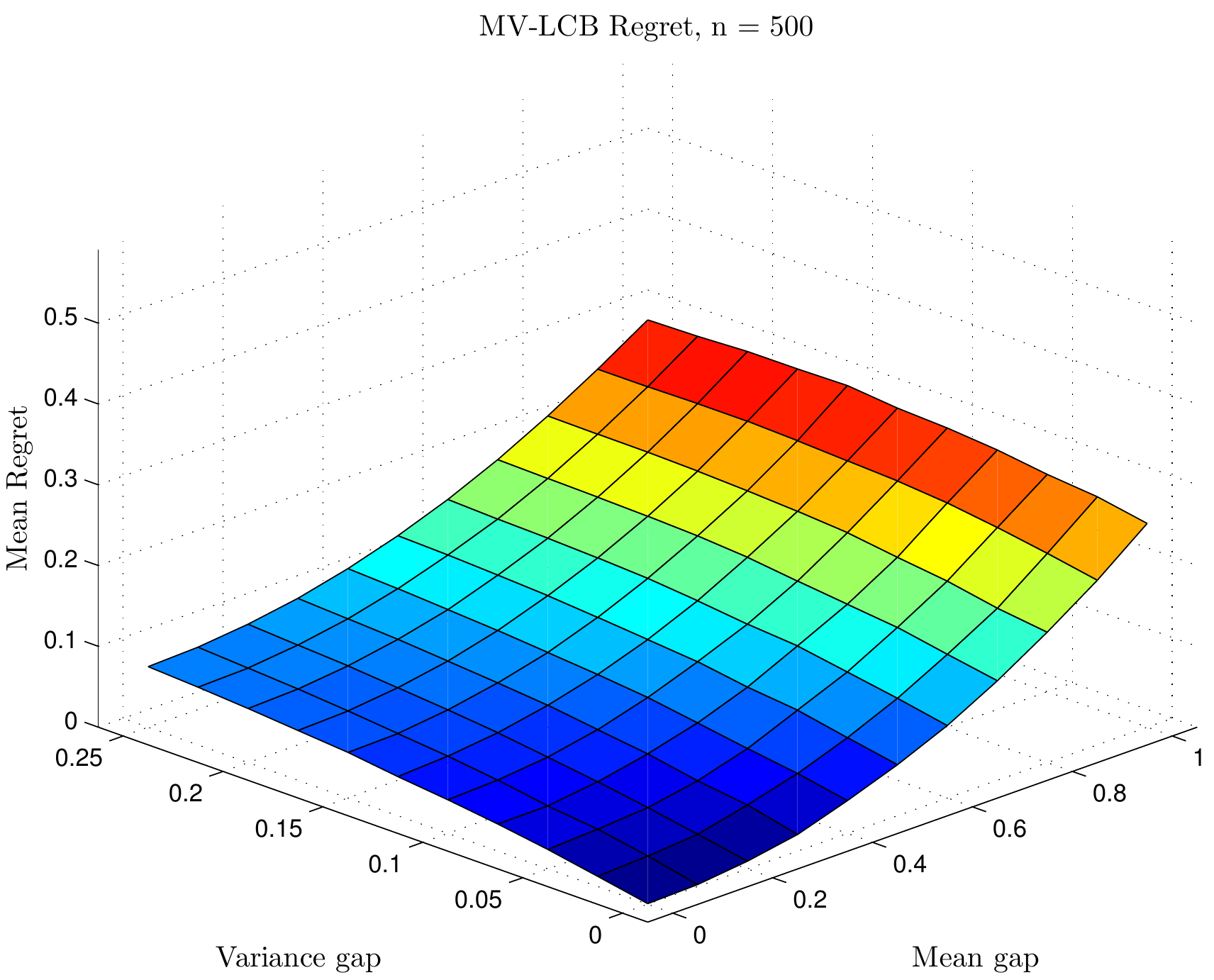}
\includegraphics[width=0.3\textwidth]{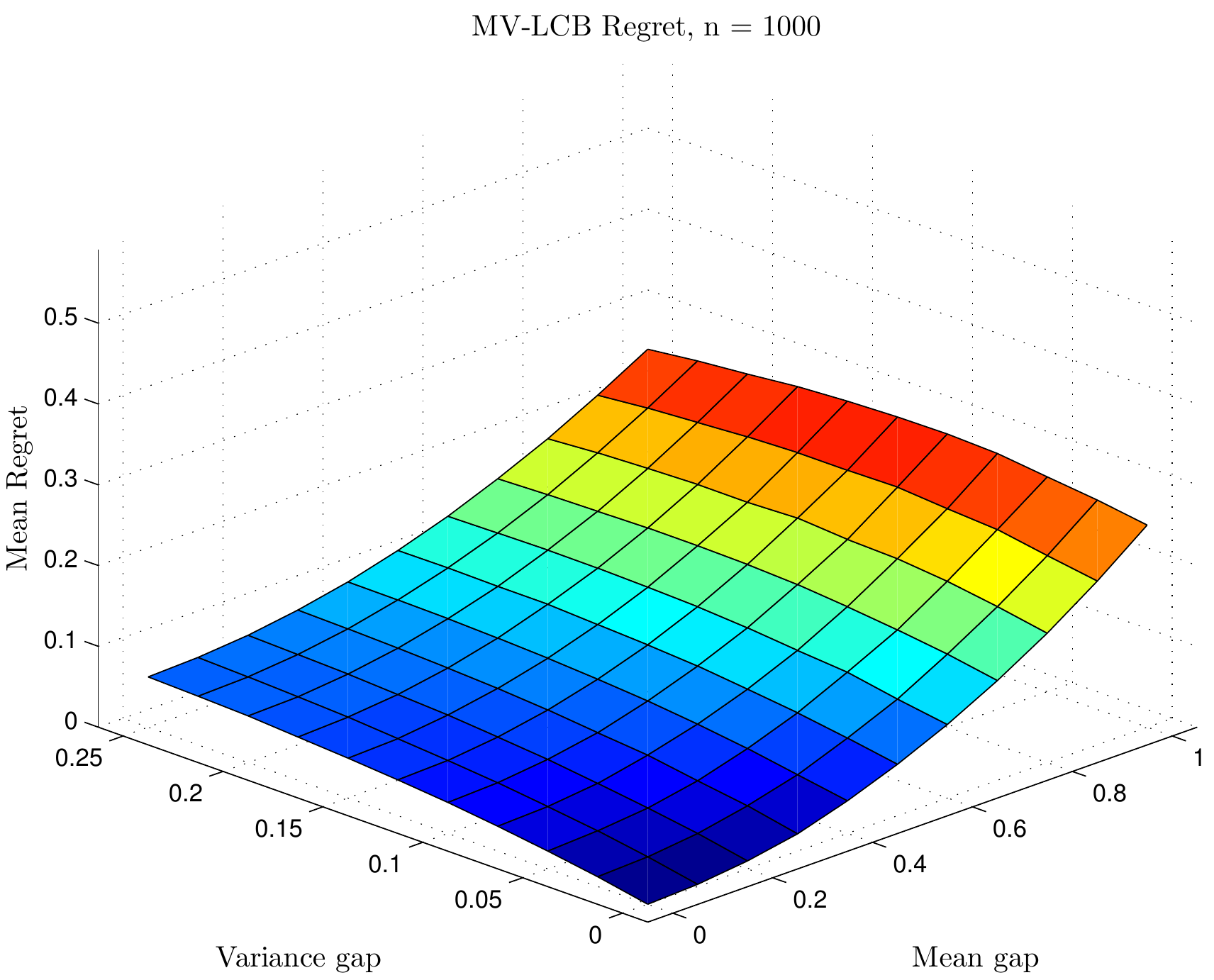}
\includegraphics[width=0.3\textwidth]{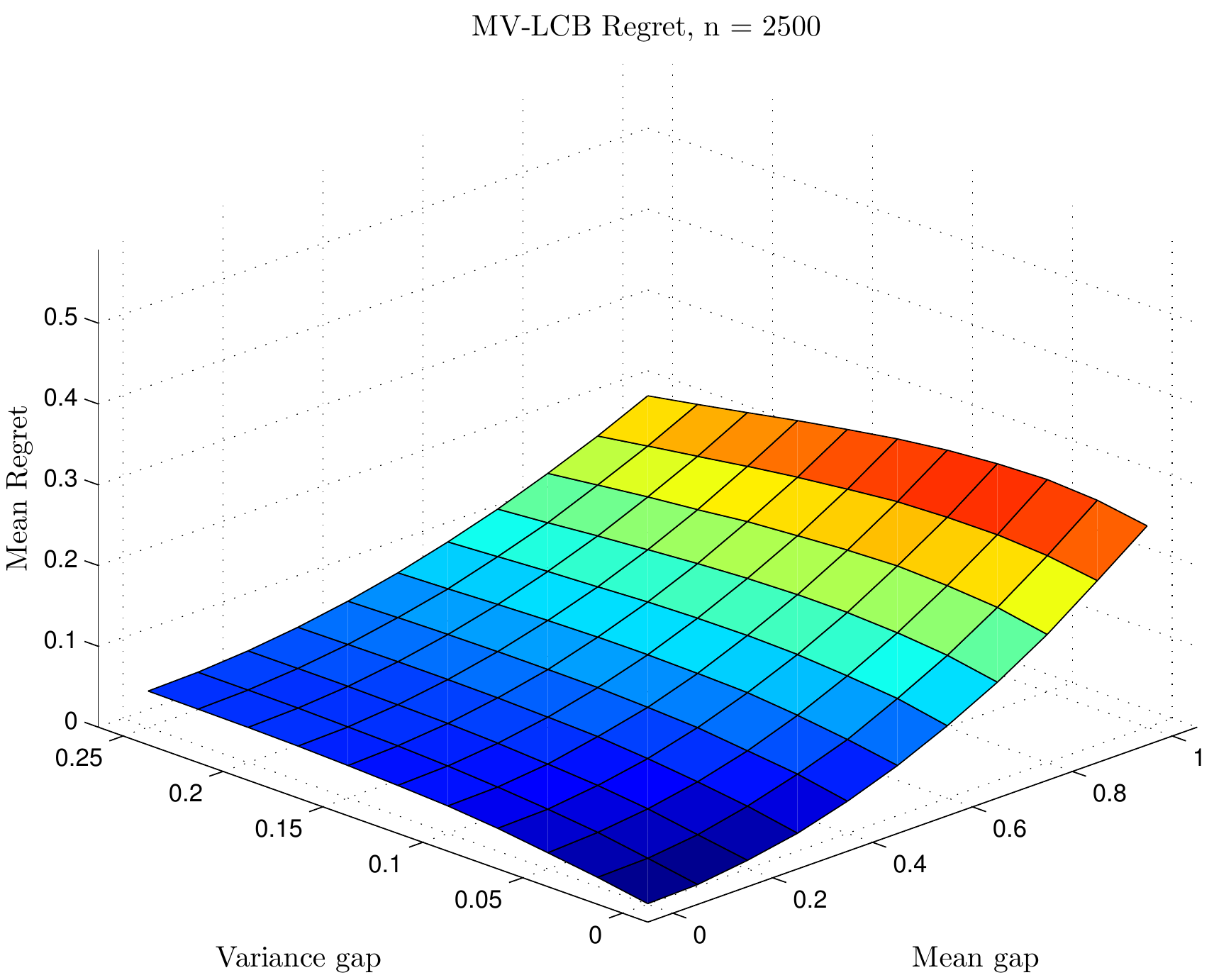}
\includegraphics[width=0.3\textwidth]{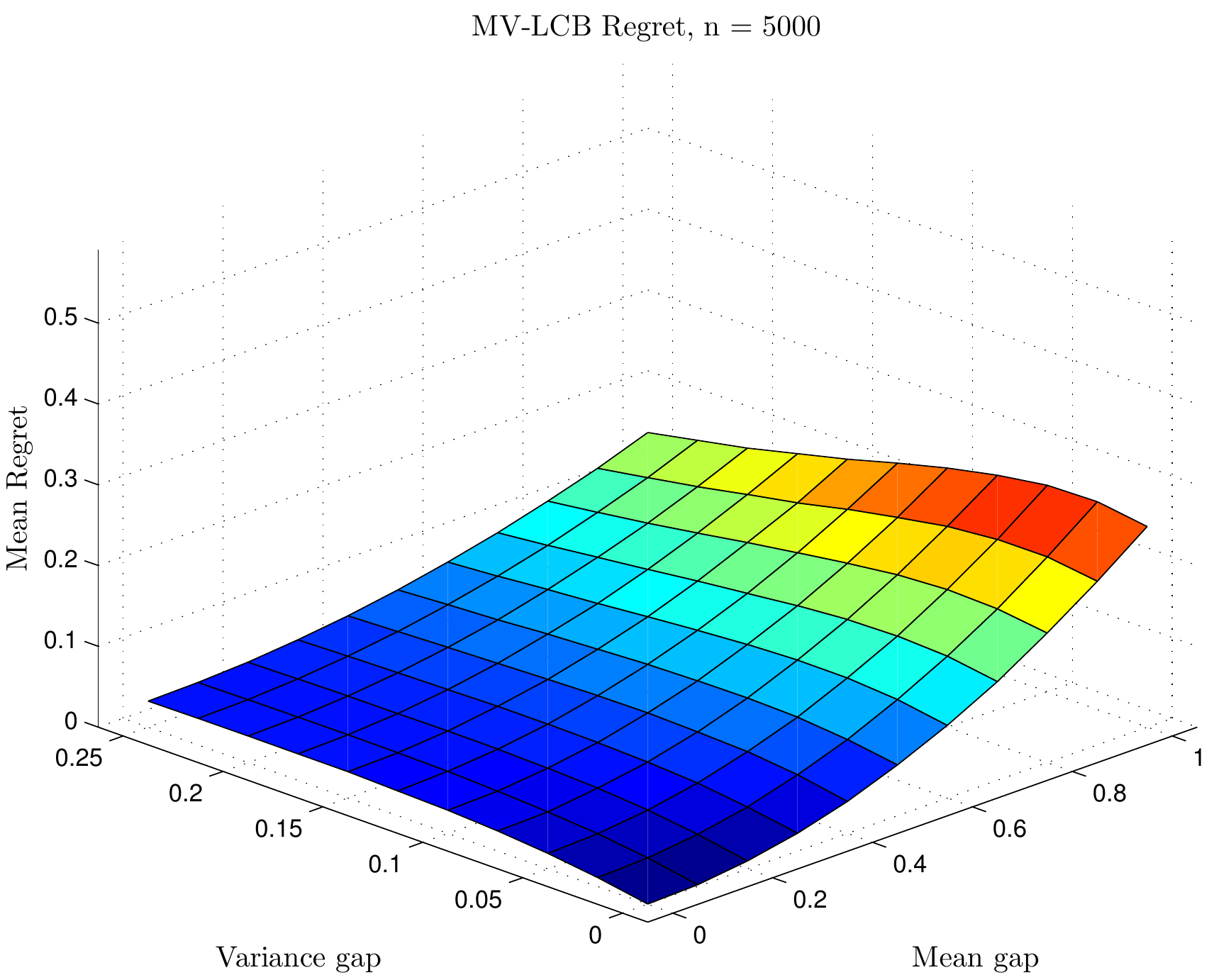}
\includegraphics[width=0.3\textwidth]{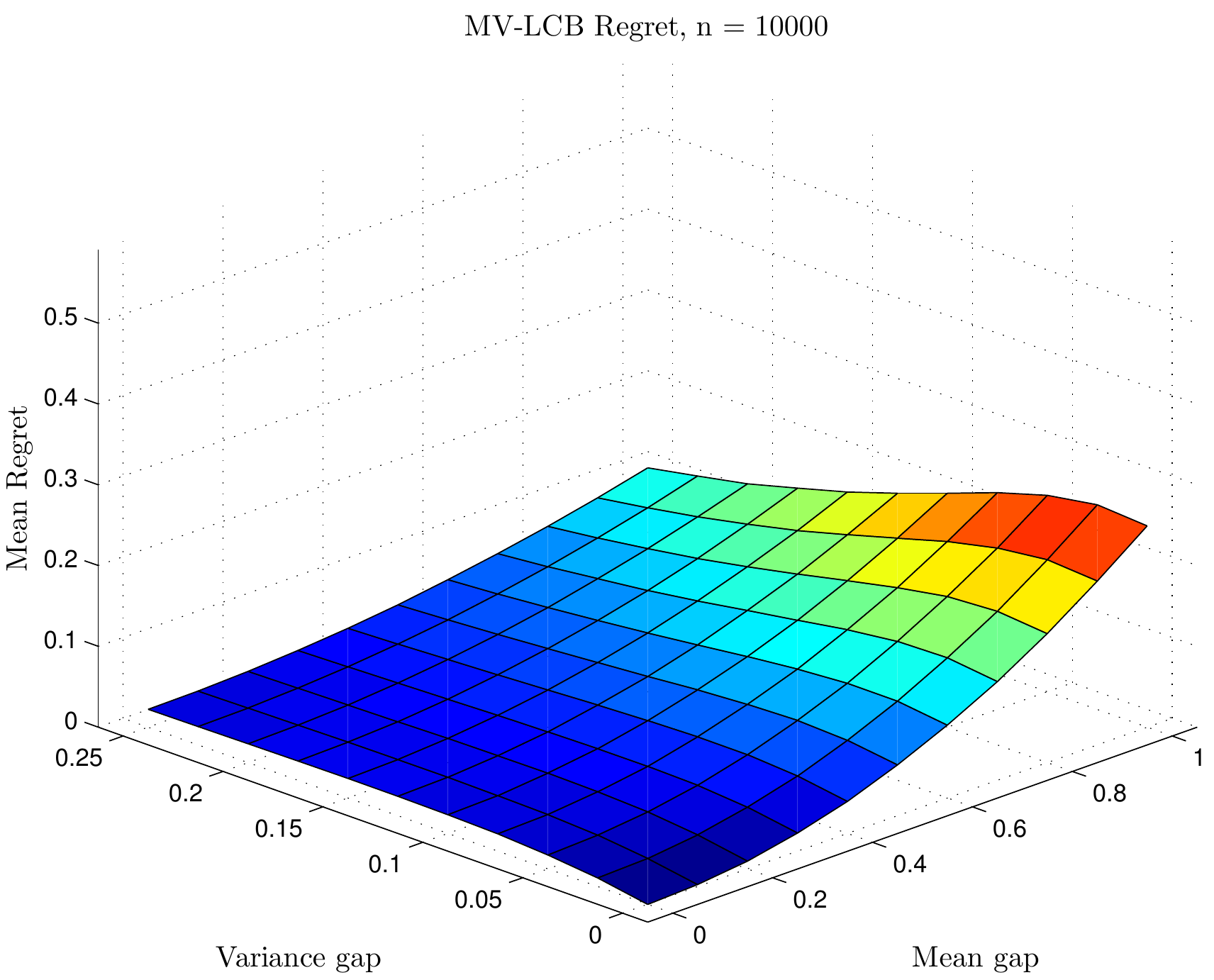}
\includegraphics[width=0.3\textwidth]{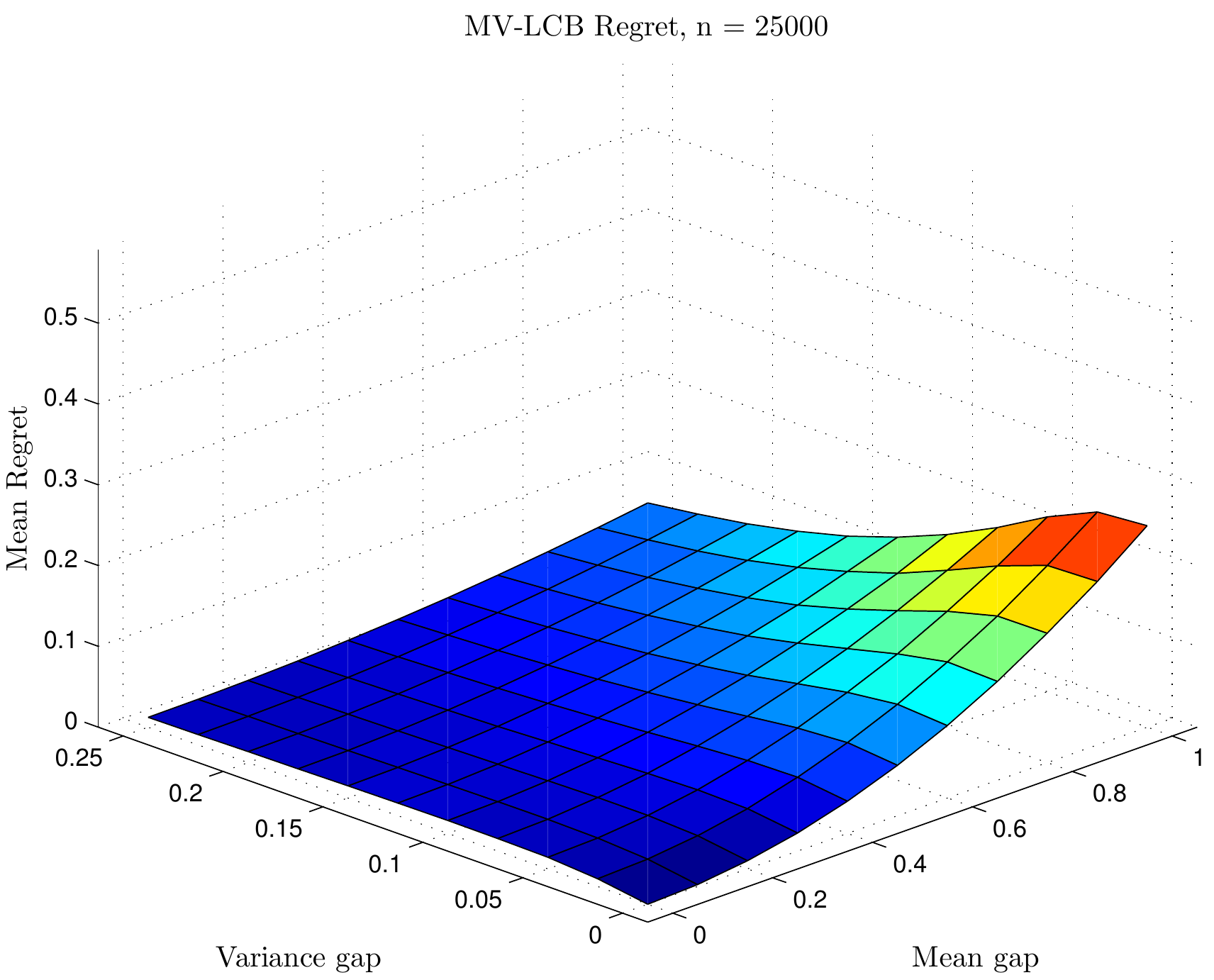}
\includegraphics[width=0.3\textwidth]{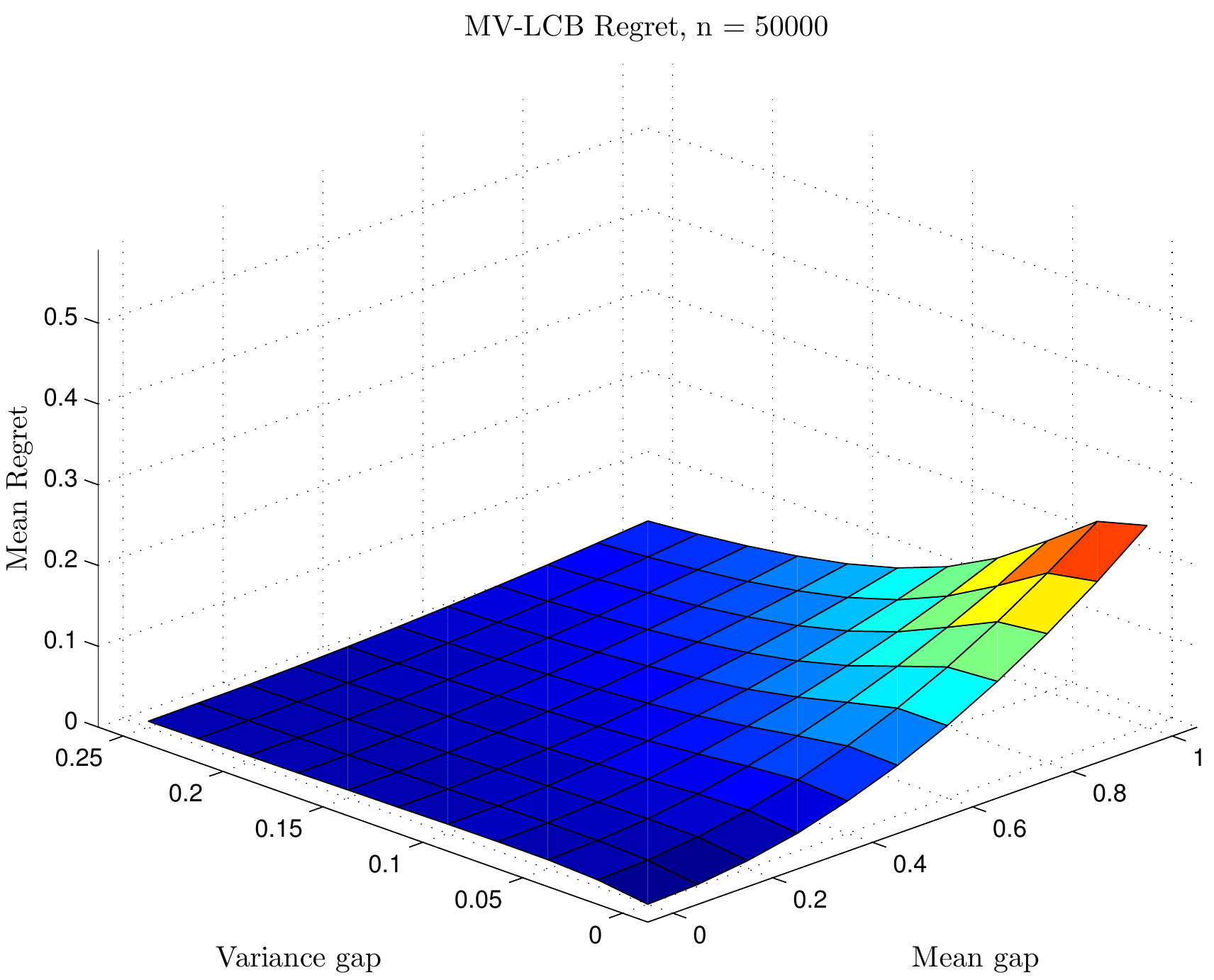}
\includegraphics[width=0.3\textwidth]{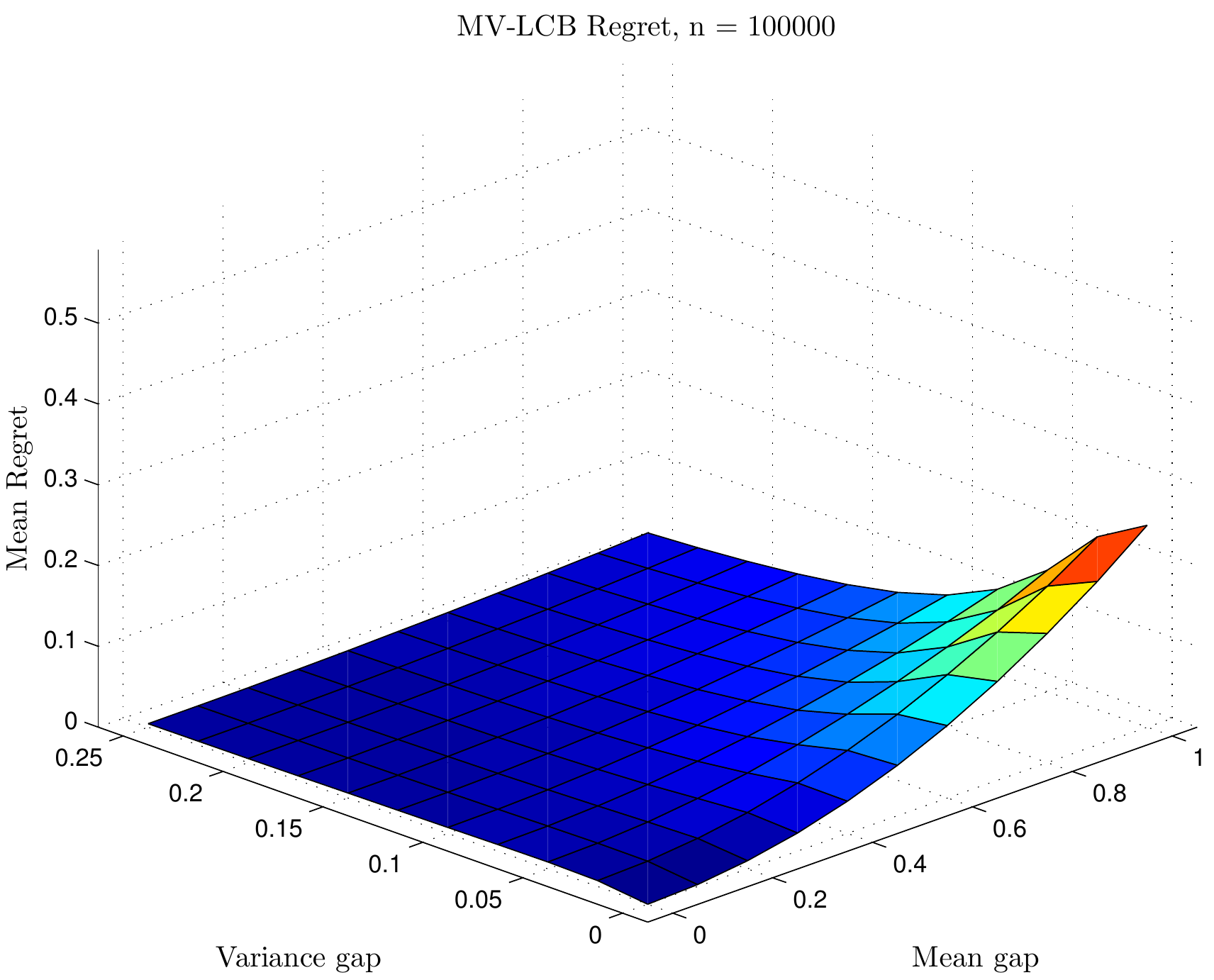}
\includegraphics[width=0.3\textwidth]{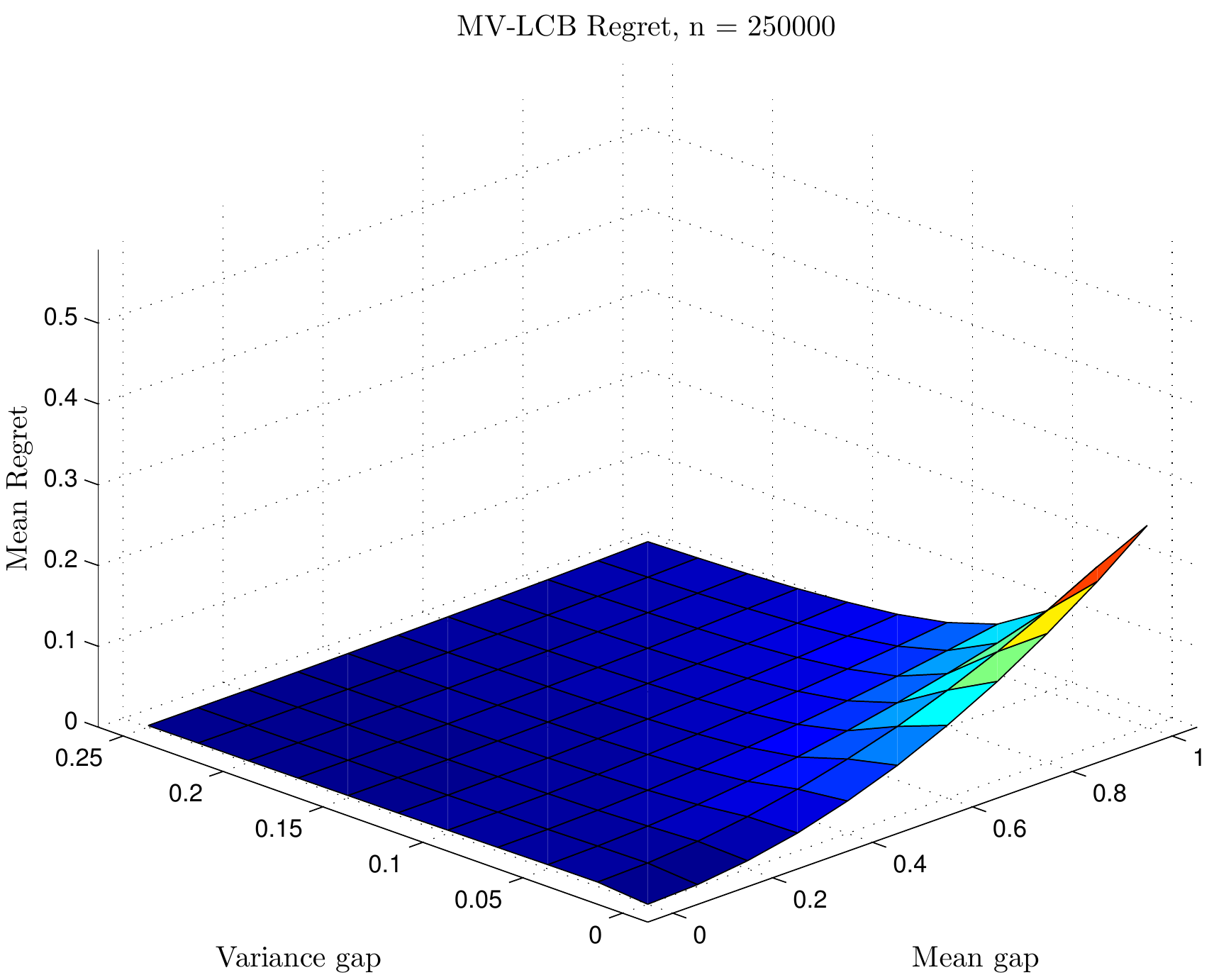}
\end{center}
\vspace{-0.4cm}
\caption{Regret $\R_n$ of \textsl{MV-LCB}.}\label{f:mvlcb.grid}
\vspace{-0.4cm}
\end{figure*}

\begin{figure*}[h]
\begin{center}
\includegraphics[width=0.32\textwidth]{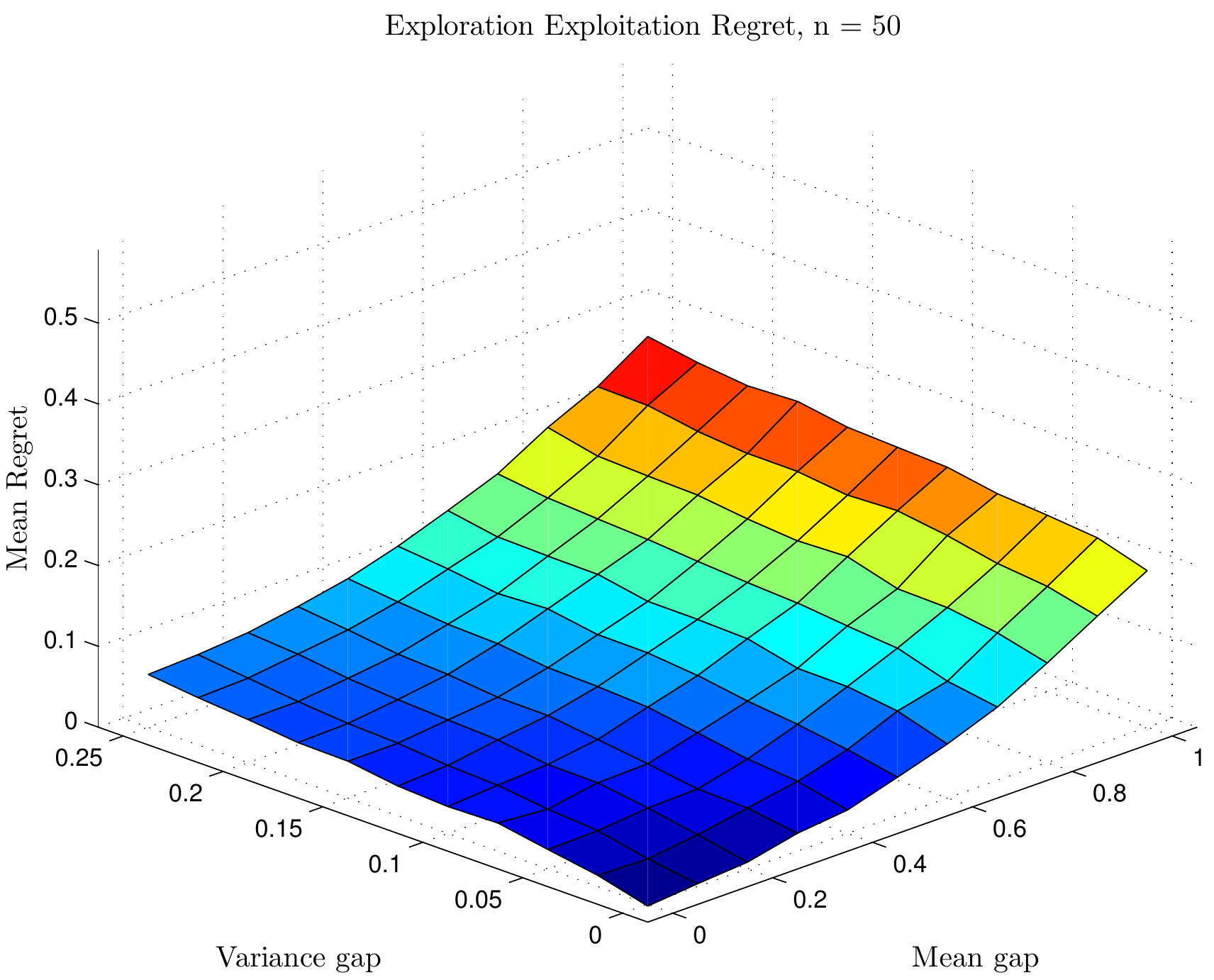}
\includegraphics[width=0.32\textwidth]{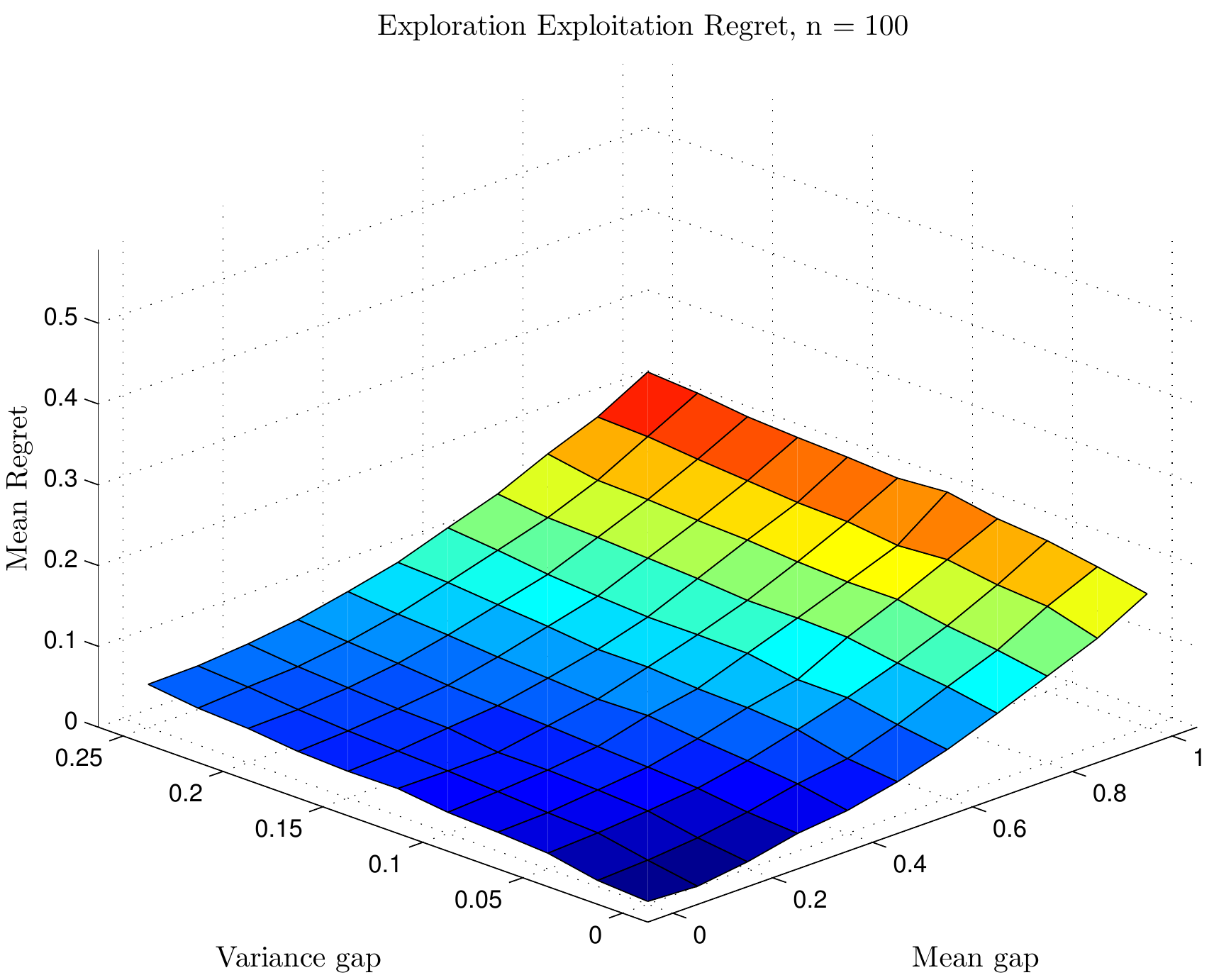}
\includegraphics[width=0.32\textwidth]{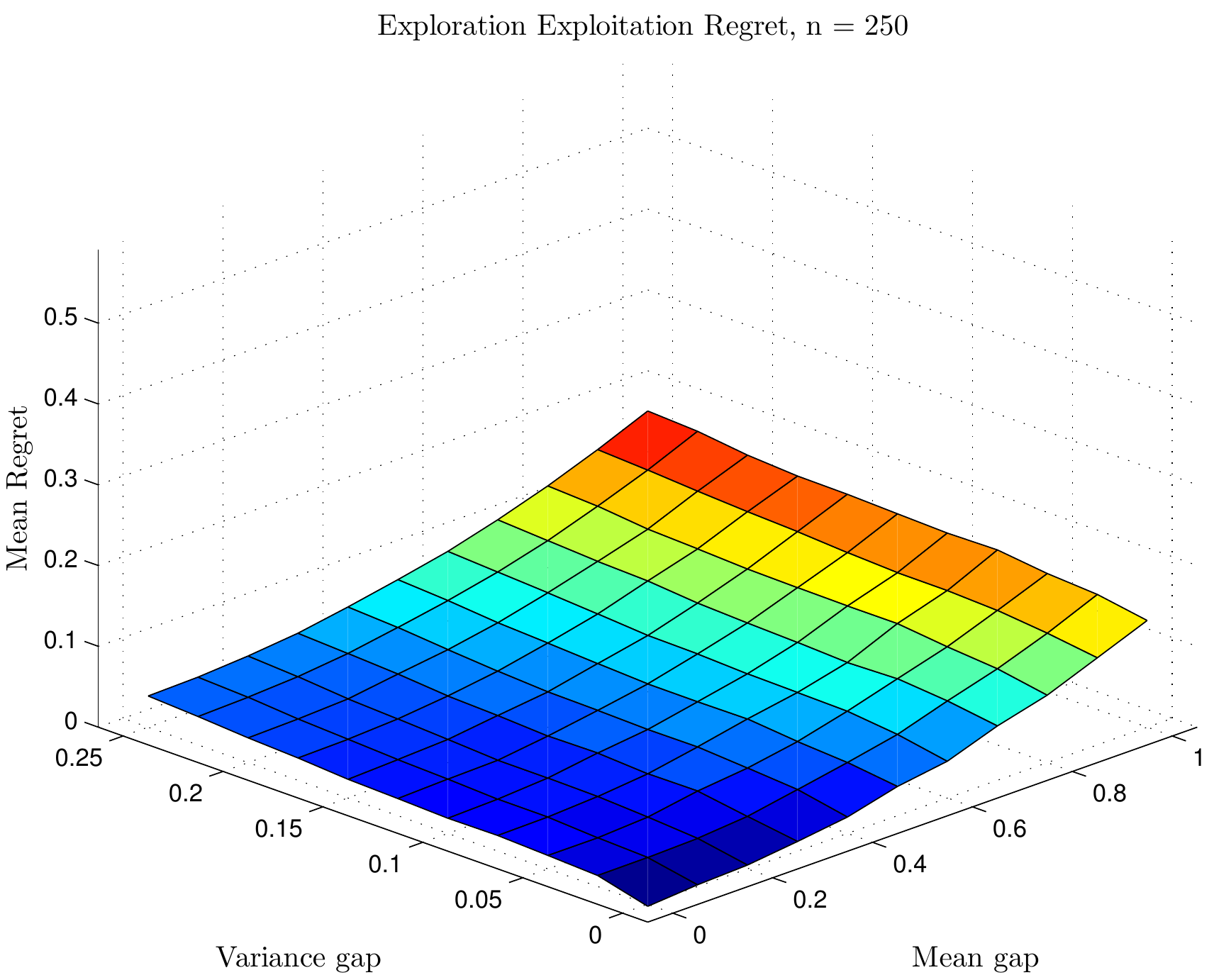}
\includegraphics[width=0.32\textwidth]{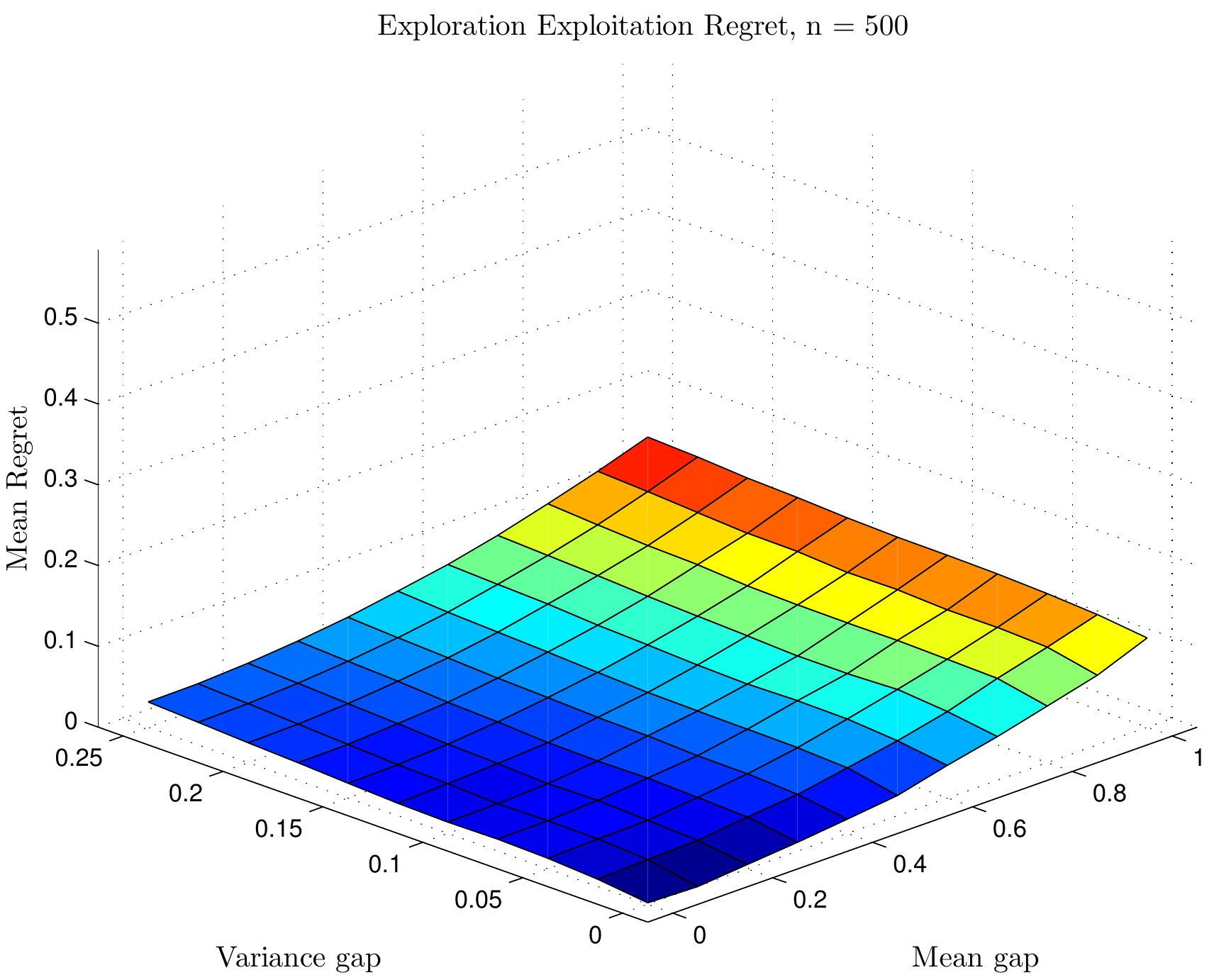}
\includegraphics[width=0.32\textwidth]{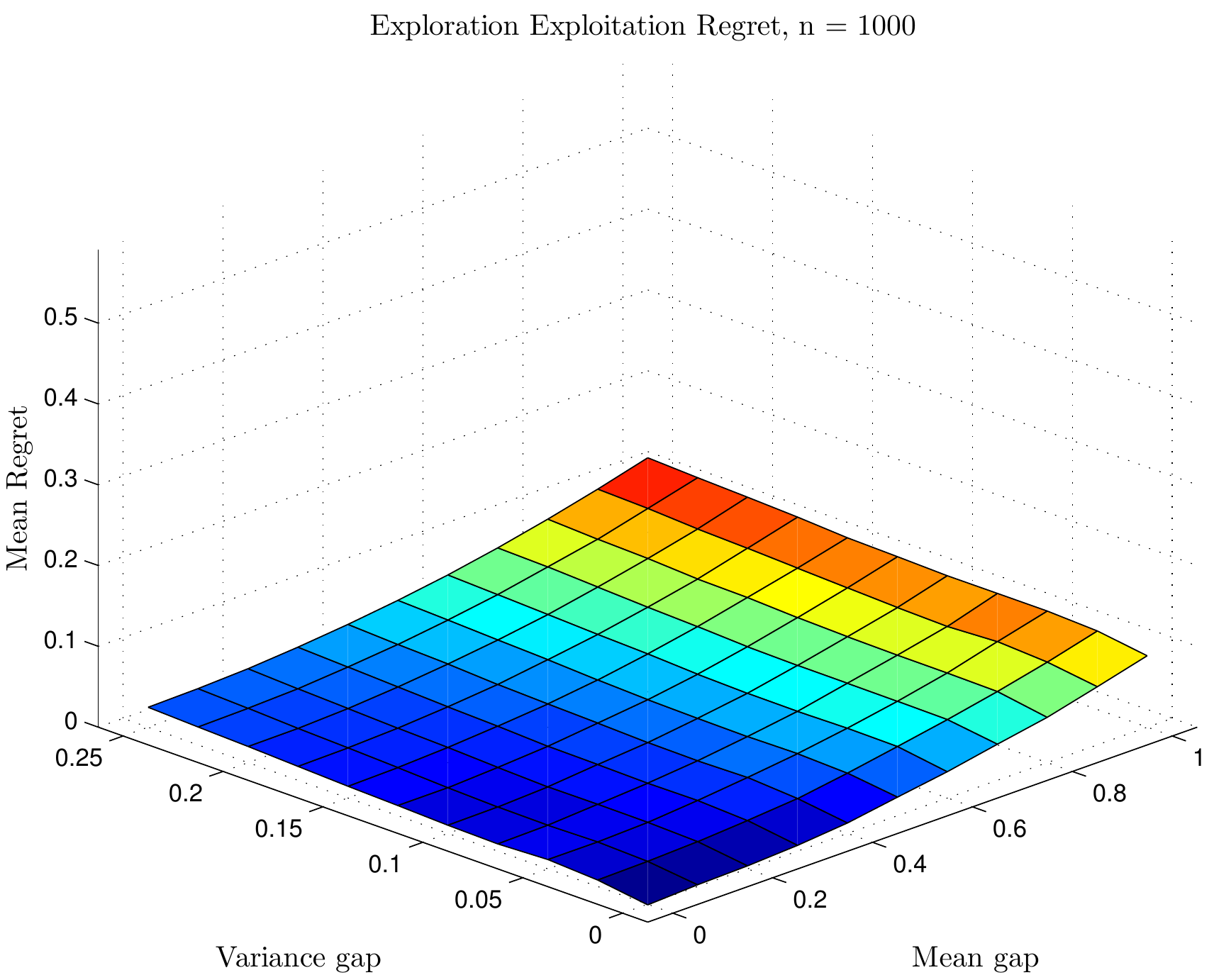}
\includegraphics[width=0.32\textwidth]{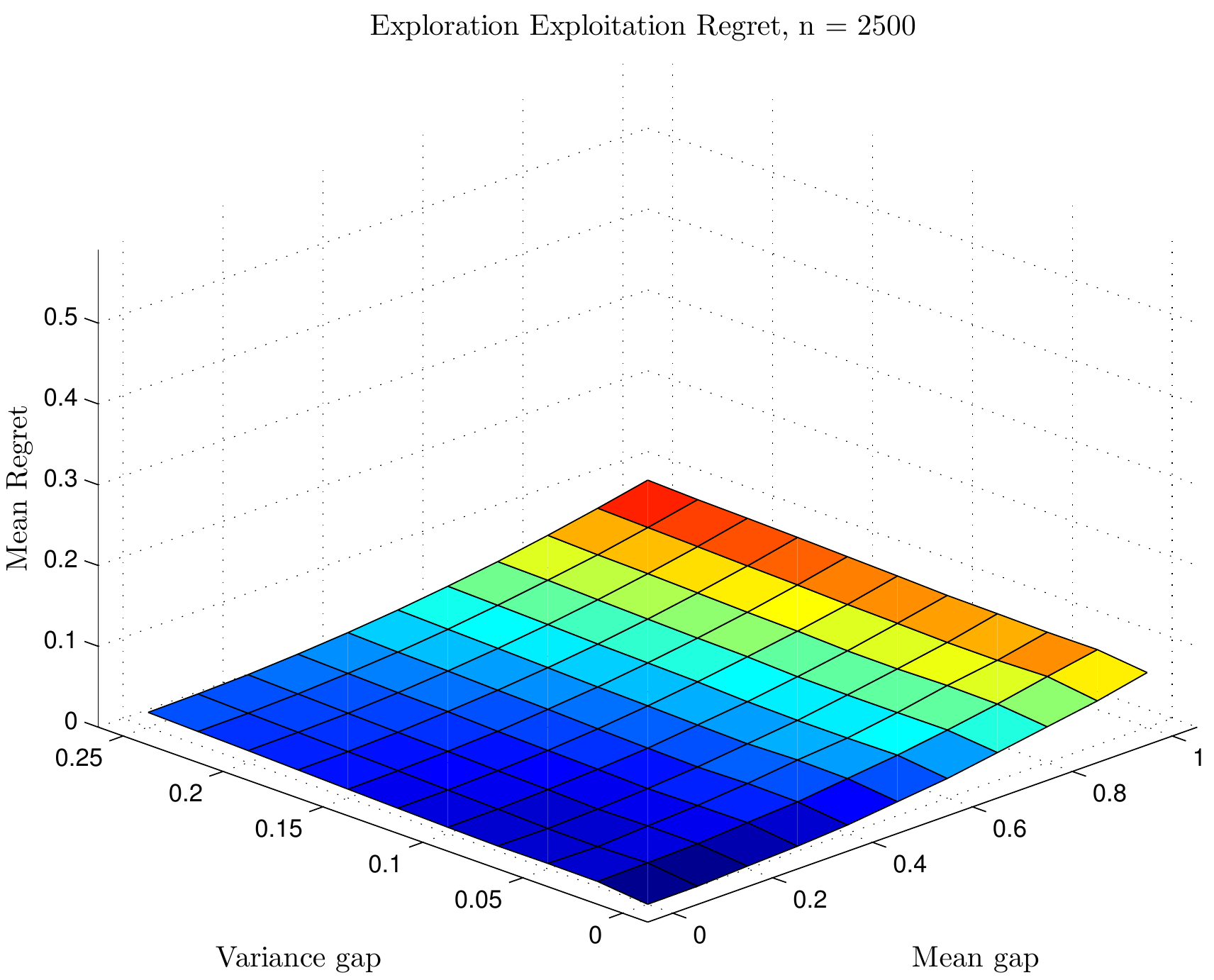}
\includegraphics[width=0.32\textwidth]{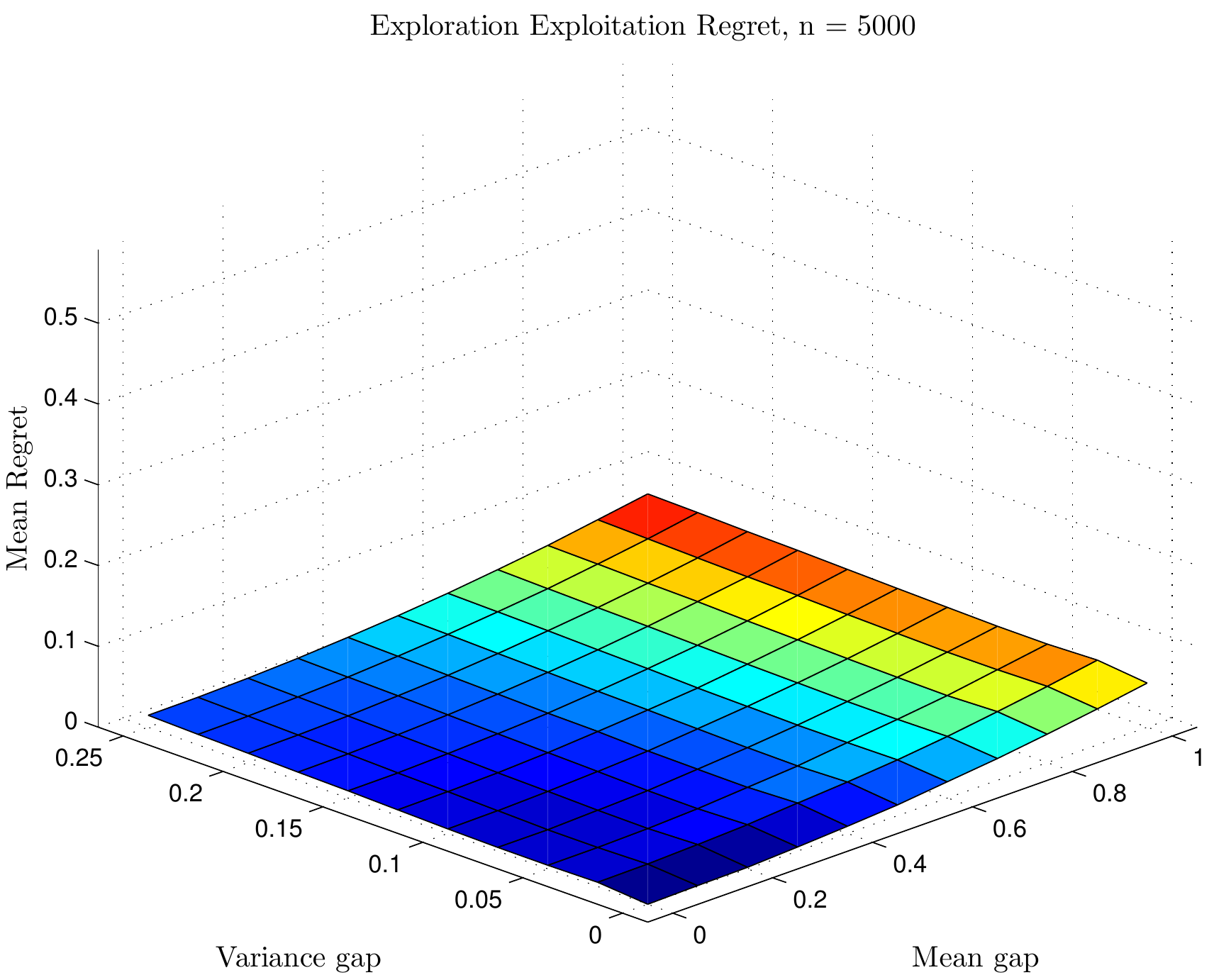}
\includegraphics[width=0.32\textwidth]{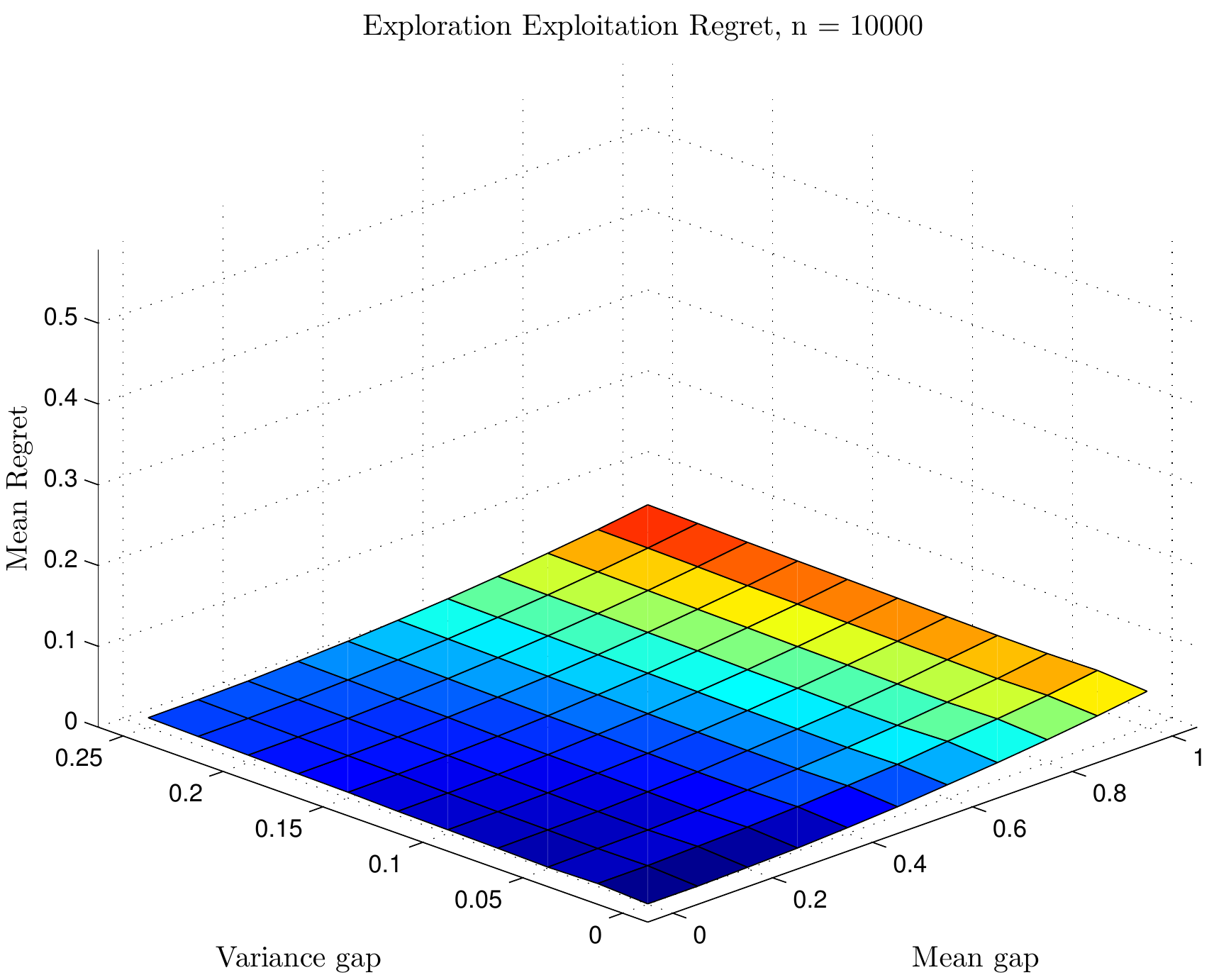}
\includegraphics[width=0.32\textwidth]{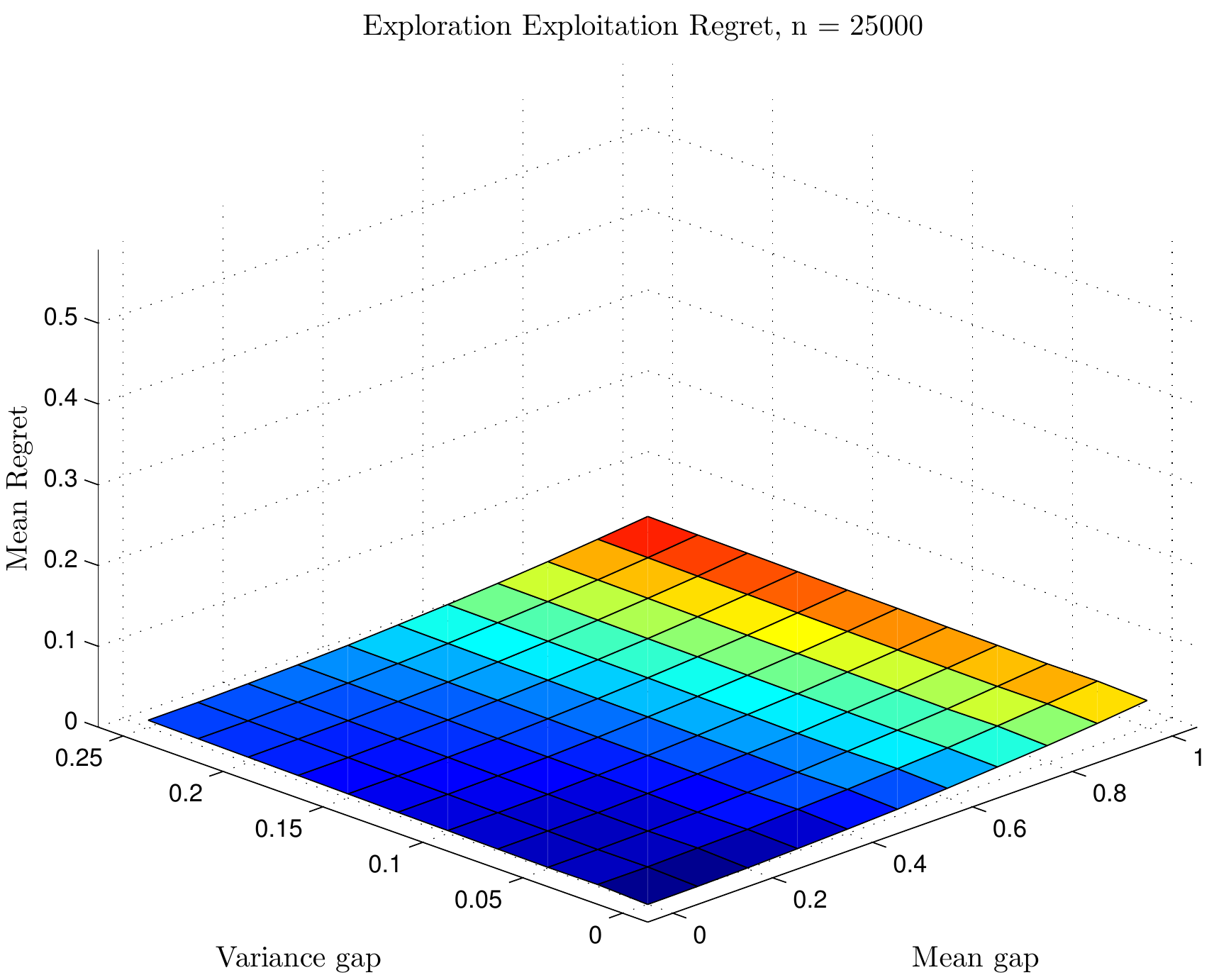}
\includegraphics[width=0.32\textwidth]{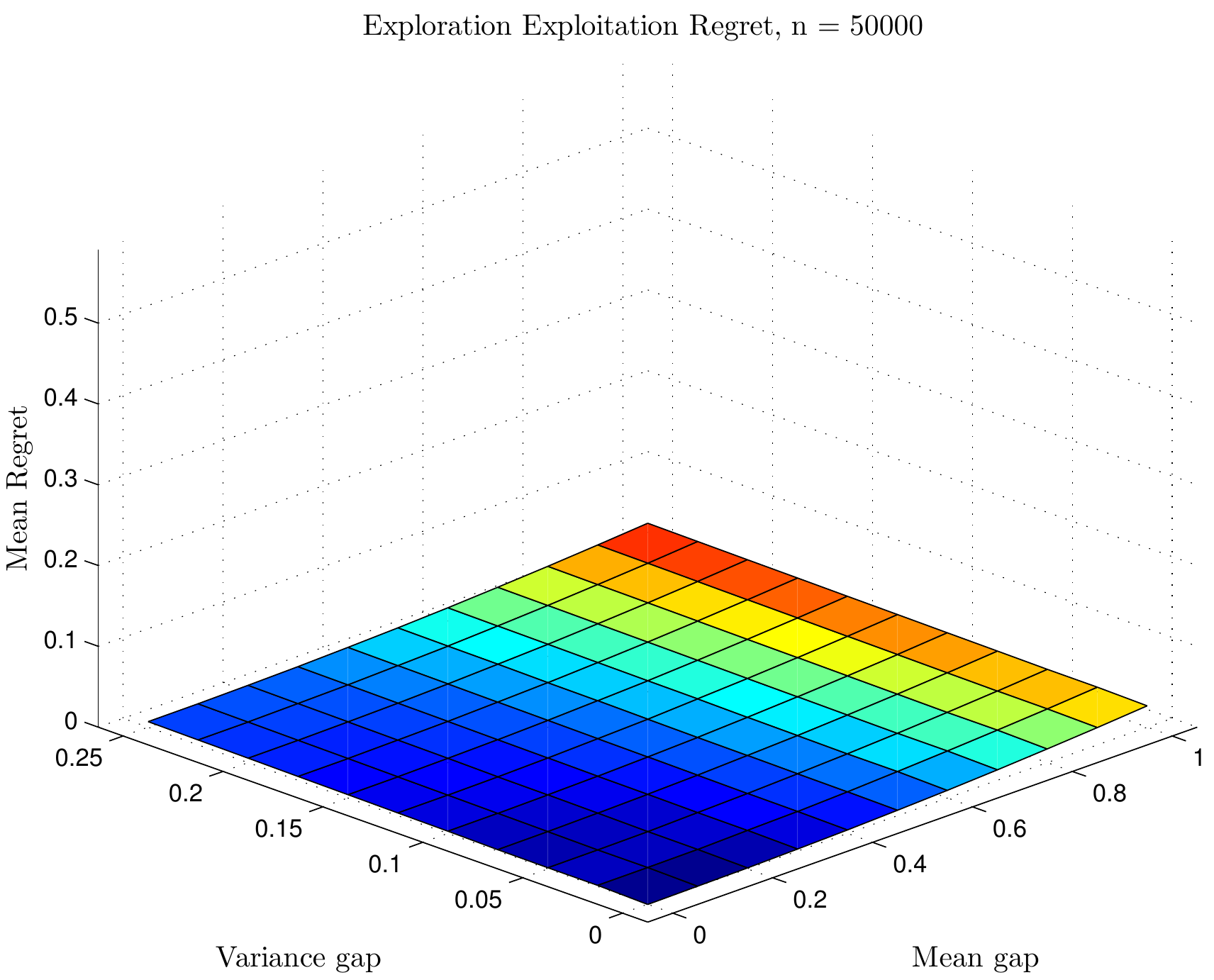}
\includegraphics[width=0.32\textwidth]{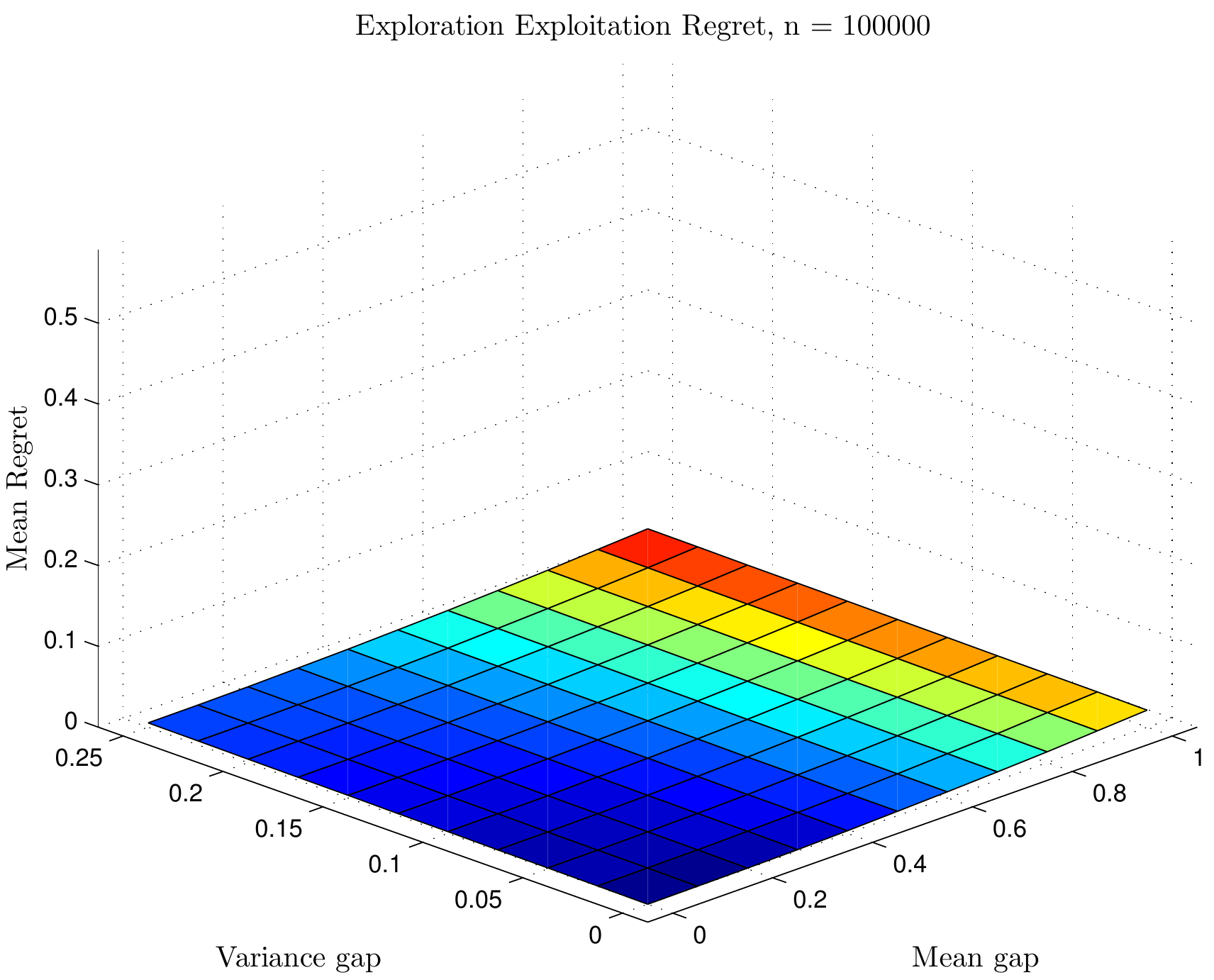}
\includegraphics[width=0.32\textwidth]{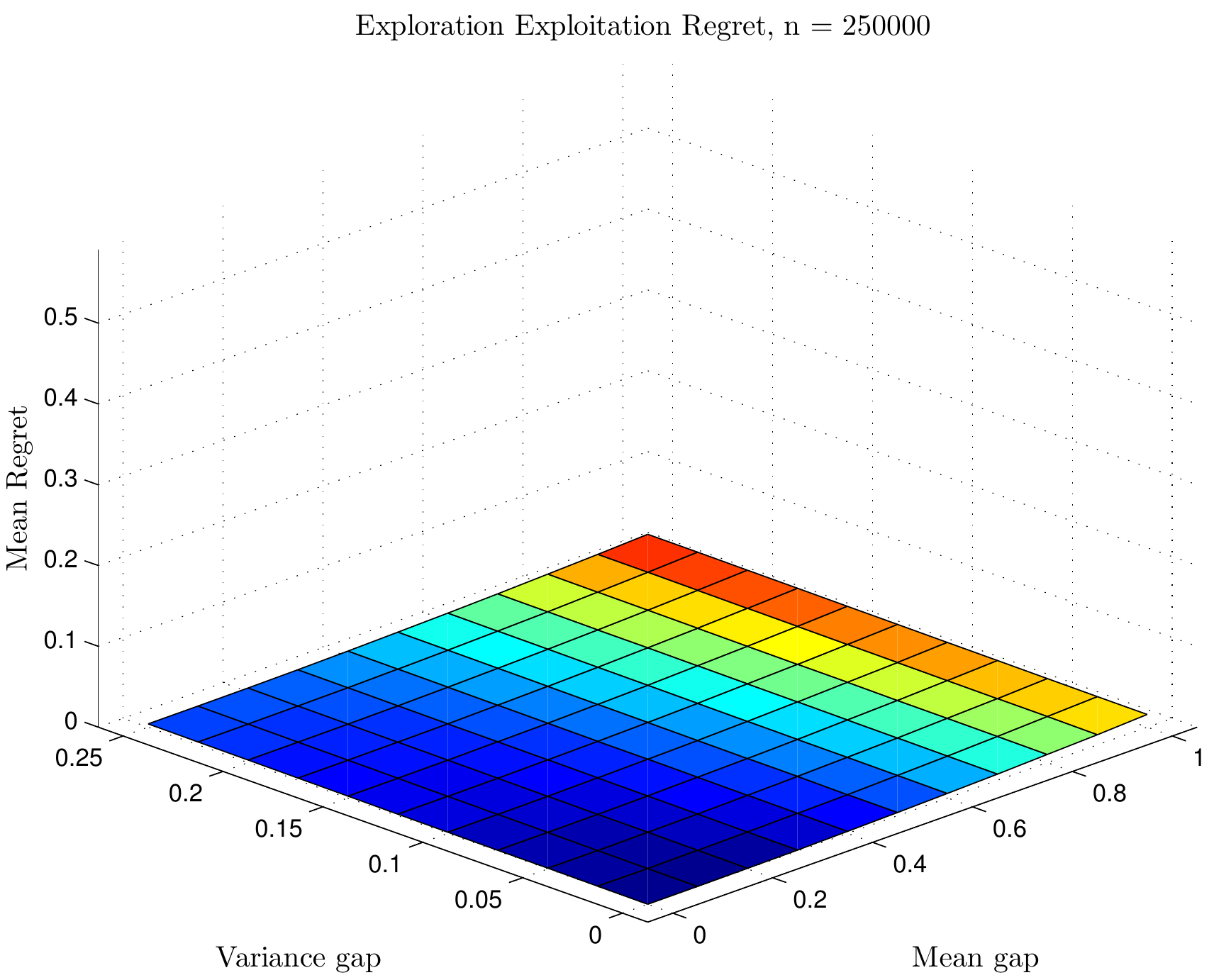}
\end{center}
\vspace{-0.4cm}
\caption{Regret $\R_n$ of \textsl{ExpExp}.}\label{f:expexp.grid}
\vspace{-0.4cm}
\end{figure*}

We consider the variance minimization problem ($\rho=0$) with $K=2$ Gaussian arms with different means and variances. In particular, we consider a grid of values with $\mu_1=1.5$, $\mu_2\in [0.4; 1.5]$, $\var_1\in [0.0; 0.25]$, and $\var_2 = 0.25$, so that $\Delta\in [0.0;0.25]$ and $\Gamma\in [0.0; 1.1]$ and number of rounds $n\in [50; 2.5\times 10^5]$. Figures~\ref{f:mvlcb.grid} and~\ref{f:expexp.grid} report the mean regret for different values of $n$. The colors are renormalized in each plot so that dark blue corresponds to the smallest regret and red to the largest regret. The results confirm the theoretical findings of Theorem~\ref{thm:mvlcb.regret} and~\ref{thm:ee.regret}. In fact, for simple problems (large gaps $\Delta$) \textsl{MV-LCB} converges to a zero--regret faster than \textsl{ExpExp}, while for $\Delta$ close to zero (i.e., equivalent arms), \textsl{MV-LCB} has a constant regret which does not decrease with $n$ and the regret of \textsl{ExpExp} slowly decreases to zero.

%%% Grid comparison
\subsection{Risk tolerance sensitivity}

\begin{figure}[ht]
\begin{scriptsize}
\begin{center}
\begin{minipage}{0.25\textwidth}
\begin{center}
\begin{tabular}{|r|c|c|}
\hline
Arm & $\mu$ & $\var$ \\
 \hline
 \hline
$\alpha_{1}$ &  0.10 & 0.05 \\
$\alpha_{2}$ & 0.20 & 0.34 \\
$\alpha_{3}$ & 0.23 & 0.28 \\
$\alpha_{4}$ & 0.27 & 0.09 \\
$\alpha_{5}$ & 0.32 & 0.23 \\
$\alpha_{6}$ & 0.32 & 0.72 \\
$\alpha_{7}$ & 0.34 & 0.19 \\
$\alpha_{8}$ & 0.41 & 0.14 \\
$\alpha_{9}$ & 0.43 & 0.44 \\
$\alpha_{10}$ & 0.54 & 0.53 \\
$\alpha_{11}$ & 0.55 & 0.24 \\
$\alpha_{12}$ & 0.56 & 0.36 \\
$\alpha_{13}$ & 0.67 & 0.56 \\
$\alpha_{14}$ & 0.71 & 0.49 \\
$\alpha_{15}$ & 0.79 & 0.85 \\
\hline
\end{tabular}
\end{center}
\end{minipage}
\begin{minipage}{0.25\textwidth}
\begin{center}
\begin{tabular}{|r|c|c|}
\hline
Arm & $\mu$ & $\var$ \\
 \hline
 \hline
$\alpha_{1}$ & 0.1 & 0.05 \\
$\alpha_{2}$ & 0.2 & 0.0725 \\
$\alpha_{3}$ & 0.27 & 0.09 \\
$\alpha_{4}$ & 0.32 & 0.11 \\
$\alpha_{5}$ & 0.41 & 0.145 \\
$\alpha_{6}$ & 0.49 & 0.19 \\
$\alpha_{7}$ & 0.55 & 0.24 \\
$\alpha_{8}$ & 0.59 & 0.28 \\
$\alpha_{9}$ & 0.645 & 0.36 \\
$\alpha_{10}$ & 0.678 & 0.413 \\
$\alpha_{11}$ & 0.69 & 0.445 \\
$\alpha_{12}$ & 0.71 & 0.498 \\
$\alpha_{13}$ & 0.72 & 0.53 \\
$\alpha_{14}$ & 0.765 & 0.72 \\
$\alpha_{15}$ & 0.79 & 0.854 \\
\hline
\end{tabular}
\end{center}
\end{minipage}
\begin{minipage}{0.4\textwidth}
\begin{center}
\includegraphics[width=1.0\textwidth]{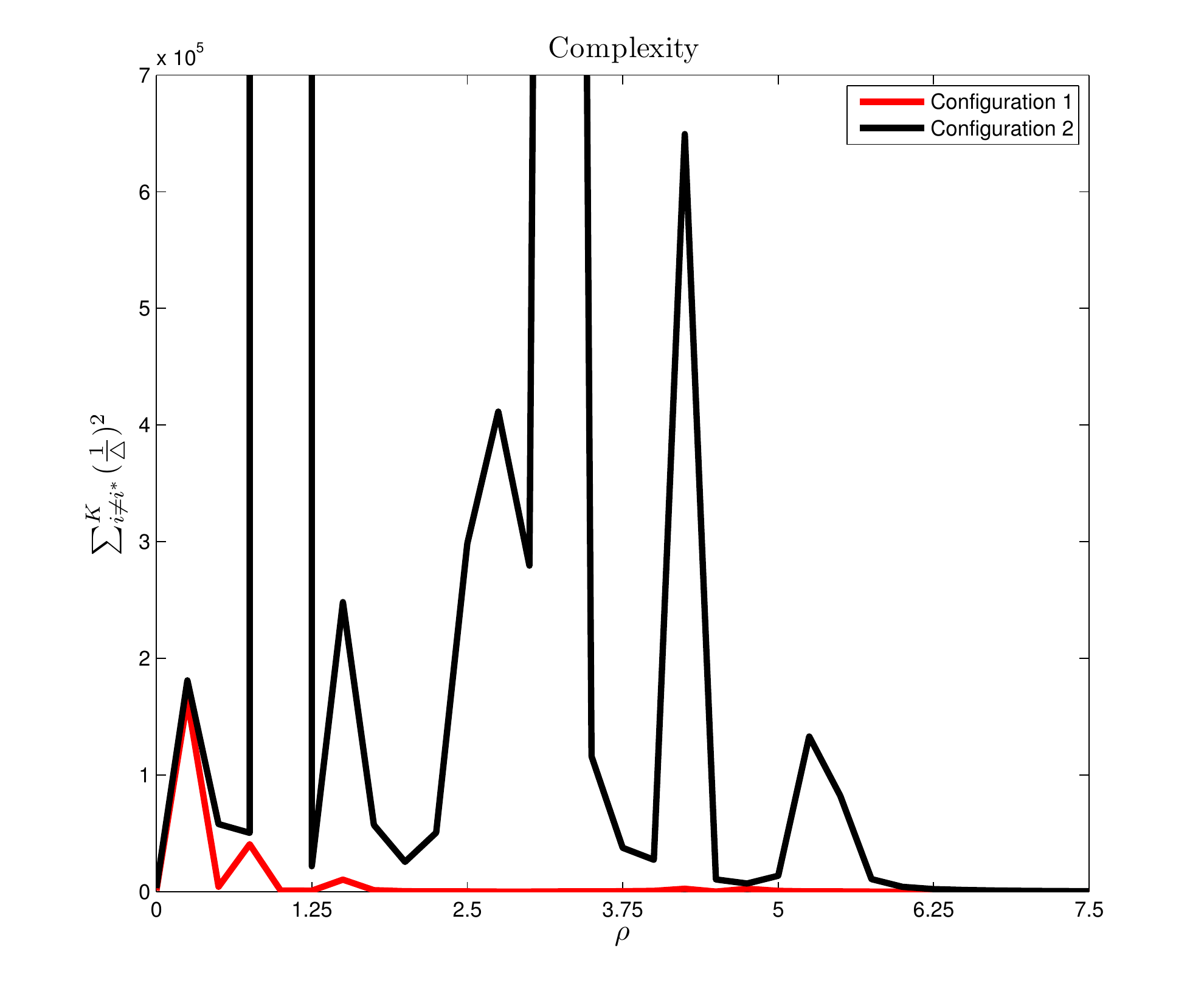}
\end{center}
\end{minipage}
\end{center}
\end{scriptsize}
\caption{Configuration 1 and Configuration 2 and their corresponding complexity.}\label{f:config}
\end{figure}

\begin{figure*}[h]
\begin{center}
\includegraphics[width=0.32\textwidth]{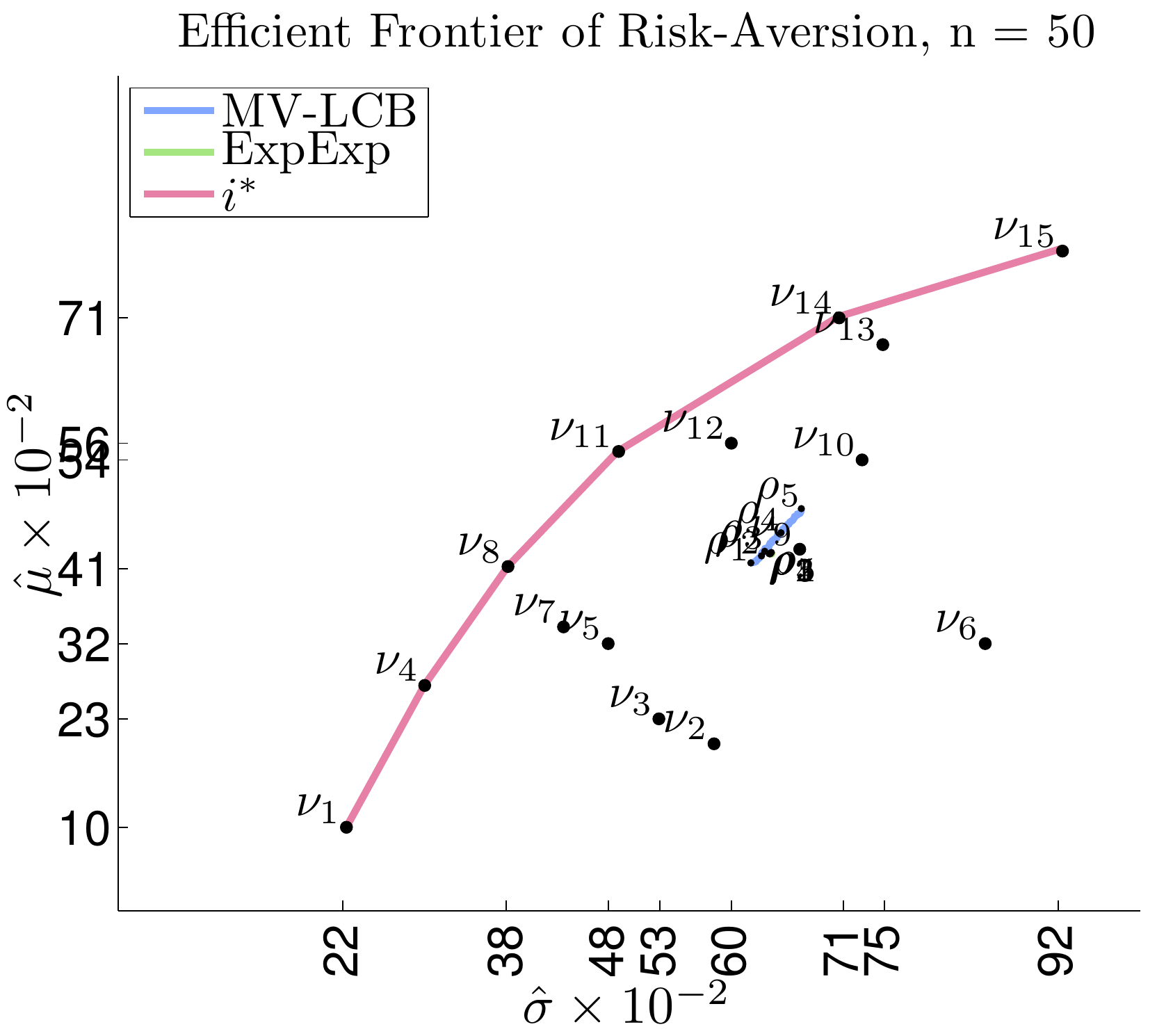}
\includegraphics[width=0.32\textwidth]{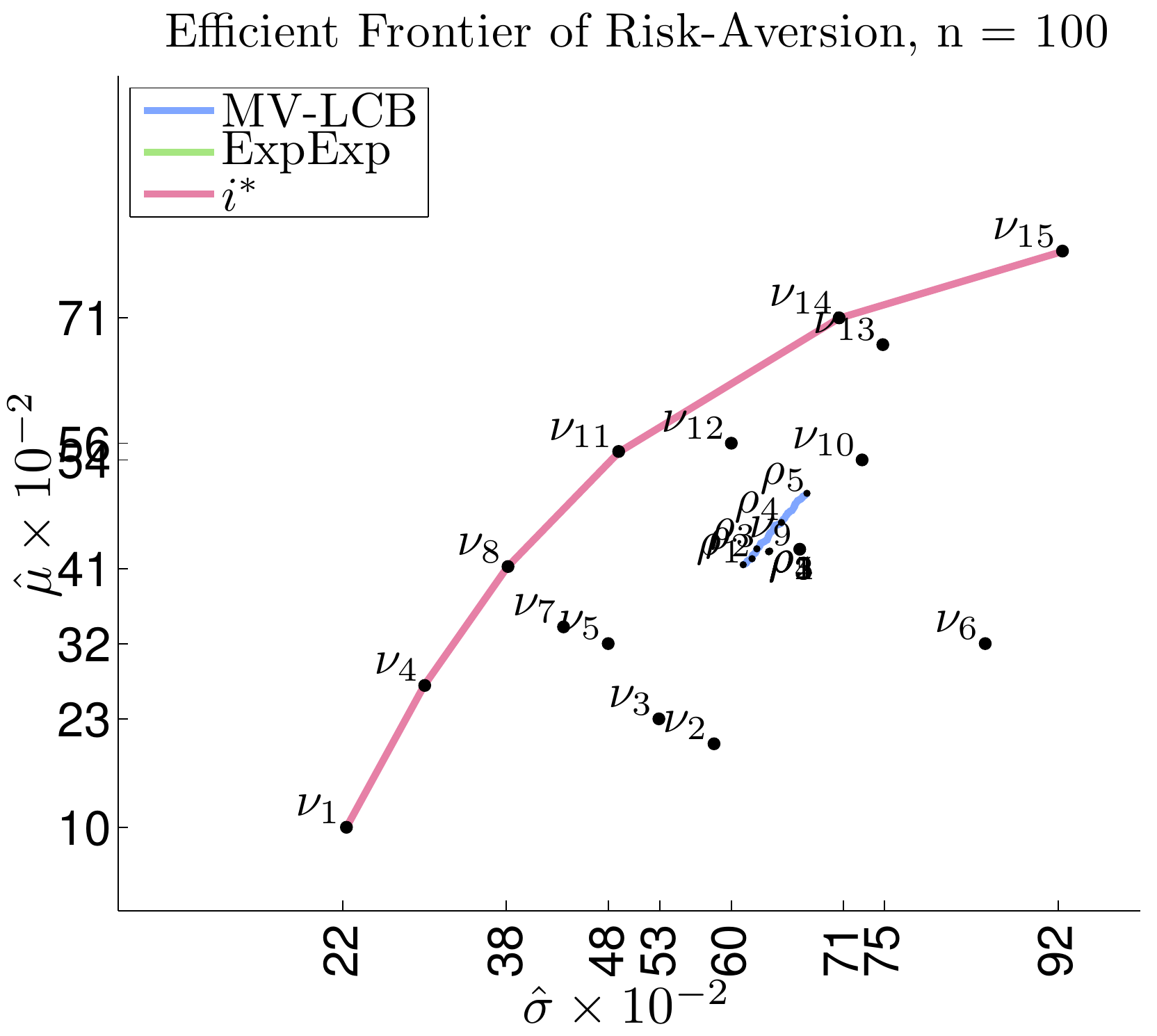}
\includegraphics[width=0.32\textwidth]{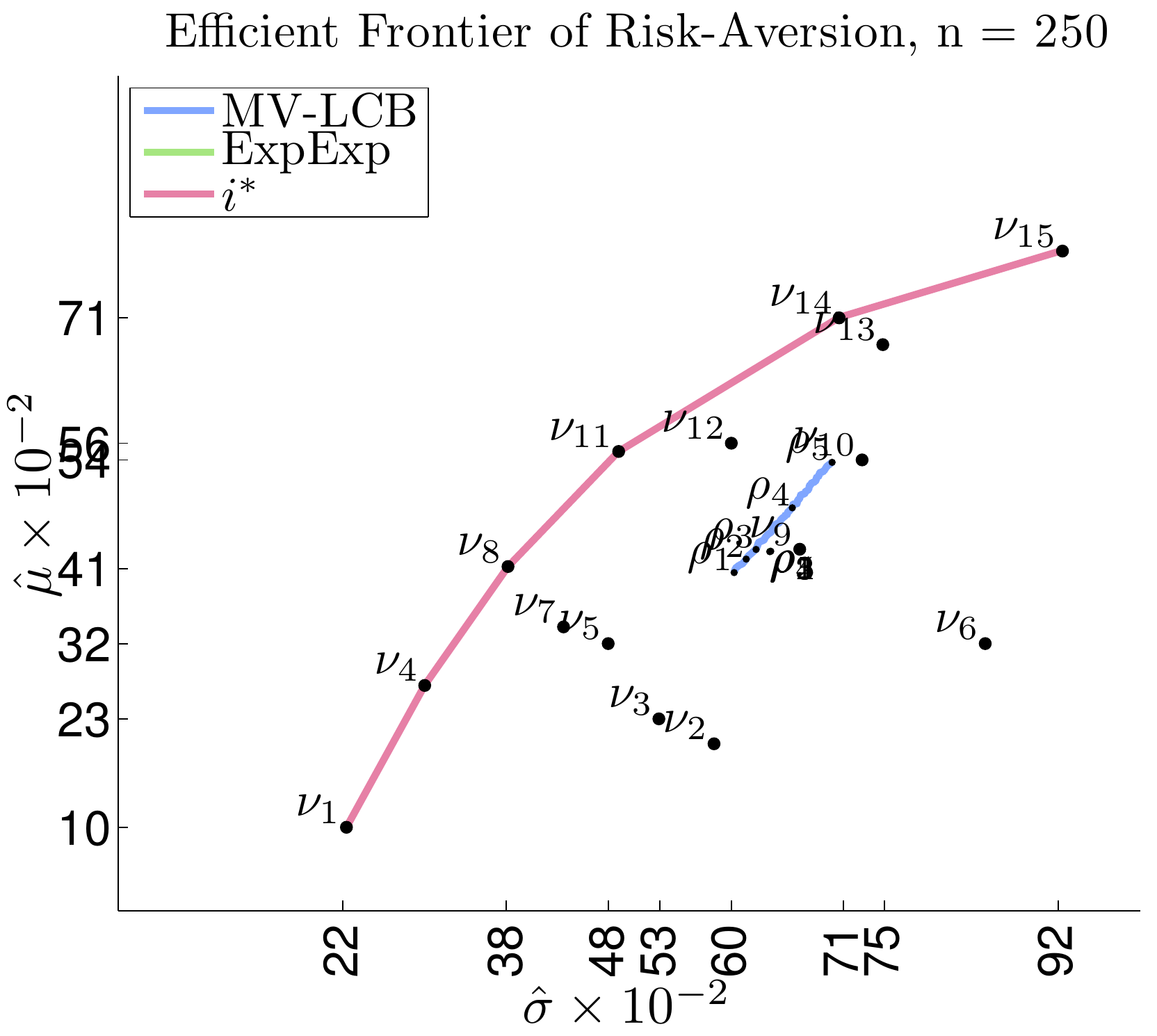}
\includegraphics[width=0.32\textwidth]{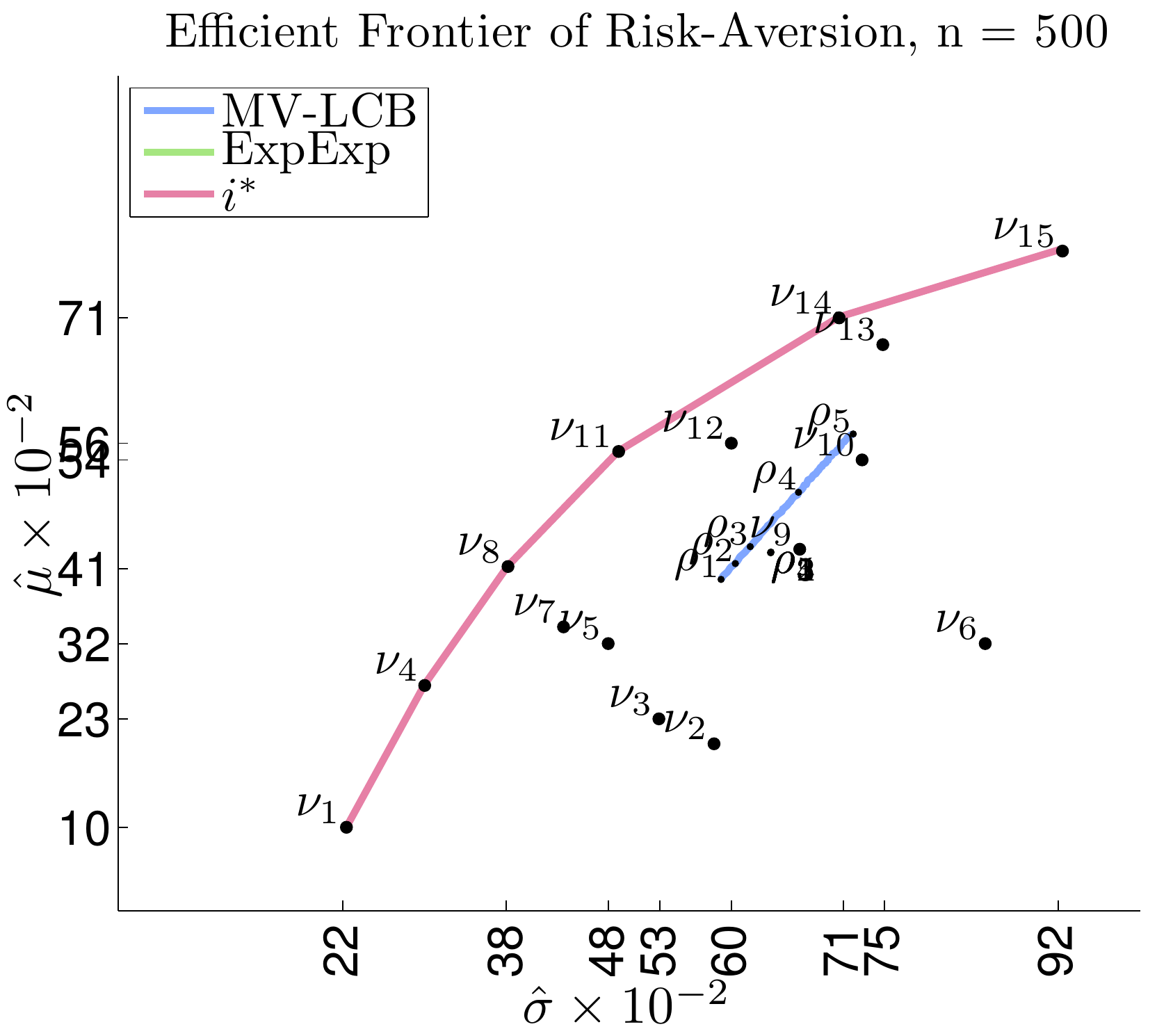}
\includegraphics[width=0.32\textwidth]{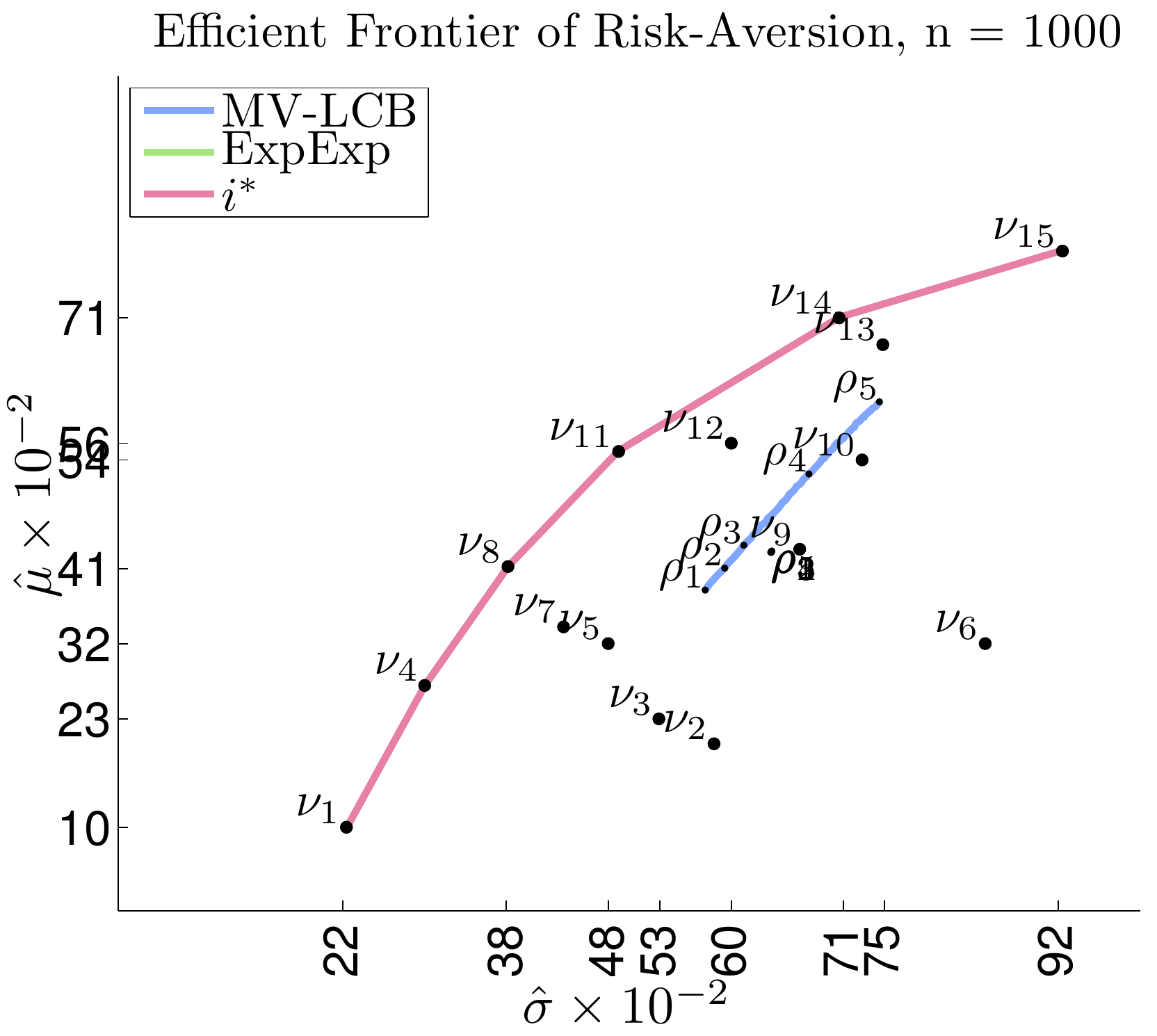}
\includegraphics[width=0.32\textwidth]{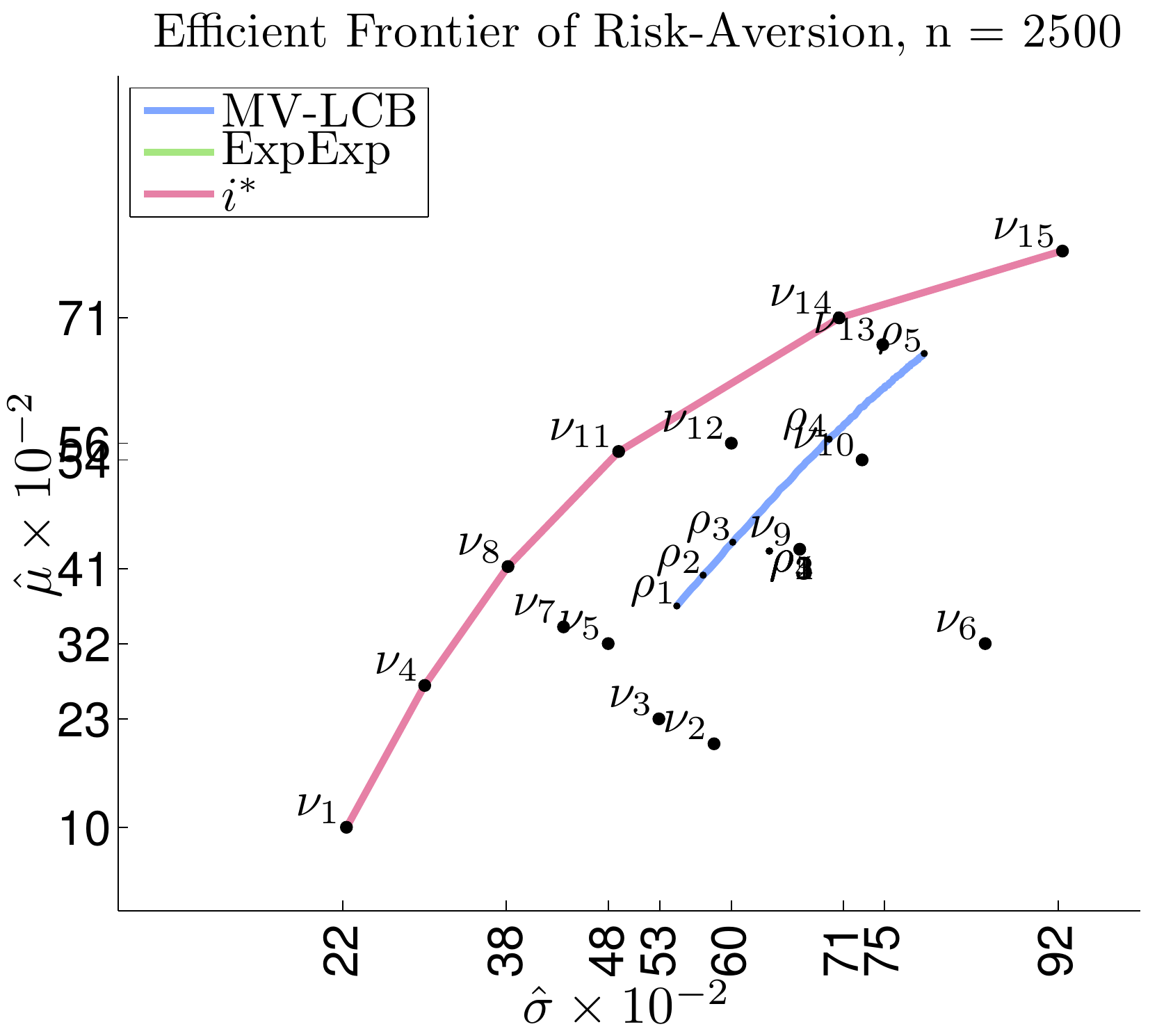}
\includegraphics[width=0.32\textwidth]{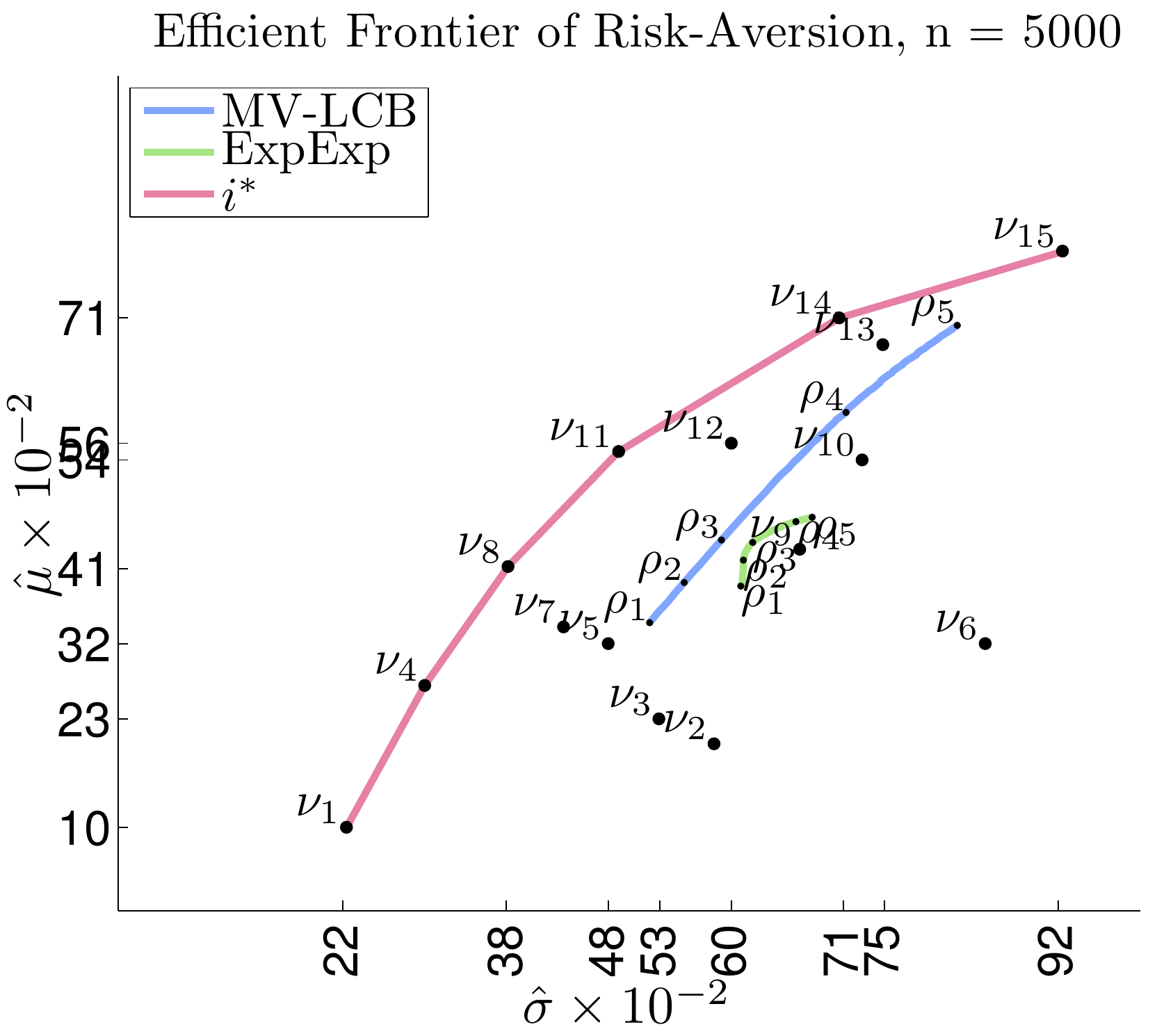}
\includegraphics[width=0.32\textwidth]{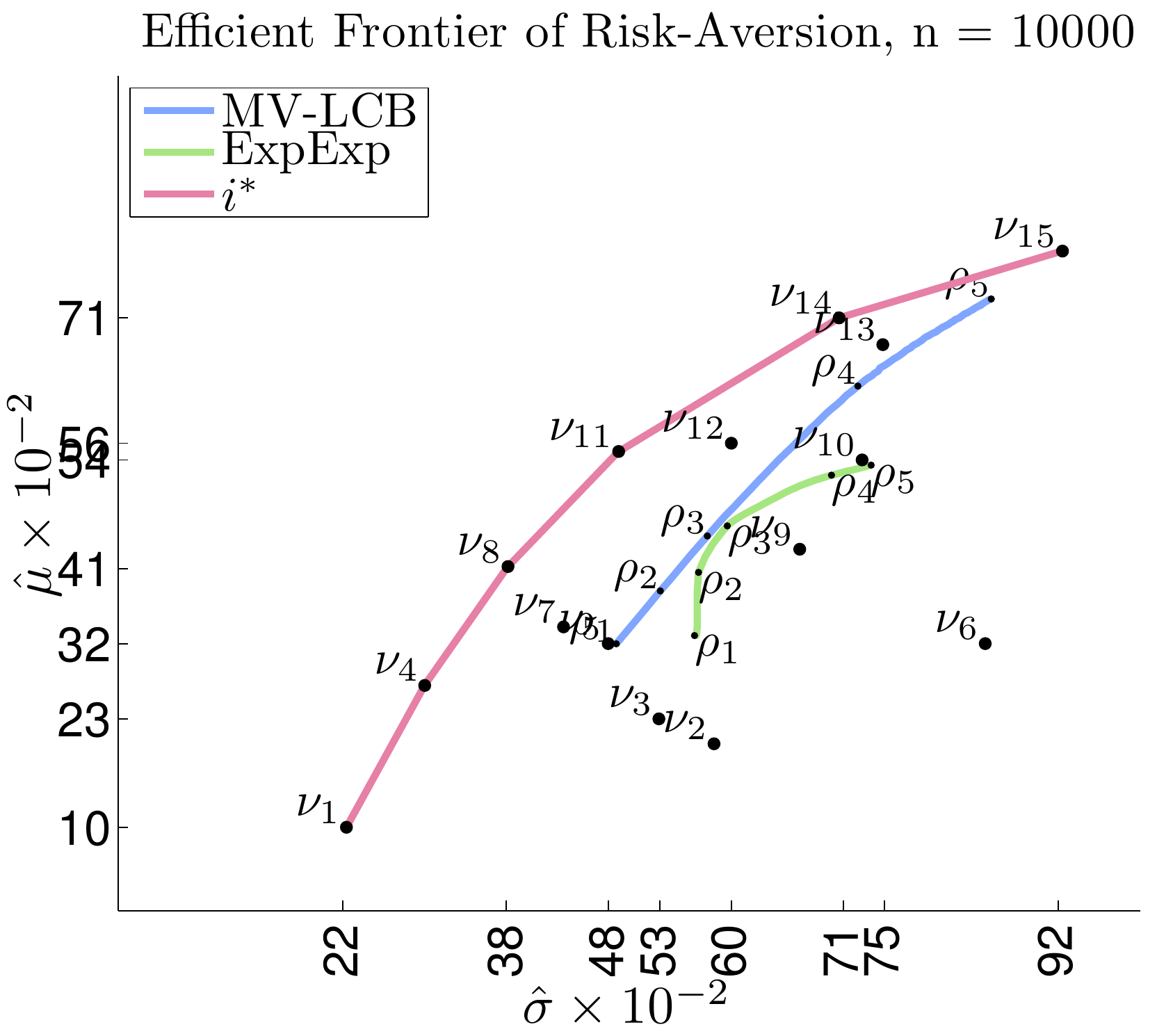}
\includegraphics[width=0.32\textwidth]{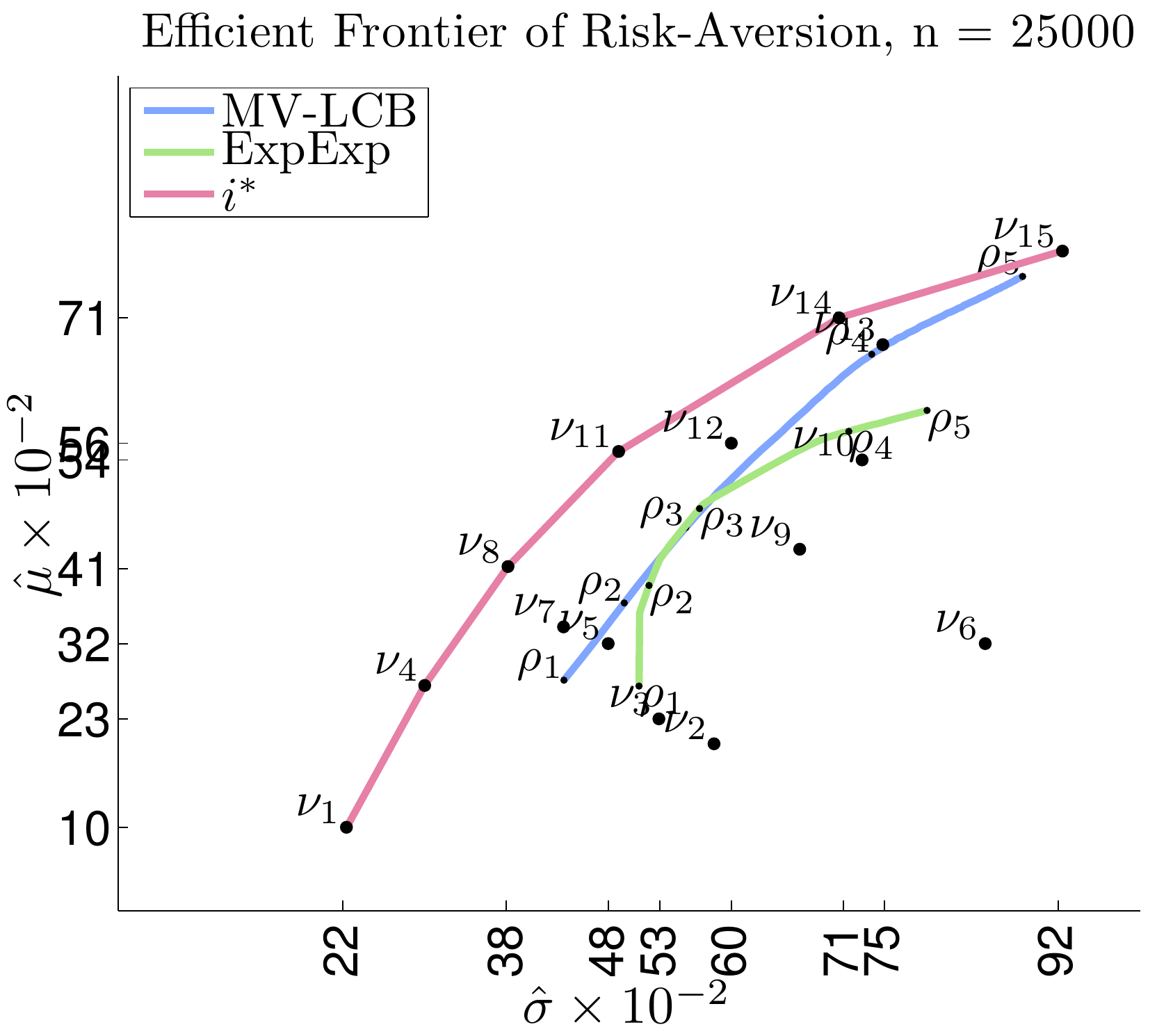}
\includegraphics[width=0.32\textwidth]{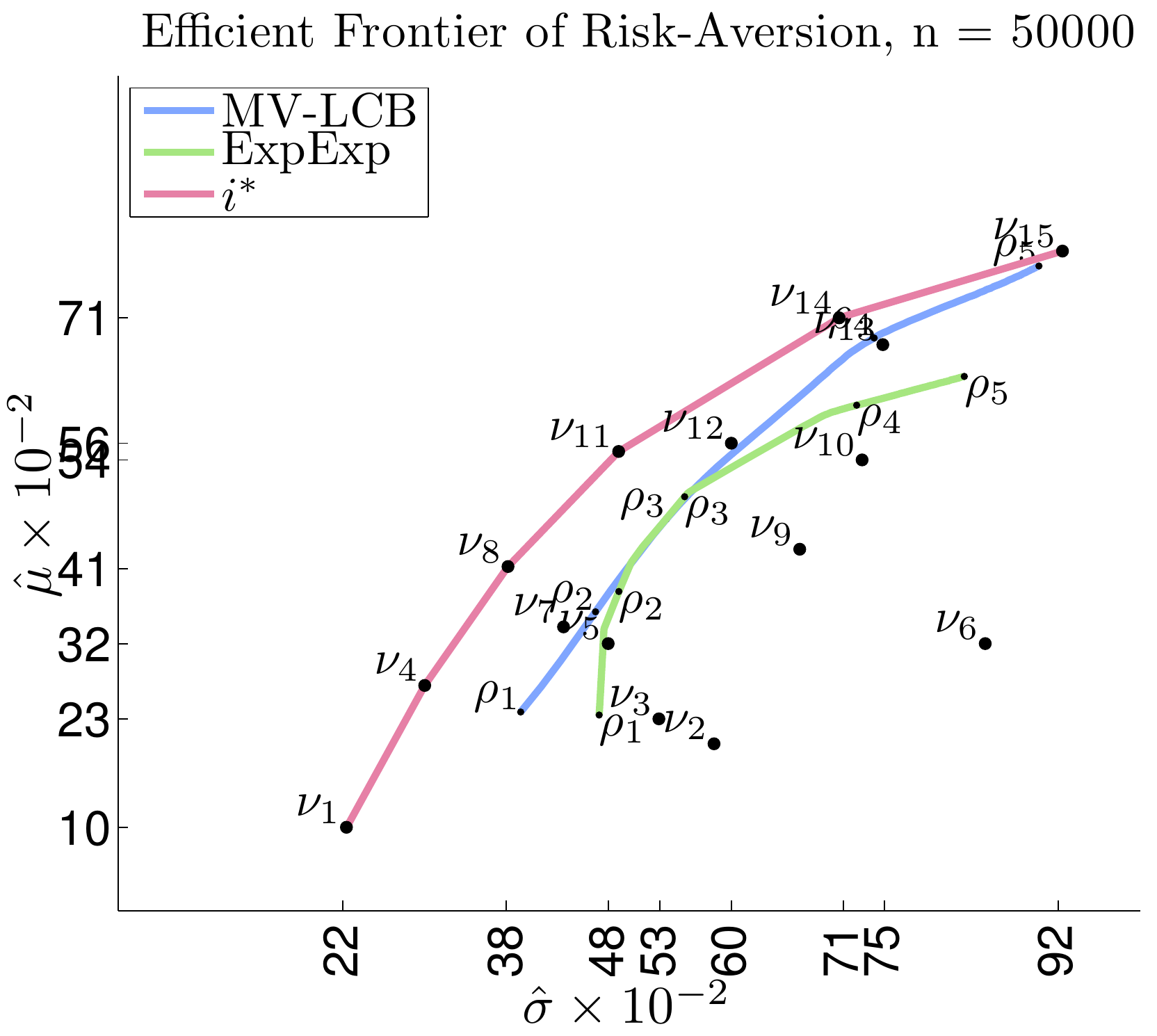}
\includegraphics[width=0.32\textwidth]{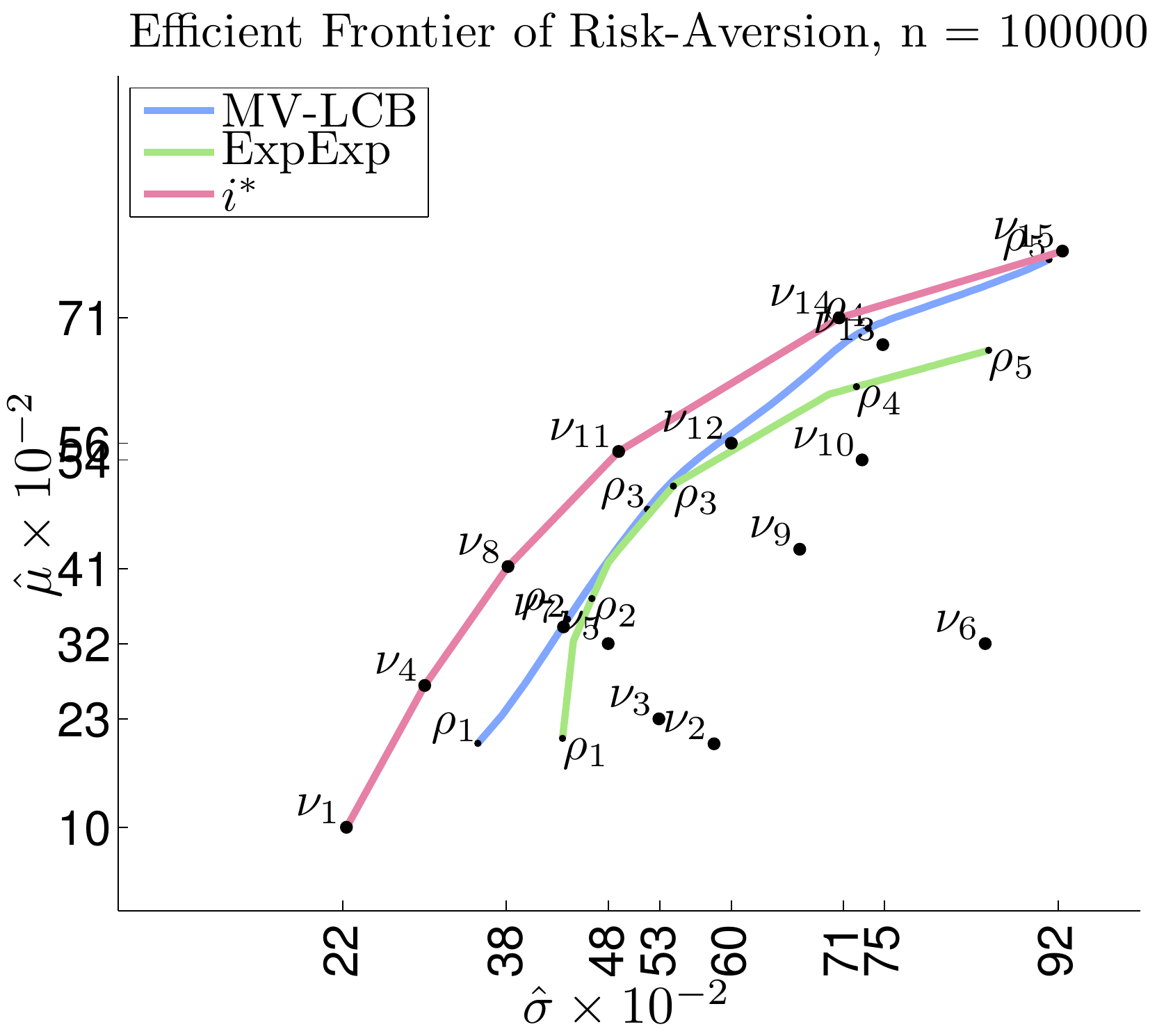}
\includegraphics[width=0.32\textwidth]{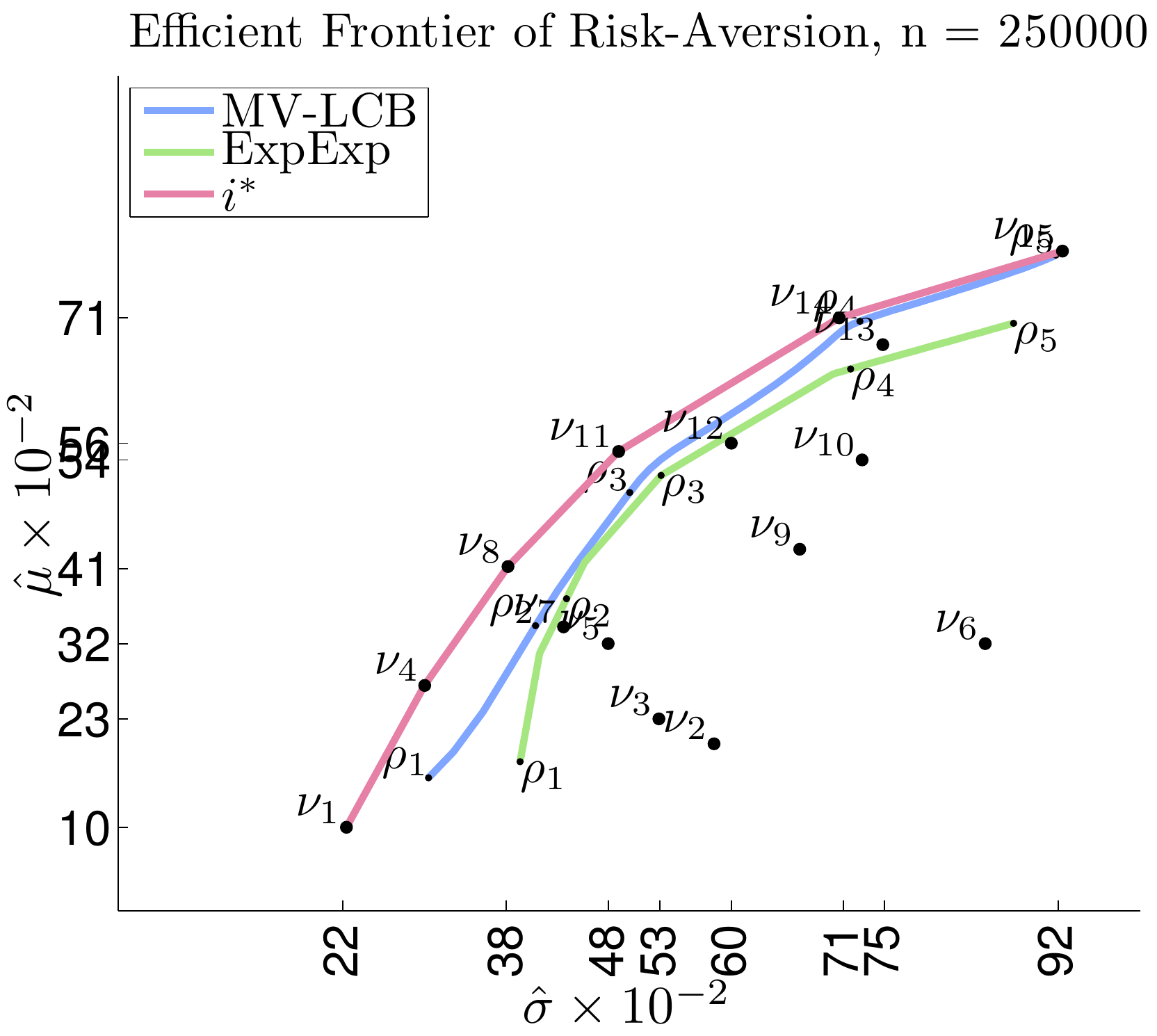}
\end{center}
\vspace{-0.4cm}
\caption{Risk tolerance sensitivity of \textsl{MV-LCB} and \textsl{ExpExp} for \textit{Configuration 1}.}\label{f:ef.easy}
\vspace{-0.4cm}
\end{figure*}

\begin{figure*}[ht]
\begin{center}
\includegraphics[width=0.32\textwidth]{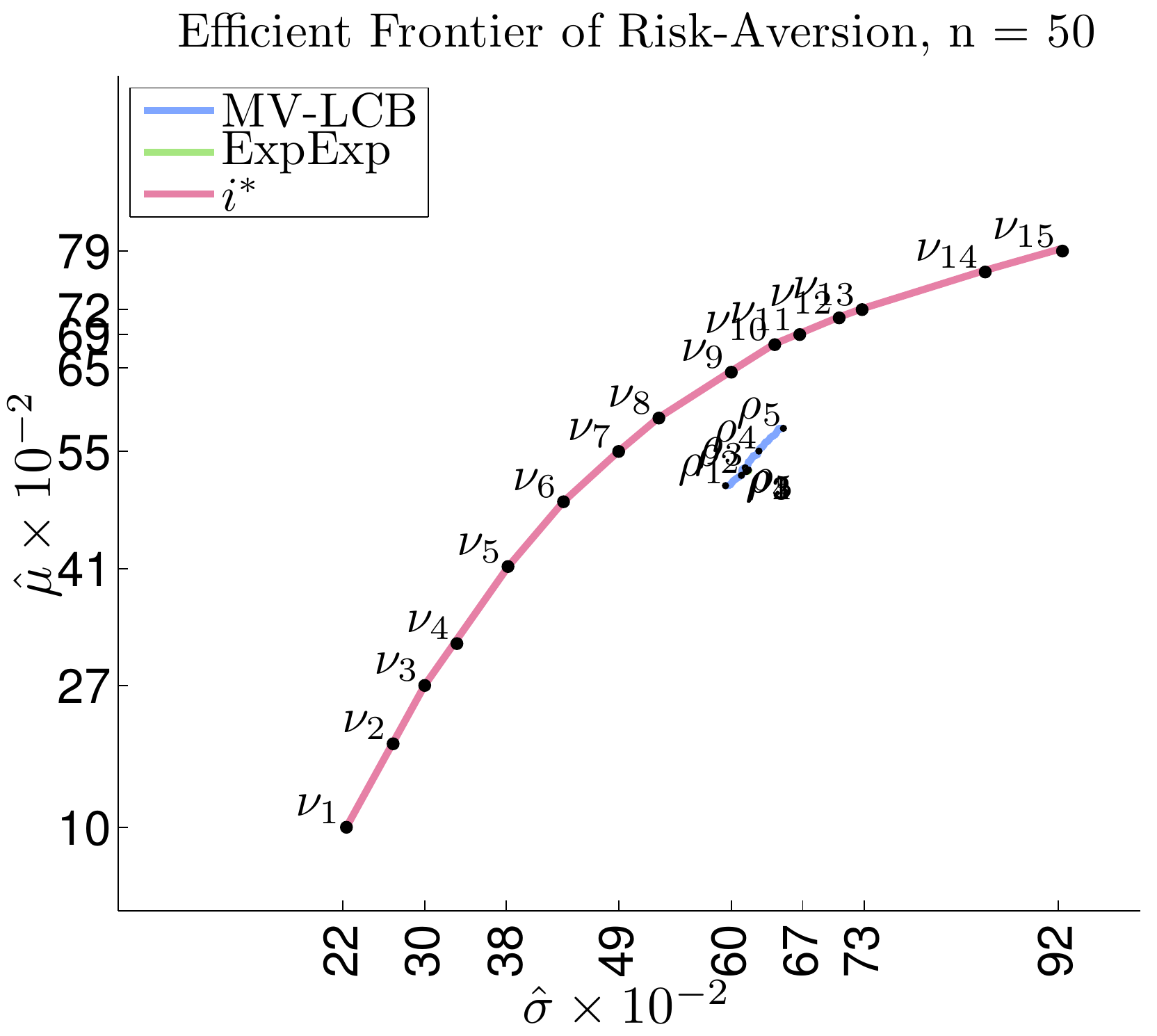}
\includegraphics[width=0.32\textwidth]{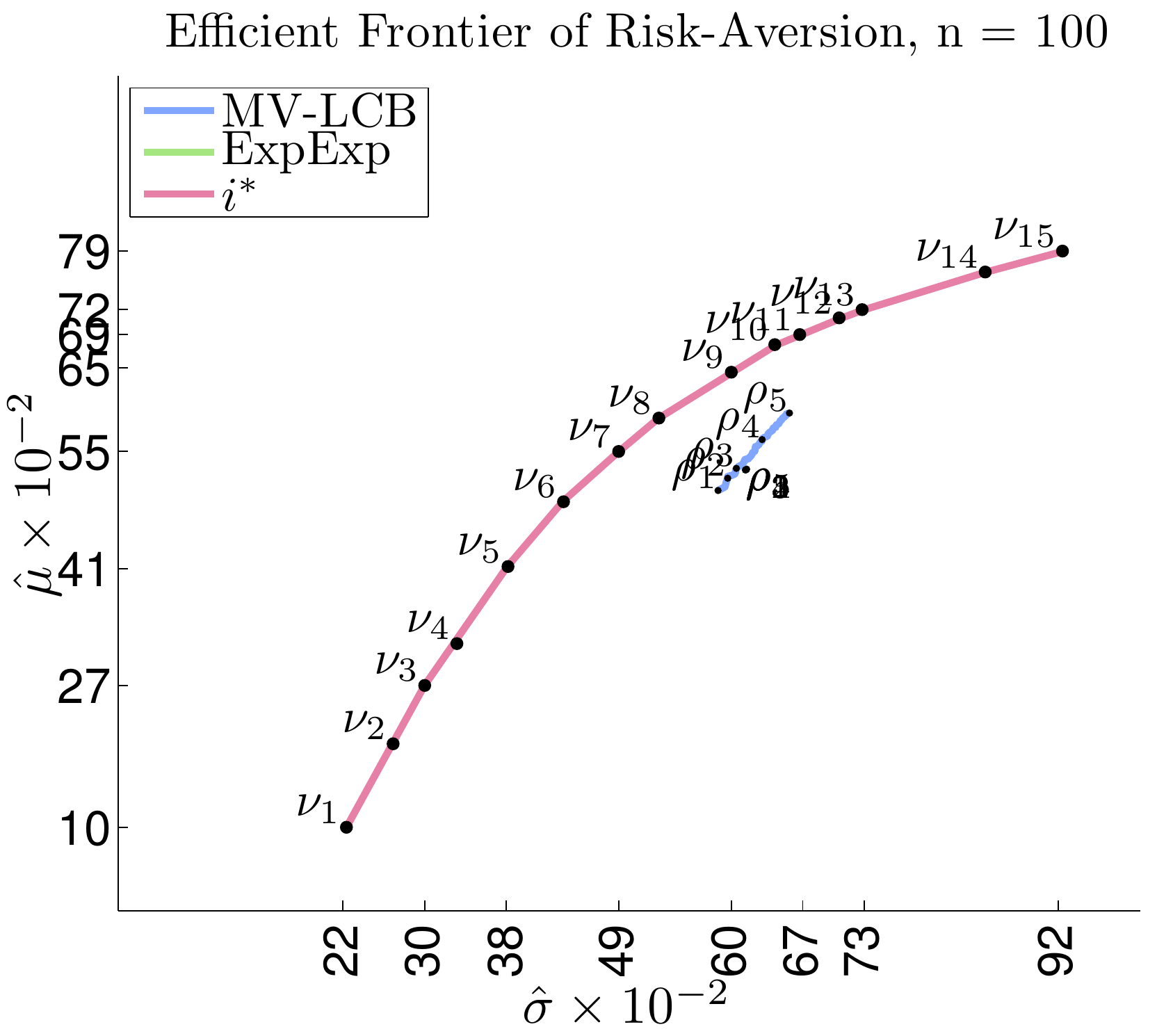}
\includegraphics[width=0.32\textwidth]{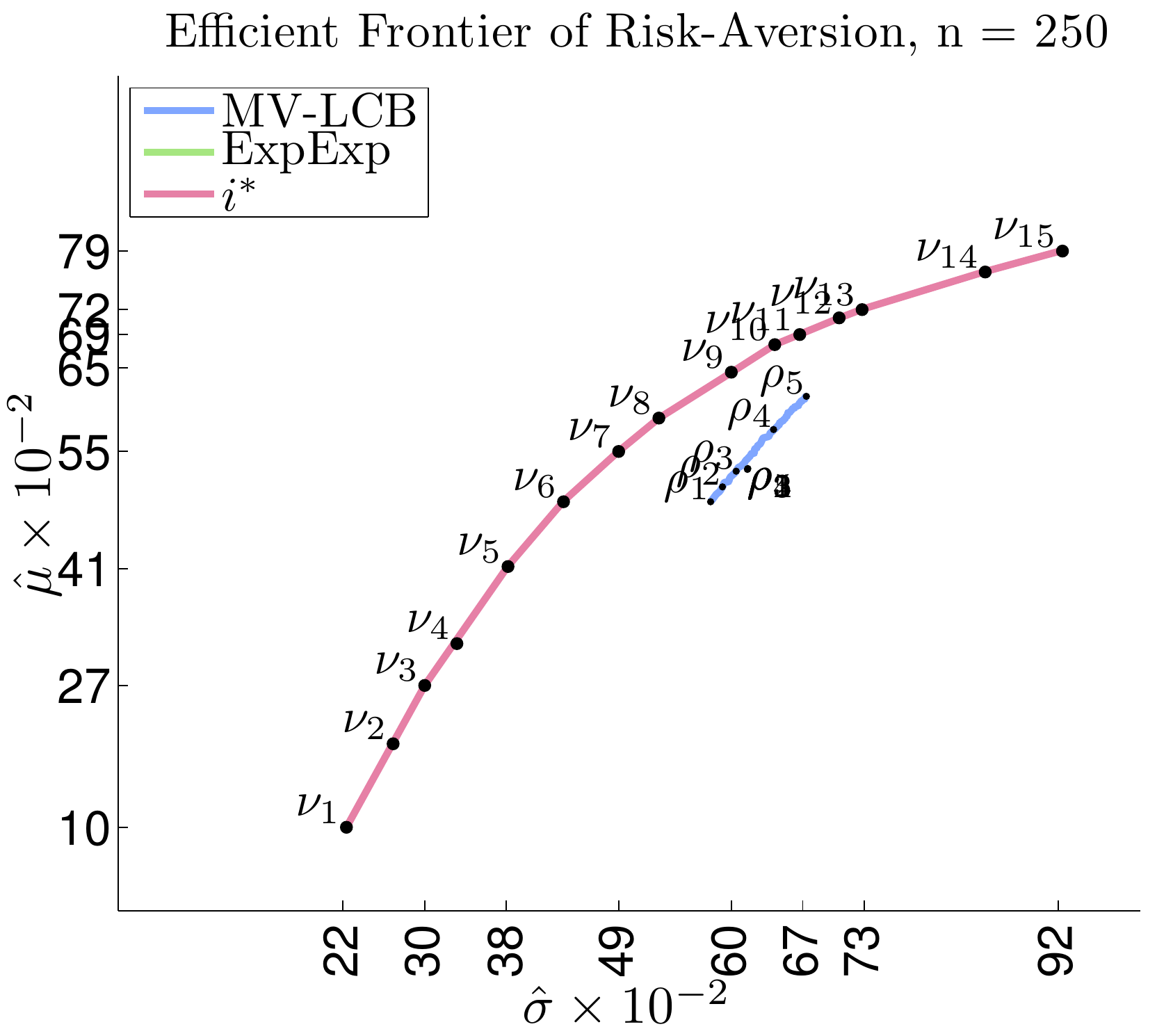}
\includegraphics[width=0.32\textwidth]{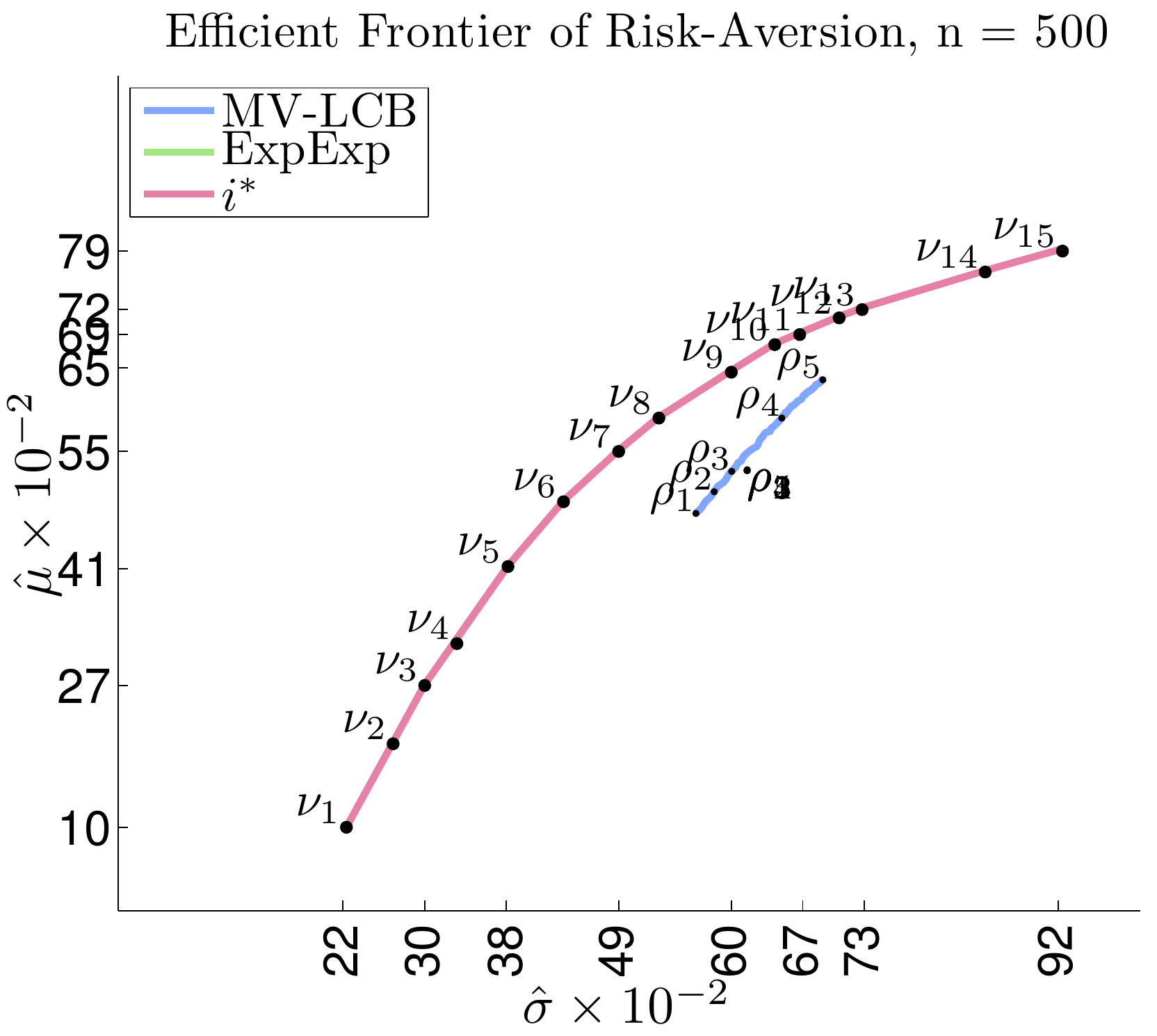}
\includegraphics[width=0.32\textwidth]{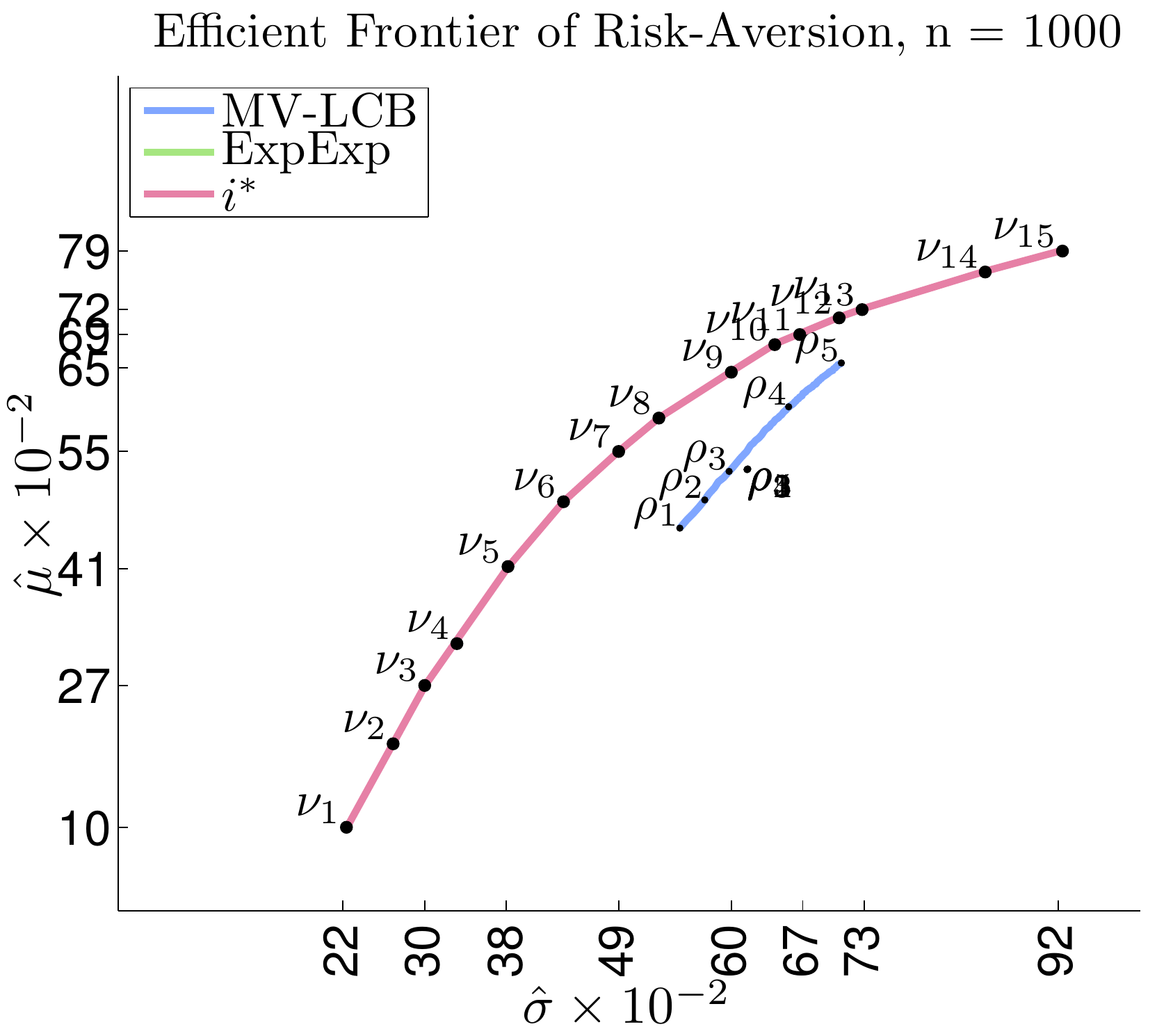}
\includegraphics[width=0.32\textwidth]{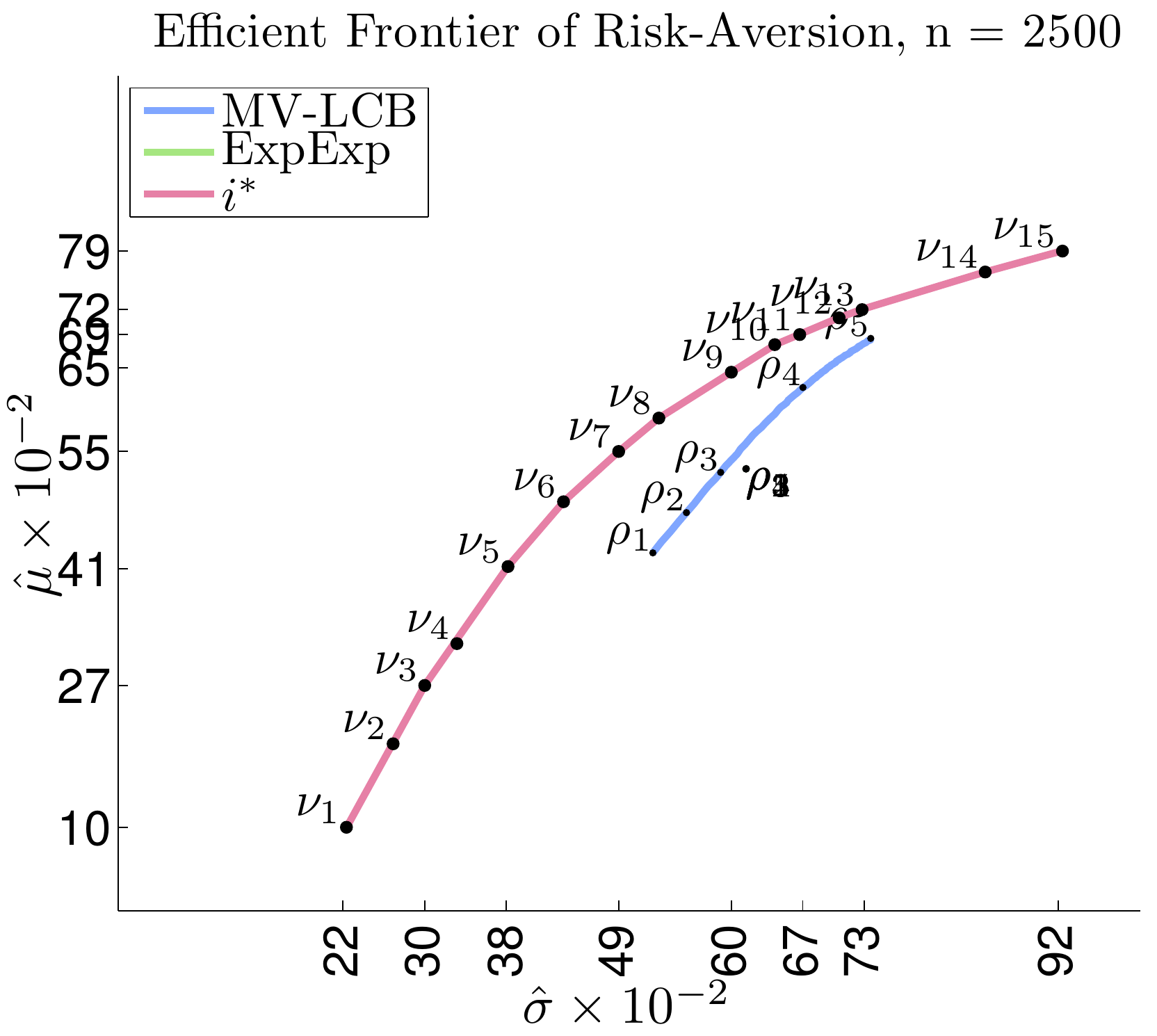}
\includegraphics[width=0.32\textwidth]{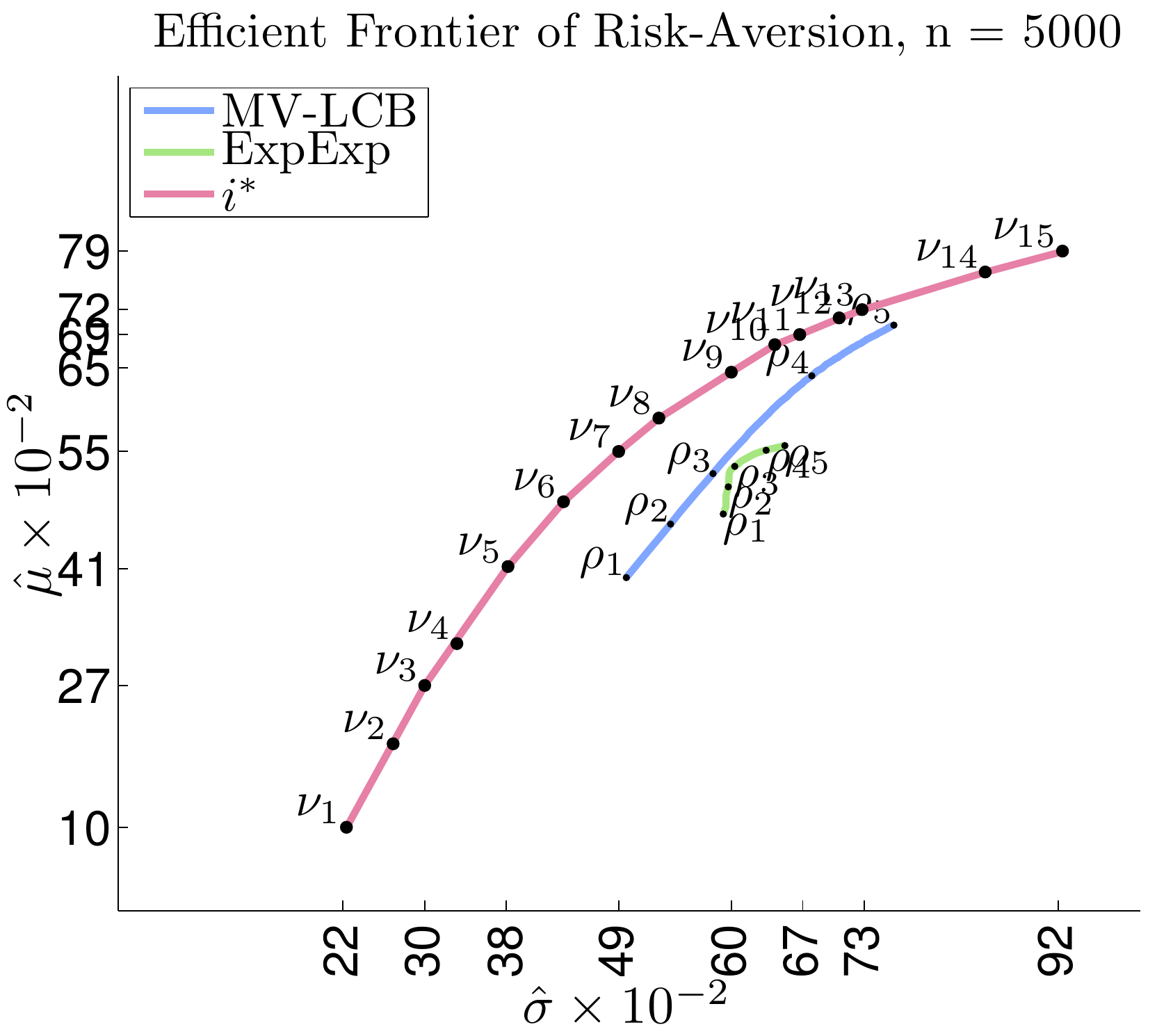}
\includegraphics[width=0.32\textwidth]{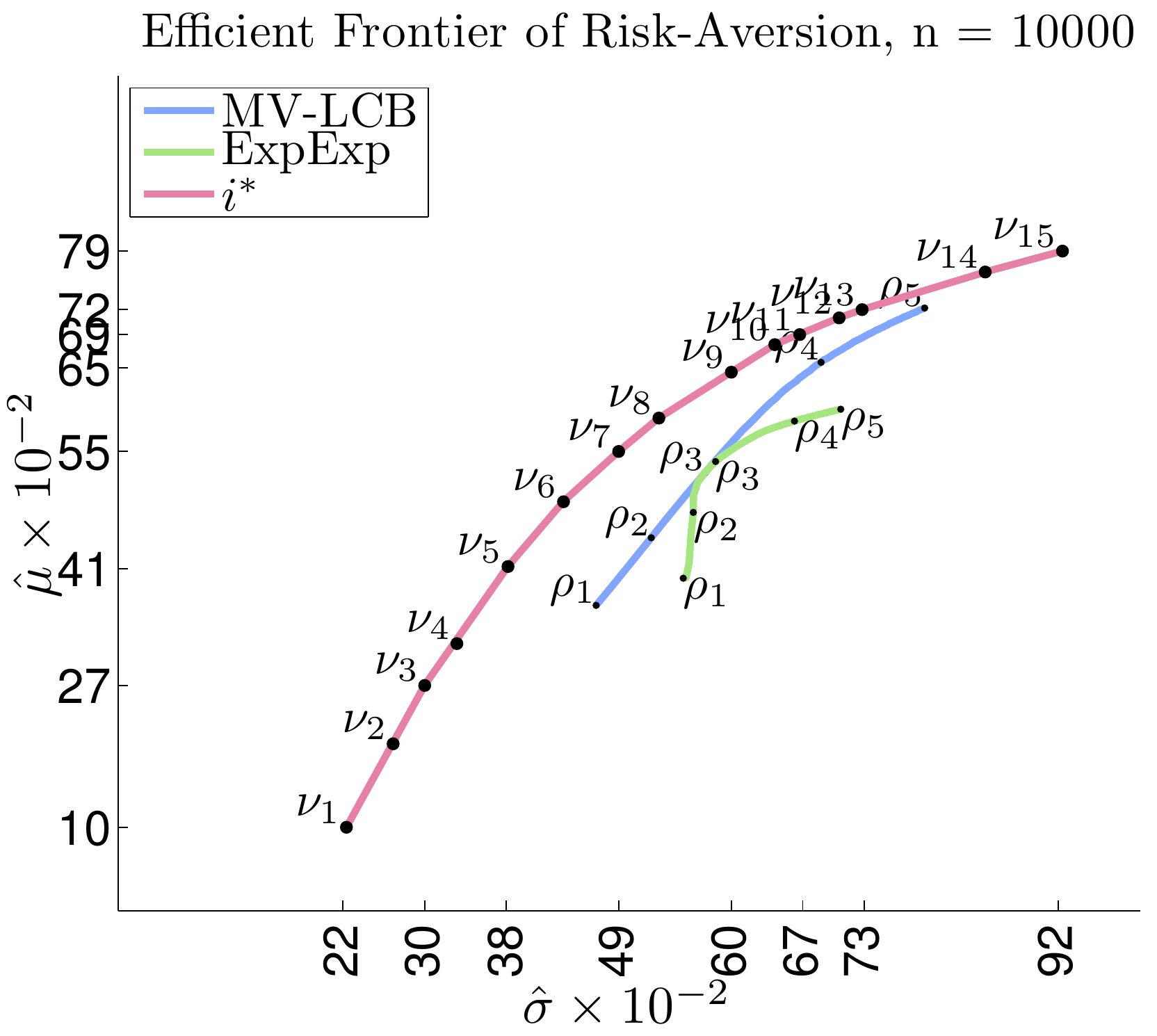}
\includegraphics[width=0.32\textwidth]{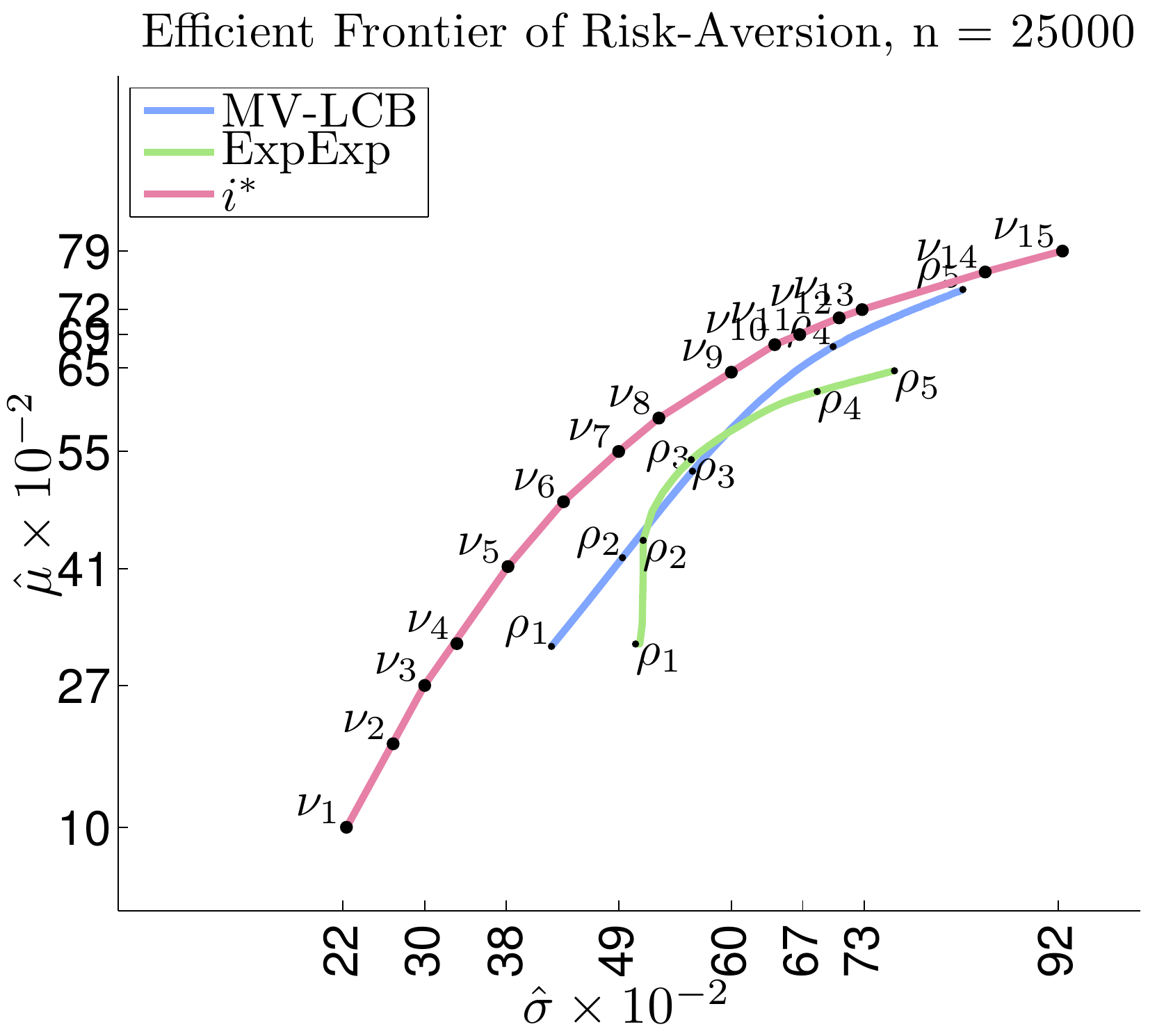}
\includegraphics[width=0.32\textwidth]{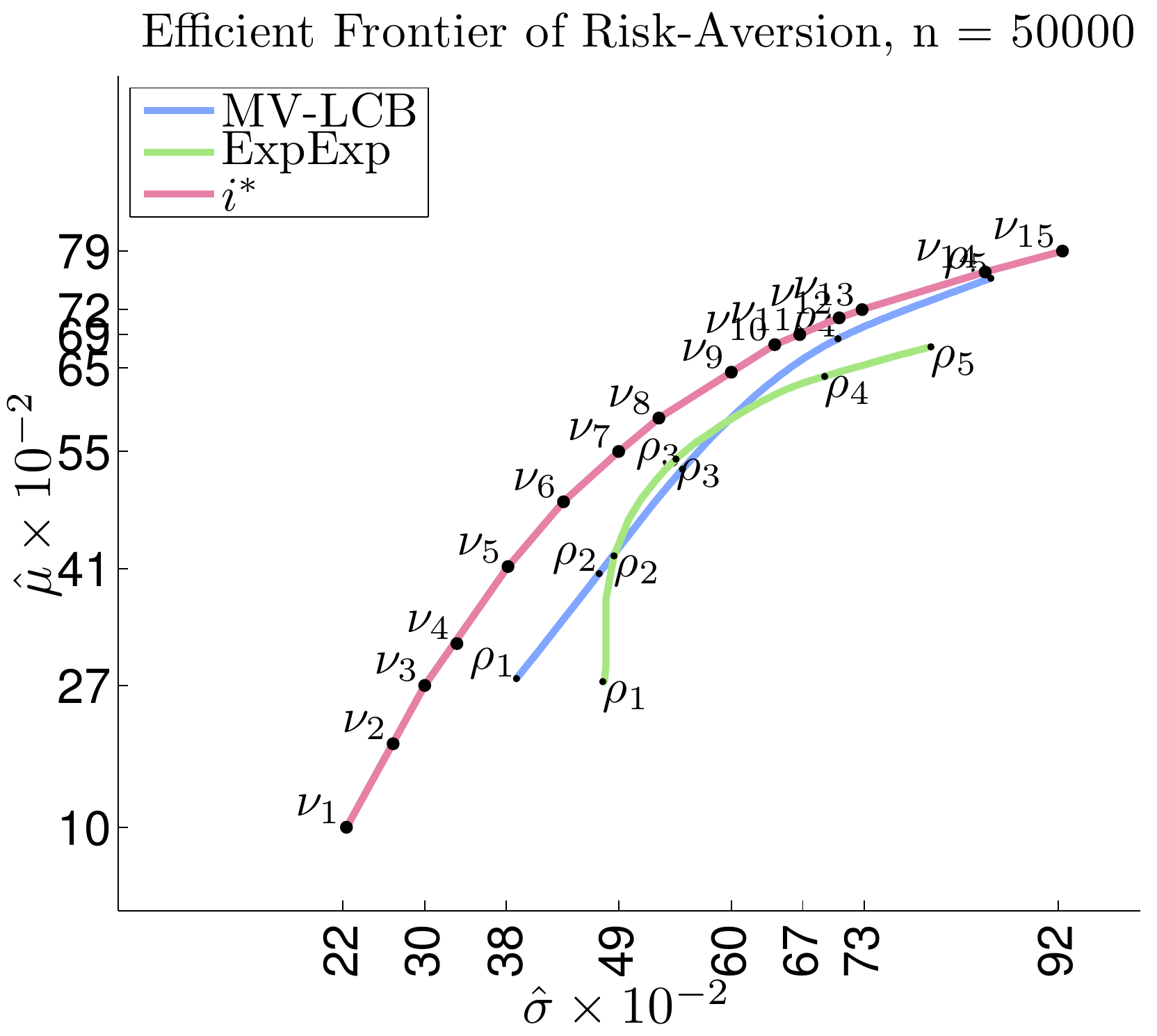}
\includegraphics[width=0.32\textwidth]{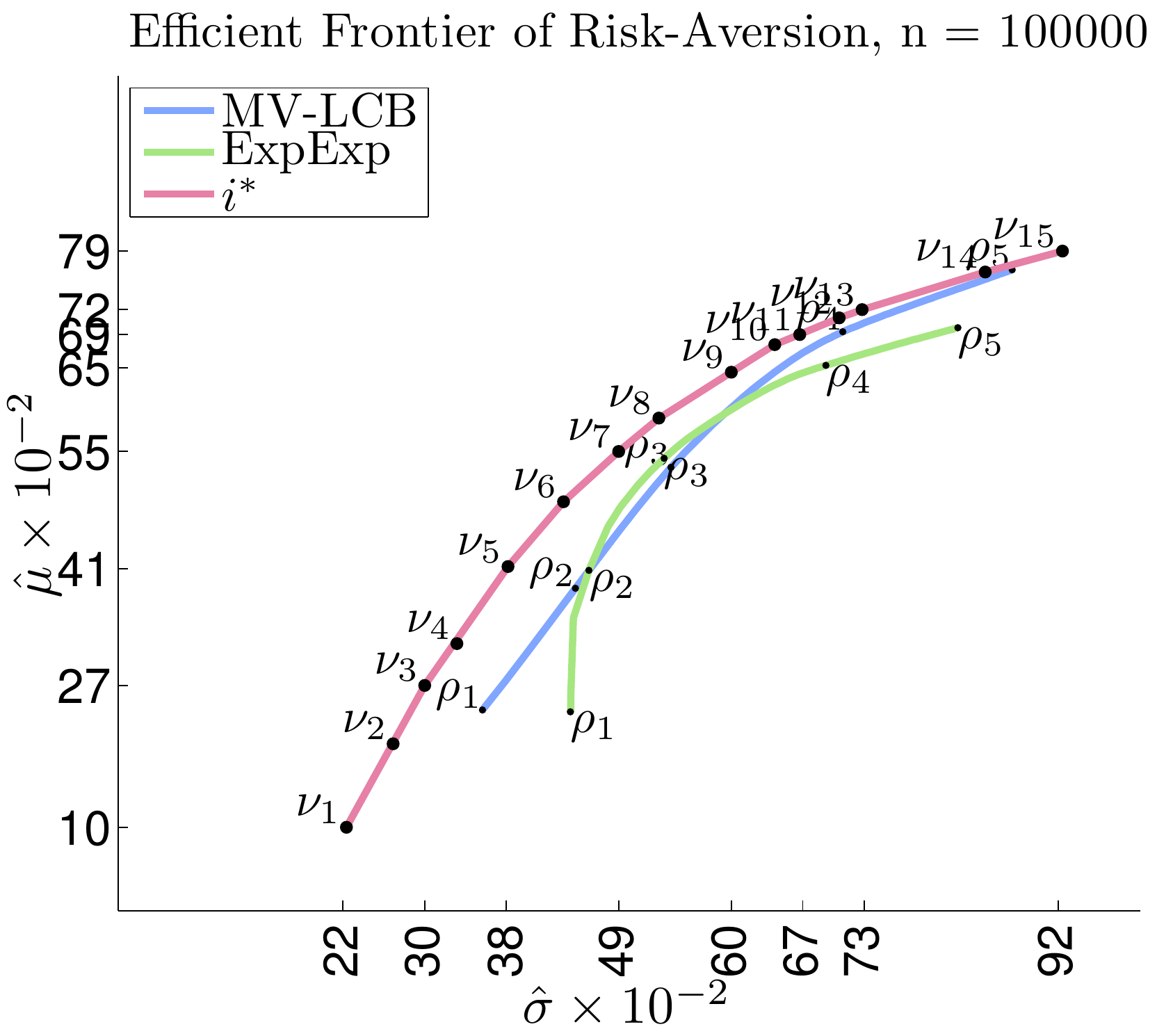}
\includegraphics[width=0.32\textwidth]{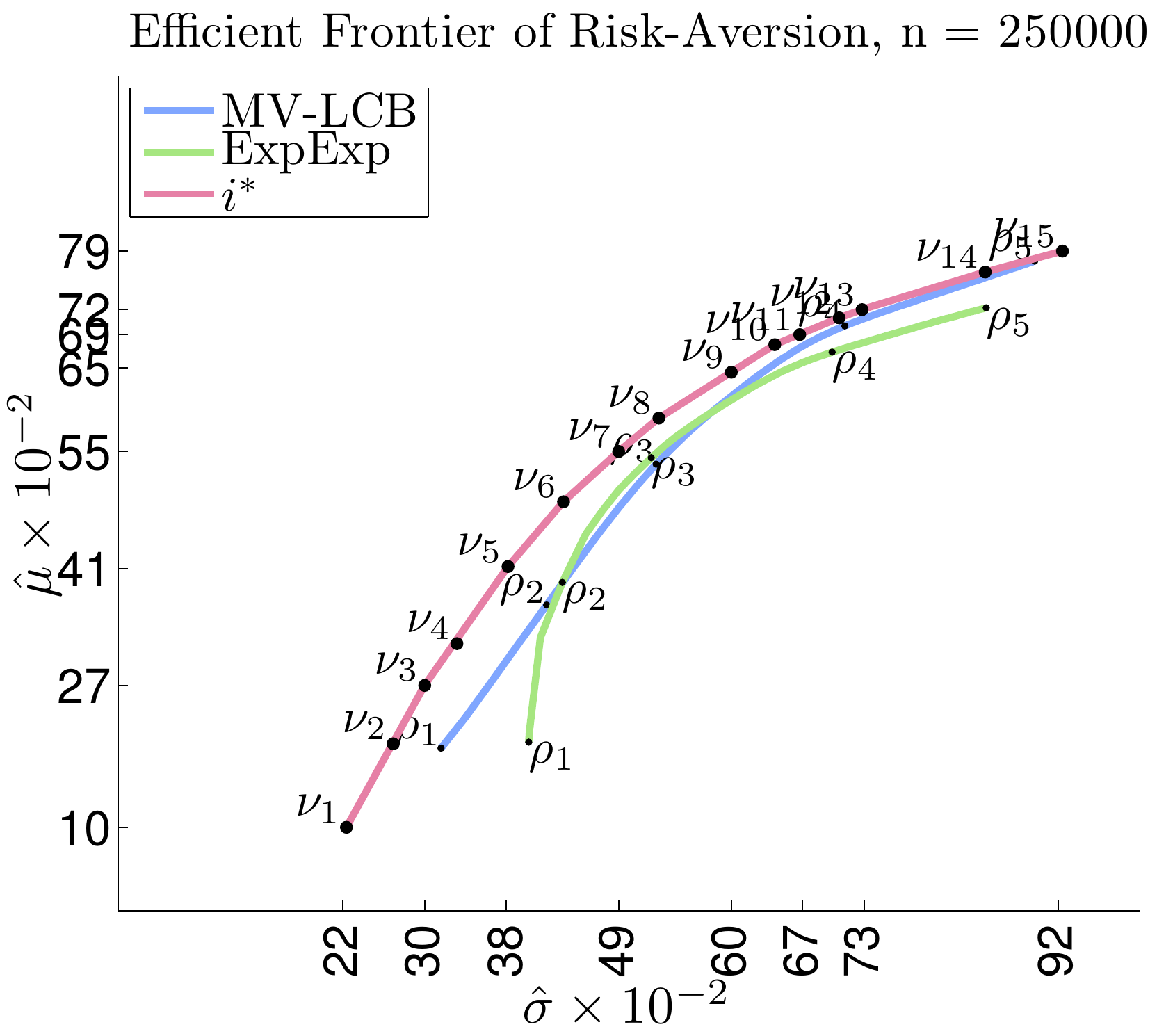}
\end{center}
\caption{Risk tolerance sensitivity of \textsl{MV-LCB} and \textsl{ExpExp} for \textit{Configuration 2}.}\label{f:ef.diff}
\vspace{-0.4cm}
\end{figure*}

In Section~\ref{s:simulations} we report numerical results demonstrating the composition of the regret and performance of algorithms with only 2 arms in the case of variance minimization. Here we report results for a wide range of risk tolerance $\rho\in [0.0;10.0]$ and $K=15$ arms. We set the mean and variance for each of the 15 arms so that a subset of arms is always dominated (i.e., for any $\rho$, $\MV^{\rho}_i > \MV^\rho_{i^*_\rho}$) demonstrating the effect of different $\rho$ values on the position of the optimal arm $i^*_\rho$.

In Figure~\ref{f:mvlcb} we arranged the true values of each arm along the red fronteir and the $\rho$-directed performance of the algorithms in a standard deviation--mean plot. The green and blue lines show the standard deviation and mean for the performance of each algorithm for a specific $\rho$ setting and finite time $n$, where each point represents the resulting mean--standard deviation of the sequence of pulls on the arms by the algorithm with that specific value of $\rho$. The gap between the $\rho$ specific performance of the algorithm and the corresponding optimal arm along the red frontier represents the regret for the specific $\rho$ value. Accordingly, the gap between the algorithm performance curves represents the gap in performance with regard to \textsl{MV-LCB} versus \textsl{ExpExp}. Where a lot of arms have big gaps (e.g., all the dominated arms have a large gap for any value of $\rho$), \textsl{MV-LCB} tends to perform better than \textsl{ExpExp}. The series of plots represent increasing values of $n$ and demonstrate the relative algorithm performance versus the optimal red frontier. The set of plots represent the two settings reported in Figure~\ref{f:config}. We chose the values of the arms so as to have configurations with different complexities. In particular, configuration 1 corresponds to ``easy'' problems for \textsl{MV-LCB} since the arms all have quite large gaps (for different values of $\rho$) and this should allow it to perform well. On the other hand, the second configuration has much smaller gaps and, thus, higher complexity. According to the bounds for \textsl{MV-LCB} we know that a good proxy for its learning complexity is represented by the term $\sum_i 1/\Delta_{i,\rho}^2$. In Figure~\ref{f:config} we report such complexity for different values of $\rho$ and, as it can be noticed, configuration 2 has always a higher complexity than configuration 1.

As we notice, in both configurations the performance of \textsl{MV-LCB} and \textsl{ExpExp} approach one of the optimal arms $i^*_\rho$ for each specific $\rho$ as $n$ increases. Nonetheless, in configuration 1 the large number of suboptimal arms (e.g., arms with large gaps) allows \textsl{MV-LCB} to outperform \textsl{ExpExp} and converge faster to the optimal arm (and thus zero regret). On the other hand, in configuration 2 there are more arms with similar performance and for some values of $\rho$ \textsl{ExpExp} eventually achieves better performance than \textsl{MV-LCB}.

\end{document}